\documentclass[examplefnt]{nowfnt} 

\usepackage{times,epsfig,graphicx}
\usepackage{algorithm,algorithmic}
\usepackage{verbatim}
\usepackage{tabularx}
\usepackage{xspace}
\usepackage{bbm}
\usepackage{bm}
\usepackage{booktabs}  

\usepackage{url}
\usepackage{lmodern}
\usepackage[utf8]{inputenc}
\usepackage{microtype}

\usepackage{amssymb,amsmath}
\usepackage{nicematrix}

\newlength{\smallimage}
        \definecolor{rel}{rgb}{.1,.6,.2}
        \definecolor{nrl}{rgb}{1,1,1}
        \definecolor{qim}{rgb}{1,1,1}

\makeatletter

\def\eg{\emph{e.g.\,}}

\def\ie{\emph{i.e.\,}}

\def\etc{\emph{etc.\,}}
\def\vs{\emph{versus\,}}
\def\wrt{w.r.t.\,}

\def\iid{i.i.d.\,}

\newcommand{\secref}[1]{Section~\ref{sec:#1}}
\newcommand{\fig}[1]{Figure~\ref{fig:#1}}
\newcommand{\tabl}[1]{Table~\ref{tab:#1}}

\def\be{\begin{equation}}
\def\ee{\end{equation}}
\def\bea{\begin{eqnarray}}
\def\eea{\end{eqnarray}}
\def\ben{\begin{eqnarray*}}
\def\een{\end{eqnarray*}}

\def\bi{\begin{itemize}}
\def\ei{\end{itemize}}

\newcommand{\btab}[1]{\begin{tabular}{#1}}
\newcommand{\etab}{\end{tabular}}
\newcommand{\ba}[1]{\begin{array}{#1}}
\newcommand{\ea}{\end{array}}

\def\Loss{\mathcal L}

\DeclareMathOperator*{\argmax}{\mathrm{argmax}}

\def\<{\langle}
\def\>{\rangle}

\newcommand{\bfx}{{\bf x}}
\newcommand{\bfy}{{\bf y}}

\newcommand{\calD}{{\mathcal D}}

\newcommand{\calS}{{\mathcal S}}
\newcommand{\calT}{{\mathcal T}}

\newcommand{\calX}{{\mathcal X}}
\newcommand{\calY}{{\mathcal Y}}

\newcommand{\Exp}{\mathbb{E}}

\newcommand{\myparagraph}[1]{\vspace{0.1cm}\noindent\textbf{#1.}}



\newcommand{\ith}{$i^\textrm{th}$\,}
\newcommand{\jth}{$j^\textrm{th}$\,}

\definecolor{DarkCoral}{rgb}{0.8, 0.36, 0.27}

\title{Semantic Image Segmentation: Two Decades of Research}

\maintitleauthorlist{
Gabriela Csurka \\
Naver Labs Europe,  France \\
Gabriela.Csurka@naverlabs.com
\and
Riccardo Volpi\\
Naver Labs Europe,  France \\
Riccardo.Volpi@naverlabs.com
\and
Boris Chidlovskii \\
Naver Labs Europe,  France \\
Boris.Chidlovskii@naverlabs.com
}

\journaltitle{}

\issuesetup
{%
 copyrightowner = xxx,
 volume        = 14,
 issue         = 1-2,
 pubyear       = 2022,
 doi           = 10.1561/0600000095,
 firstpage     = 1,
 lastpage      = 162
 }

\usepackage{mwe}

\author[1]{Csurka,Gabriela}
\author[2]{Volpi,Riccardo}
\author[3]{Chidlovskii,Boris}

\affil[1]{Naver Labs Europe,  France; Gabriela.Csurka@naverlabs.com}
\affil[2]{Naver Labs Europe,  France; Riccardo.Volpi@naverlabs.com}
\affil[3]{Naver Labs Europe,  France; Boris.Chidlovskii@naverlabs.com}

\articledatabox{\nowfntstandardcitation}

\begin{document}

\makeabstracttitle

\begin{abstract}

Semantic image segmentation (SiS) plays a fundamental role in a broad variety of computer vision applications, providing key information for the global understanding of an image. This survey is  an effort to summarize two decades of research in the field of SiS, where we propose a literature review of solutions starting from early historical methods followed by an overview of more recent deep learning methods including the latest trend of using transformers. We complement the review by discussing particular cases of the weak supervision and side machine learning techniques that can be used to improve the semantic segmentation such as curriculum, incremental or self-supervised learning.

State-of-the-art SiS models rely on a large amount of annotated samples, which are more expensive to obtain than labels for tasks such as image classification. Since unlabeled data is instead significantly cheaper to obtain, it is not surprising that Unsupervised Domain Adaptation (UDA) reached a broad success within the semantic segmentation community. Therefore, a second core contribution of this book is to summarize five years of a rapidly growing field, Domain Adaptation for Semantic Image Segmentation (DASiS) which embraces the importance of semantic segmentation itself and a critical need of adapting segmentation models to new environments. In addition to providing a comprehensive survey on DASiS techniques, we unveil also newer trends such as multi-domain learning, domain generalization, domain incremental learning, test-time adaptation and source-free domain adaptation. Finally, we conclude this survey by describing datasets and benchmarks most widely used in SiS and DASiS and briefly discuss related tasks such as instance and panoptic image segmentation, as well as applications such as medical image segmentation. 

We hope that this book will provide researchers across academia and industry with a comprehensive reference guide and will help them in fostering new research directions in the field.

\end{abstract}

\chapter*{Preface}

\textit{Semantic image segmentation} (SiS) plays a fundamental role 
towards a general understanding of the image content and context. 
In concrete terms, the goal is to label image pixels with the corresponding semantic classes and to provide boundaries of the class objects, easing the understanding of object appearances and the spatial relationships between them. Therefore, it represents an important task towards the design of artificial intelligent systems. Indeed, systems such as intelligent robots or autonomous cars should have the ability to coherently understand visual scenes, in order to perceive and reason about the environment holistically.

Hence, semantic scene understanding is a key element of advanced driving assistance systems (ADAS) and autonomous driving (AD) \citep{TeichmannIVS18MultiNetRealTimeJointSemReasoningAD,HofmarcherBC19VisualSceneUnderstandingAutonomousDriving} as well as robot navigation \citep{ZurbruggX22EmbodiedActiveDASiSInformativePathPlanning}.
The information derived from visual signals is generally combined with other sensors such as radar and/or LiDAR to increase the robustness of the artificial agent's perception of the world \citep{YurtseverIEEE20ASurveyADCommonPracticesEmergingTechnologies}. Semantic segmentation fuels applications in the fields of robotic control and task learning~\citep{FangICRA18MultiTaskDADeepInstanceGraspingSimulation,HongIJCAI18VirtualToRealLearningToControlSemSeg}, medical image analysis (see section~\ref{sec:MedSegm}),  
augmented reality~\citep{DeChicchisTR20SemanticUnderstandingAR,TurkmenTR19SceneUnderstandingThroughSemSegmAR}, satellite imaging~\citep{MaJPRS19DeepLearningRemoteSensingApplications} and many others.

The growth of interest in these topics has also been caused by recent advances in deep learning, which allowed a significant performance boost in many computer vision tasks -- including semantic image segmentation. Understanding a scene at the semantic level has long been a major topic in computer vision, but only  recent progresses on the field have allowed to train machine learning systems robust enough to be integrated in real-world applications.

The success of deep learning methods typically depends on the availability of large amounts of annotated training data, but manual annotation of images with pixel-wise semantic labels is an extremely tedious and time consuming process. As the major bottleneck in SiS is the high cost of manual annotation, many methods rely on graphics platforms and game engines to generate synthetic data and use them to train  segmentation models. The main advantage of such synthetic rendering pipelines is that they can produce a virtually unlimited amount of labeled data. Due to constantly increasing photo-realism of the rendered datasets, the models trained on them yield good performance when tested on real data. Furthermore, they allow to easily diversify data generation, simulating various environments and weather/seasonal conditions, making such data generation pipeline suitable to support the design and training of SiS models for the real world.

While modern SiS models trained on such simulated images can already perform relatively well on real images, their performance can be further improved by domain adaptation (DA) -- and even with \textit{unsupervised domain adaptation} (UDA) not requiring any target labels. This is due to the fact that DA allows to bridge the gap caused by the {\it domain shift} between the synthetic and real images. For the aforementioned reasons, sim-to-real adaptation represents one of the leading benchmarks to assess the effectiveness of~\textit{domain adaptation for semantic image segmentation} (DASiS).

The aim of our book is to overview the research field of SiS. On the one hand, we propose a literature review of semantic image segmentation solutions designed in the last two decades -- including early historical methods and more recent deep learning ones, also covering the recent trend of using transformers with attention mechanism. On the other hand, we devote a large part of the book to survey methods designed \textit{ad hoc} for DASiS. While our work shares some similarities with some of the previous surveys on this topic, it covers a broader set of DASiS approaches and departs from these previous attempts pursuing different directions that are detailed below.

Amongst the existing works surveying SiS methods, we can mention \citet{ThomaX16ASurveySemanticSegmentation} who gives a brief overview of some of the early semantic segmentation and low-level segmentation methods. \citet{LiICDMWS18ASurveyOnSemanticSegmentation} and \citet{GouIJMIR19AReviewSemanticSegmentationDNN} discuss some of the early deep learning-based solutions for SiS. A more complete survey on deep SiS models has been proposed by~\citet{MinaeeX20ImageSegmUsingDeepLearningSurvey}, while~\citet{ZhangAIR20ASurveyOfSemiWeaklySupervisedSiS} focus on reviewing semi- and weakly supervised semantic segmentation models. We cover most of these methods in Chapter \ref{c:semsegm}, where we provide a larger spectrum of the traditional SiS methods in \secref{histSS}. Then, in \secref{deepSS}, we organize the deep SiS methods according to their \textit{most important characteristics}, such as the type of encoder/decoder, attention or pooling layers, solutions to reinforcing local and global consistency. In contrast to the previous surveys, this chapter also includes the latest SiS models that use attention mechanisms and transformers as encoder and/or decoder. One of the core contributions of this chapter is \tabl{semsegm_methods}, which presents a broad set of deep models proposed in the literature, and summarized according to the above mentioned characteristics. Finally, in~\secref{beyond_sis} we review not only semi- and weakly supervised SiS solutions, but also new trends whose goal is improving semantic segmentation, such as curriculum learning, incremental learning and self-supervised learning.

In Chapter~\ref{c:DAsemsegm}, we present and categorize a large number of approaches devised to tackle the DASiS task. Note that previous DA surveys~\citep{GopalanFTCGV15DomainDAVisualRecogn,CsurkaBC17AComprehensiveSurveyDAForVisualApplications,KouwPAMI21AReviewDAWithoutTargetLabels,ZhangX19TransferAdaptationLearningADecadeSurvey,VenkateswaraB20DAinCVwithDeepLearning,SinghB20DomainAdaptationVisualUnderstandings,CsurkaX20DeepVisualDomainAdaptation,WangNC18DeepVisualDASurvey,WilsonTIST20ASurveyDeepUDA} address generic domain adaptation approaches that mainly cover image classification and mention only a few adaptation methods for SiS. Similarly, in recent surveys on domain generalization~\citep{WangX20GeneralizingToUnseenDomainsSurveyDG,ZhouX20DomainGeneralizationSurvey}, online learning~\citep{HoiX18OnlineLearningComprehensiveSurvey} and robot perception~\citep{GargFTR20SemanticsForRoboticMappingPerceptionIInteractionSurvey}, several DA solutions are mentioned, but yet DASiS received only a marginal attention here. The most complete survey -- and therefore most similar to the content of our Chapter~\ref{c:DAsemsegm} -- is by~\citet{ToldoX20UnsupervisedDASSReview}, which also aimed at reviewing the recent trends and advances developed for DASiS. Nevertheless, we argue that our survey extends and enriches it in multiple ways. 
First, our survey is more recent in such a quickly evolving field as DASiS, so we address an important set of recent works appeared after their survey. 
Second, while we organize the DASiS methods according to how domain alignment is achieved similarly to \citep{ToldoX20UnsupervisedDASSReview} -- namely on {\it image, feature or output level} -- we  complement it with different ways of grouping DASiS approaches, namely based on their most important~\textit{characteristics}, such as the backbone used for the segmentation network, the type and levels of domain alignments, any complementary techniques used and finally the particularity of each method compared to the others. We report our schema in \tabl{semsegm_methods}, which represents one of core contributions of this book.
Third, we survey a large set of complementary techniques in \secref{dass_improvements} that can help boosting the adaptation performance, such as self-training, co-training, self-ensembling and model distillation. 

Finally, in \secref{beyond_dass} we propose a detailed categorization of some of the \textit{related DA tasks} -- such as multi-source, multi-target domain adaptation, domain generalization, source-free adaptation, domain incremental learning, etc. -- and survey solutions proposed in the literature to address them. None of the previous surveys has such a comprehensive survey on these related DA tasks, especially what concerns semantic image segmentation.
 
To complement the above two Chapters, which represent the core contributions of our book, 
we further provide in Chapter~\ref{c:benchmarks} a list of the datasets and benchmarks typically used to evaluate SiS and DASiS methods, covering the main metrics  and discuss different SiS and DASiS evaluation protocols.
Furthermore, in Chapter~\ref{c:relatedtasks} we propose a short overview of the literature for three tasks strongly related to SiS, namely instance segmentation in~\secref{instanceSegm}, panoptic segmentation in~\secref{panopticSegm} and medical image segmentation in~\secref{MedSegm}.

We hope that our book, with its comprehensive survey of the main trends
in the field of semantic image segmentation, will provide researchers both across academia and in the industry a solid bases and a helpful background to help them developing new methods and fostering new research directions.


\chapter{Semantic Image Segmentation (SiS)}
\label{c:semsegm}

Semantic image segmentation (SiS) -- sometimes referred to as content-based image segmentation --  is a computer vision problem where the task is to determine to which semantic class each pixel of an image belongs to. 
Typically, this problem is approached in a supervised learning fashion, by relying on a dataset of images annotated at pixel level, and training with them a machine learning model to perform the task. 
This task is inherently more challenging than image classification, where the aim is to predict a single label for a given image. Furthermore, the task is more than the extension of image classification to  pixel-level classification, as in contrast to image classification where each image can be considered independently from the others, in SiS the neighboring pixels are strongly related with each other and their labeling should be considered together, tackling the problem as an image partitioning into semantic regions. Hence, while the models in general indeed tries to minimize the \textit{pixel-level cross-entropy loss} between the \textit{ground-truth} (GT) segmentation map and the \textit{predicted} segmentation map, additional constraints or regularizing terms are necessary in general to ensure, for example, local labeling consistency or to guide segmentation boundary smoothness\footnote{For more details on different losses for SiS we refer the reader to \secref{SiS_losses}}.

The name of the task, \textit{semantic image segmentation}, reflects the goal of determining the nature, \ie semantics, of different parts of an image. Semantic labels in general refers to
\textit{things} such as ``car'', ``dog'', ``pedestrian'' or \textit{stuff} such as ``vegetation'', ``mountain'', ``road'', ``sky''. Things  and stuff are terms extensively used in the literature, where the former includes classes associated with \textit{countable} instances and the latter indicates  classes associated with the \textit{layout of a scene}. Note that a related, still different problem is low-level image segmentation (not addressed in this survey), which consists in an unsupervised partitioning of an image into coherent regions according more to some low-level cues, such as color, texture or depth. 
Another related field is instance segmentation (discussed in~\secref{instanceSegm}), which differs from semantic segmentation as the latter treats multiple object instances with the same semantics as a single entity, while the former treats multiple objects of the same class as distinct individual objects (or instances). The extension of instance segmentation to panoptic segmentation, where \textit{stuff} is also taken into account, is further discussed in~\secref{panopticSegm}.

The aim of this chapter is to provide a comprehensive literature review of SiS methods proposed since the beginning of the field. It is organized as follows. 
In \secref{histSS} we first provide an overview of the historical SiS methods preceding the deep learning era. 
Then, in \secref{deepSS}, we focus on deep learning-based models proposed for SiS, following~\citet{MinaeeX20ImageSegmUsingDeepLearningSurvey}, and propose to categorize them by their main principles. In particular we collect the methods in~\tabl{semsegm_methods} 
detailing their main characteristics, such as the encoder and decoder used, whether they rely on attention modules, how they tackle the semantic consistency within regions, on what kind of data the models were tested on, and what are the main specific of each model. 

Finally, we conclude this chapter with~\secref{beyond_sis} where we discuss some of the semantic segmentation solutions that depart from the classical setting, such as exploiting the unlabeled data (\secref{SSLearn}), relying on weak or none annotations (\secref{WSSiS}), 
exploiting curriculum learning strategies 
(\secref{currLearn}), learning the semantic classes incrementally (\secref{class-inc}) or fine-tuning a self-supervised pre-trained model (\secref{SelfLearn}). 
Note that the models proposed in this chapter have generally been tested \textit{in domain} -- that is, training and testing data come from the same data distribution.
The case when training and testing data come from two different distributions, -- \ie the model  trained on a source domain (\eg synthetic environment) needs to be adapted to a new target domain (\eg real world), -- is discussed in details in~Chapter~\ref{c:DAsemsegm}.

\section{Historical SiS Methods}
\label{sec:histSS}

\begin{figure}[ttt]
\begin{center}
\includegraphics[width=0.9\textwidth]{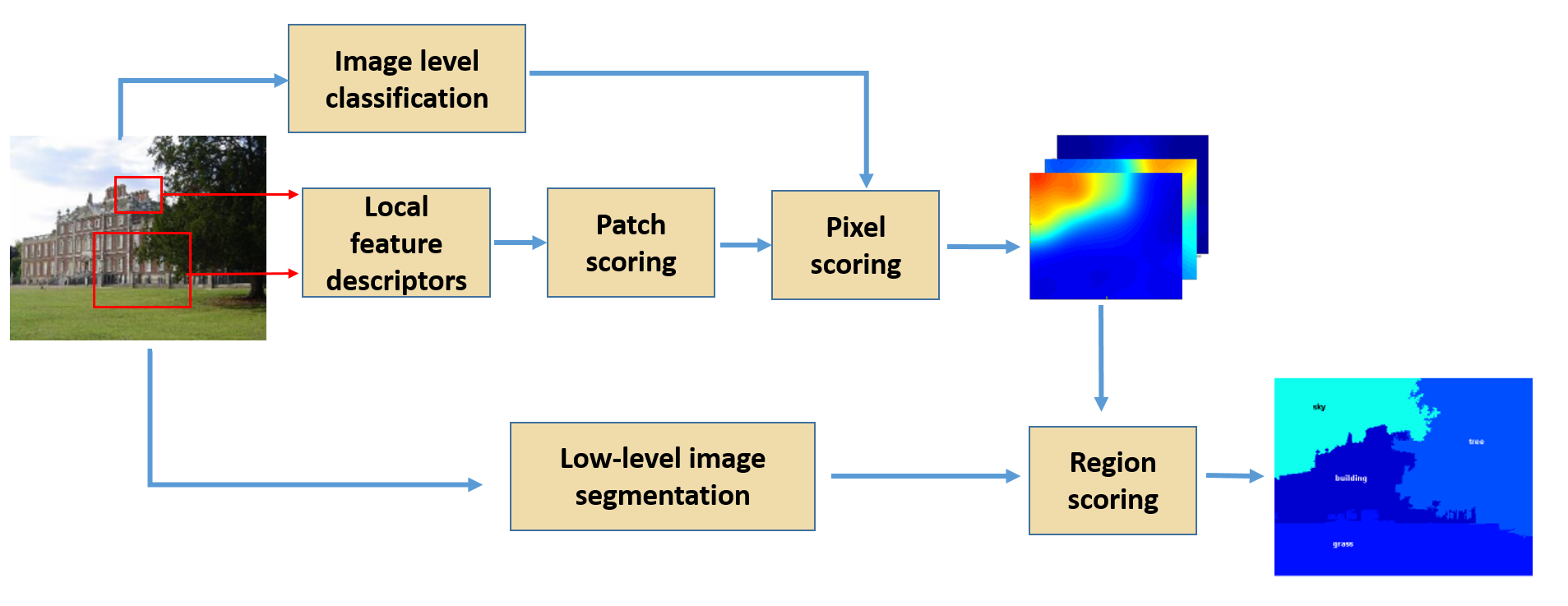}
\caption{In the model proposed in~\citep{CsurkaIJCV11AnEfficientApproachSemSegm}, the local appearance, global and local consistencies are addressed independently. First, for patches considered at multiple scale SIFT~\citep{LoweIJCV04DistinctiveImageFeaturesScaleInvariantKeypoints} and local color statistics are extracted and transformed into high-level Fisher vector representations~\citep{PerronninCVPR07FisherKernels} allowing fast and efficient patch scoring. The global consistency is addressed by an image-level classifier, which is used to filter out improbable classes, while the local consistency is ensured by low-level segmentation assigning to each super-pixel the semantic label based on averaged class probabilities (Figure based on~\citep{CsurkaIJCV11AnEfficientApproachSemSegm}).}
\label{fig:PatchScoringSIS}
\end{center}
\end{figure}

Methods preceding the deep learning era mainly focused on three directions to approach the segmentation problem: 1) local appearance of semantic classes, 2) local consistency of the labeling between locations and 3) how to incorporate  prior knowledge into the pipeline to improve the segmentation quality. These three aspects are addressed independently in the semantic segmentation pipeline as illustrated example in~\fig{PatchScoringSIS}; they can also be approached within a unified probabilistic framework such as a Conditional Random Field (CRF), as described in \secref{globloc_consistency}. The latter methods enable at training time a joint estimation of the model parameters and therefore ensure at test time a globally consistent labeling. Yet, they carry a relatively high computational cost. Note that the three aspects are also addressed by the deep learning models, where they are jointly learned in an end-to-end manner, together with the main supervised task, as we will see in~\secref{deepSS}. 

In which follows we briefly discuss how the above three components were addressed and combined by the methods proposed  before the deep learning era.

\subsection{Modeling local appearance}
\label{sec:apperance}

The local appearance can be defined at different levels, including a representation proposed at every pixel location \citep{HeCVPR04MultiscaleCRFImageLabeling,KumarICCV05AHierarchicalFieldUnifiedContextBasedClassif,SchroffICVGIP06SingleHistogramClassModelsImgSegm,LiCVPR09TowardsTotalSceneUnderstanding}, patches on a regular grid  \citep{VerbeekCVPR07RegionClassificationMRFMAspectModels,LarlusIJCV10CategoryLevelObjectSegmBOVDirichletProcessesRF}, positions of interest points~\citep{LeibeECCVWS04CombinedObjectCatSegmWithImplicitShapeModel,CaoICCV07SpatiallyCoherentLatentTopicModelObsSegmClass,YangCVPR07MultipleClassSegmentationUnifiedFrameworkMeanShiftPatches} or regions obtained through low-level segmentation referred to as \textit{super-pixels}~\citep{BorensteinECCV04LearningToSegment,CaoICCV07SpatiallyCoherentLatentTopicModelObsSegmClass,YangCVPR07MultipleClassSegmentationUnifiedFrameworkMeanShiftPatches}. Note that a sparse description in general enables faster processing and still provides excellent accuracy compared to the dense description. The same method can sometimes consider the combination of multiple representations such as using interest points and regions~\citep{CaoICCV07SpatiallyCoherentLatentTopicModelObsSegmClass,YangCVPR07MultipleClassSegmentationUnifiedFrameworkMeanShiftPatches} or using dense sampling and regions~\citep{KumarICCV05AHierarchicalFieldUnifiedContextBasedClassif}.

Amongst early local appearance features we can mention   
raw image representations~\citep{SchroffICVGIP06SingleHistogramClassModelsImgSegm}, combination of Gaussian filter outputs, colors, and locations computed for each pixel called textons~\citep{ShottonECCV06TextonBoostJointAppearanceShapeContextSemSegm}, SIFT~\citep{LoweIJCV04DistinctiveImageFeaturesScaleInvariantKeypoints}, local color statistics~\citep{ClinchantCLEFWN07XRCEParticipation}. As mentioned above, the local features are often computed on image patches extracted either on a (multi-scale) grid~\citep{VerbeekNIPS07SceneSegmCRFfromPartiallyLabeledImages,CsurkaIJCV11AnEfficientApproachSemSegm} or at detected interest point  locations~\citep{CaoICCV07SpatiallyCoherentLatentTopicModelObsSegmClass,YangCVPR07MultipleClassSegmentationUnifiedFrameworkMeanShiftPatches}. 

These local representations are often clustered into so called \textit{visual words} \citep{CsurkaECCVWS04VisualCategorizationBagsKeypoints,JurieICCV05CreatingEfficientCodebooksVisRecogn} and the local image entity (pixel, patch, super-pixels) is labeled by simply assigning the corresponding feature to the closest visual word~\citep{SchroffICVGIP06SingleHistogramClassModelsImgSegm} or fed into a classifier~\citep{PlathICML09MultiClassImSegmCRFGlobalClassif}. 
Alternatively, these low-level local features can also be used to build higher level representations such as Semantic Texton Forest~\citet{ShottonECCV06TextonBoostJointAppearanceShapeContextSemSegm}, Bag of Visual Words~\citep{CsurkaECCVWS04VisualCategorizationBagsKeypoints}, Fisher Vectors~\citep{PerronninCVPR07FisherKernels}, which are fed into a classifier that predicts class labels at patch level \citep{CsurkaIJCV11AnEfficientApproachSemSegm,LadickyICCV09AssociativeHierarchicalCRFsSiS}, pixel level \citep{ShottonIJCV09TextonBoostForImageUnderstanding} or region level \citep{YangCVPR07MultipleClassSegmentationUnifiedFrameworkMeanShiftPatches,GonfausCVPR10HarmonyPotentialsJointClassifSegm,HuICVGIP12OnRegionsSemanticImgSegm}.

Topic models, such as probabilistic Latent Semantic Analysis~\citep{HofmannML01UnsupervisedLearningPLS} and Latent Dirichlet Allocation~\citep{BleiJMLR03LatentDirichletAllocation} consider the bag-of-words as a mixture of several \textit{topics} and represent a region as a distribution over visual words. Such representations have been extended to image segmentation by explicitly incorporating spatial coherency in the model to encourage similar latent topic assignment for neighboring regions with similar appearance~\citep{CaoICCV07SpatiallyCoherentLatentTopicModelObsSegmClass} or by combining topic models with Random Fields~\citep{OrbanzECCV06SmoothImageSegmNonparametricBayesianInference,VerbeekCVPR07RegionClassificationMRFMAspectModels,LarlusIJCV10CategoryLevelObjectSegmBOVDirichletProcessesRF}.

\subsection{Reinforcing local and global consistency}
\label{sec:globloc_consistency}

To reinforce the segmentation consistency, the local appearance representation and its context  are generally incorporated within a Random Field (RF) framework, mainly the Markov Random Field (MRF)~\citep{VerbeekCVPR07RegionClassificationMRFMAspectModels,GouldIJCV08MultiClassSegmRelativeLocationPrior,KatoFTSP12MarkovRandomFieldsImageSegm} or the Conditional Random Field (CRF)~\citep{ShottonECCV06TextonBoostJointAppearanceShapeContextSemSegm,HeCVPR04MultiscaleCRFImageLabeling,VerbeekNIPS07SceneSegmCRFfromPartiallyLabeledImages}. While the MRF is generative in nature, the CRF directly models the conditional probability of the labels given the features. 

Note that the \textit{unary potentials} in these RF models can be pixels~\citep{ShottonECCV06TextonBoostJointAppearanceShapeContextSemSegm}, patches~\citep{VerbeekNIPS07SceneSegmCRFfromPartiallyLabeledImages,PlathICML09MultiClassImSegmCRFGlobalClassif,LarlusIJCV10CategoryLevelObjectSegmBOVDirichletProcessesRF} or super-pixels~\citep{LucchiICCV11AreSpatialGlobalConstraintsReallyNecessarySIS,LempitskyNIPS11APylonModelSemSegm} represented by a corresponding appearance feature as described in Section~\ref{sec:apperance}. 

\begin{figure}[ttt]
\begin{center}
\includegraphics[width=0.9\textwidth]{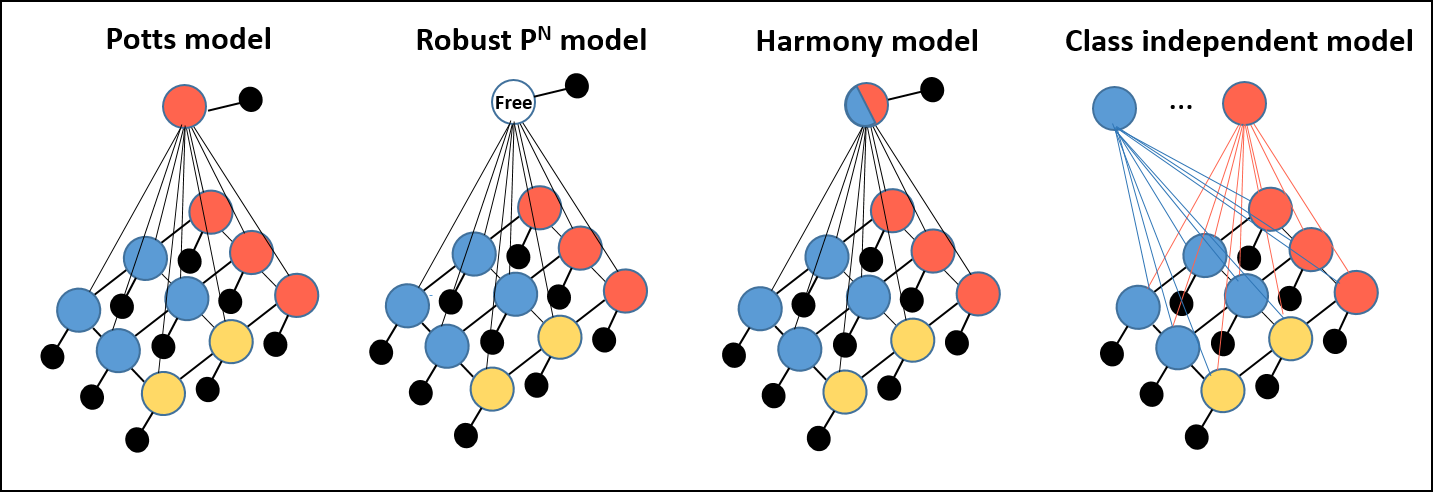}
\caption{Conditional Random Field (CRF) with different increasingly more sophisticated edge potentials. From left to right: \textbf{Potts model~\citep{WuRMP82ThePottsModel,BoykovICCV01InteractiveGraphCuts4OptimalBoundary}} penalizing all local nodes with a label different from the global node;~\textbf{The Robust $P^N$ model \citep{KohliIJCV09RobustHigherOrderPotentialsEnforcingLabelConsistency}} that 
adds an extra “free label” to the Potts model in order to not penalize local nodes;~\textbf{Harmony model~\citep{GonfausCVPR10HarmonyPotentialsJointClassifSegm}} allowing different labels to coexist in a power set;~\textbf{Class independent model \citep{LucchiICCV11AreSpatialGlobalConstraintsReallyNecessarySIS}} modeling each classes with its own global node to make the inference more tractable (Figure based on~\citep{LucchiICCV11AreSpatialGlobalConstraintsReallyNecessarySIS}).}
\label{fig:CRF}
\end{center}
\end{figure}

In these probabilistic frameworks, label dependencies are modeled by a random field (MRF or CRF), and an optimal labeling is determined usually by energy minimization. Prior information can be imposed through \textit{clique potentials} between the nodes in the RF graph (as illustrated in \fig{CRF}). The most often used \textit{edge potentials} are the Potts model~\citep{WuRMP82ThePottsModel}, which penalizes class transitions between neighboring nodes, and the contrast-sensitive Potts  model~\citep{BoykovICCV01InteractiveGraphCuts4OptimalBoundary}, which includes a term reducing the cost of a transition in high contrast regions likely corresponding to object boundaries. 

To enforce region-level consistency, higher order potentials can be added to the CRF model in order to ensure that all pixels within a low-level region have the same label (see examples in \fig{CRF}). As such, \citet{KohliIJCV09RobustHigherOrderPotentialsEnforcingLabelConsistency}
propose the Robust $P^N$ model that adds an extra \textit{free label} to the Potts model in order not to penalize local nodes. \citet{KrahenbuhlNIPS11EfficientInferenceFullyConnectedCRFswithGaussianEdgePotentials} propose a fully connected dense CRF that models the pairwise dependencies between all pairs of pixels with pairwise edge potentials defined by a linear combination of Gaussian kernels, making the inference highly efficient.

The associative hierarchical CRF model by \citet{LadickyICCV09AssociativeHierarchicalCRFsSiS} allows to incorporate context from multiple quantization levels (pixel, segment, and segment union/intersection) in a joint optimisation framework using graph cut-based move-making algorithms. The Harmony potentials~\citep{GonfausCVPR10HarmonyPotentialsJointClassifSegm} model global preferences where any possible combination of class labels can be encoded; this enforces the consistency between local and global label assignments of the nodes. In the Pylon model~\citep{LempitskyNIPS11APylonModelSemSegm}, each image is represented by a hierarchical segmentation tree, and the resulting energy -- combining unary and boundary terms -- is optimized using the graph cut.
\citet{PlathICML09MultiClassImSegmCRFGlobalClassif} uses a CRF-based on an  multi-scale pre-segmentation of the image into patches, which couples local image features with image-level multi-class SVM to provide the local patch evidences. Instead,~\citet{LucchiICCV11AreSpatialGlobalConstraintsReallyNecessarySIS} propose to model each class with its own global node to make the inference more tractable (see~\fig{CRF}). In the LayoutCRF model~\citep{WinnCVPR06TheLayoutConsistentRFSiSPartiallyOccludedObjects},
the pairwise potentials are asymmetric and impose local spatial constraints which ensures having a consistent layout whilst allowing to cope with object deformations.

In alternative to the RF framework, segmentation methods often ensure local consistency by relying
on images decomposed into super-pixels. Such unsupervised partitioning of the image is obtained with low-level segmentation methods such as a Mean Shift~\citep{ComanicuPAMI02MeanShiftRobustApproachFeatureSpaceAnalysis} or hierarchical image segmentation \citep{ArbelaezPAMI11ContourDetectionHierarchicalSegmentation}. A class label is assigned for each super-pixel either in a post processing step~\citep{CsurkaIJCV11AnEfficientApproachSemSegm} (see \eg \fig{PatchScoringSIS}) or by relying on region descriptors and a model predicting class labels at super-pixel level~\citep{YangCVPR07MultipleClassSegmentationUnifiedFrameworkMeanShiftPatches,PantofaruECCV08ObjectRecognitionIntegratingMultipleImgSegm}. The main limitation of these methods is that there is no possible recovery if a region includes multiple classes. To overcome this limitation, several works propose to consider multiple segmentations, exploiting overlapping sets of regions \citep{GouldICCV09DecomposingSceneGeometricSemanticallyConsistentRegions,PantofaruECCV08ObjectRecognitionIntegratingMultipleImgSegm}, a hierarchy of regions~\citep{GuCVPR09RecognitionUsingRegions,HuICVGIP12OnRegionsSemanticImgSegm}, or a graph of regions~\citep{ChenCVPR11PiecingTogetherSegmentationJigsawUsingContext}.

\subsection{Using prior knowledge} 
\label{sec:priorKnowledge}

Amongst different types of prior knowledge, the global image classification is most often considered -- as easy to obtain, -- where the global scale information is used to filter or to improve the estimation at local scale~\citep{CsurkaIJCV11AnEfficientApproachSemSegm,PlathICML09MultiClassImSegmCRFGlobalClassif,VerbeekNIPS07SceneSegmCRFfromPartiallyLabeledImages}. Further priors considered for SiS are object shape priors used to guide the segmentation process~\citep{KumarCVPR05ObjCut,YangCVPR07MultipleClassSegmentationUnifiedFrameworkMeanShiftPatches} or bounding boxes obtained from object detectors~\citep{LempitskyICCV09ImageSegmBoundingBoxPrior}. \citet{LiCVPR09TowardsTotalSceneUnderstanding} employ the user tags provided by Flickr as an additional cue to infer the presence of an object in the image, while ~\citet{HeECCV06LearningIncorporatingTopDownCuesSIS} use an environment-specific class distribution prior to guide the segmentation. ~\citet{GouldIJCV08MultiClassSegmRelativeLocationPrior} and~\citet{HeCVPR04MultiscaleCRFImageLabeling} explicitly model spatial relationships between different classes.

\section{Deep Learning-based SiS}
\label{sec:deepSS}

\begin{table*}
\begin{center}
{
\tiny
\setlength{\tabcolsep}{4pt}
\begin{NiceTabular}{|l|c|c|c|c|c|c|c|}[code-before = 
\rowcolor{gray!10}{4}
\rowcolor{gray!10}{6}
\rowcolor{gray!10}{8}
\rowcolor{gray!10}{10}
\rowcolor{gray!10}{12}
\rowcolor{gray!10}{14}
\rowcolor{gray!10}{16}
\rowcolor{gray!10}{18}
\rowcolor{gray!10}{20}
\rowcolor{gray!10}{22}
\rowcolor{gray!10}{24}
\rowcolor{gray!10}{26}
\rowcolor{gray!10}{28}
\rowcolor{gray!10}{30}
\rowcolor{gray!10}{32}
\rowcolor{gray!10}{34}
\rowcolor{gray!10}{36}
\rowcolor{gray!10}{38}
\rowcolor{gray!10}{40}
\rowcolor{gray!10}{42}
\rowcolor{gray!10}{44}
\rowcolor{gray!10}{46}
]
\toprule

\textbf{Citation} &  \textbf{Model}  & \textbf{Enco-} 
   &  \textbf{Deco-}&   \textbf{(Self-)}    & \textbf{HR}  & \textbf{Data}   &\textbf{Specificity}\\
 & \textbf{Name} &   \textbf{der} &  \textbf{der} &   \textbf{Attention}  & \textbf{Segm}   & \textbf{type} &  \\
\midrule


\color{blue}{Pinheiro et al. (2014)}
& rCNN & RNN & -  &  - & MSP & IP & different input patch sizes \\

\citet{ByeonCVPR15SceneLabelingLSTM} & 2D-LSTM & LSTM &  - & - &  - & IP & connected LSTM blocks \\

\citet{ChenICLR15SemanticImgSegmFullyConnectedCRFs} & DeepLab & RbFCN & -  & -    & ASPP+dCRF & Obj &  CNN driven cost function \\ 

\citet{DaiICCV15BoxSupExploitingBoundingBoxesSuperviseSiS} & BoxSup & FCN & -& -&  ObjP & Obj & weakly (BBox) supervised \\

\citet{LongCVPR15FullyConvolutionalNetworksSegmentation} & FCN & FCN &  -   & - & C2FR & Obj &  skip connections\\

 \citet{Mostajabi2015FeedforwardSemanticSegmentation} & Zoom-Out & CNNs & - & - & SP & Obj & purely feed-forward architecture \\

\citet{NohICCV15LearningDeconvolutionNNSegmentation} & DeConvNet & CNN &  DCN &   - & ObjP & Obj  & candidate object proposals   \\

\citet{RonnebergerMICCAI15UNetSegmentation} & UNet & FCN & DCN & -   & - &   Med  &  skip connections\\


\citet{YuX15MultiScaleContextAggregationDilatedConvolutions} & MSCA & dCN & - & -    & CRF-RNN & Obj & dilated convolutions \\

\cite{ZhengICCV15ConditionalRandomFieldsAsRNN} & CRF/RNN & FCN &  - &   &  CRF & Obj & RNN inference \\

\color{blue}{Chandra et al. (2016)}
& DG-CRF & FCN & - & - & dCRF & Obj  & Gaussian-CRF modules\\

\citet{ChenCVPR16AttentionToScaleScalAwarSemSegm} & DLab-Att & DeepLab &  - & PFw & CRF & Obj &  multi-scale soft spatial weights \\

\color{blue}{Ghiasi et al. (2016)}
& LRR & FCN & - & - & FP & Obj/AD & reconstr. by deconvolution \\

\citet{LiuICLRW16ParseNetLookingWiderToSeeBetter}  & ParseNet  &  FCN & - &- &  - & Obj  & append global representation\\

\citet{PaszkeX16ENetDeepNNForRealTimeSemSegm} & ENet & dFCN &  PUP & - & - &  AD & optimized for fast inference \\
  
\citet{VisinCVPRWS16ReSegRNNSemSegm} & ReSeg & RNNs &  - & - & - & AD & Gated Recurrent Unit (GRU) \\

\cite{BadrinarayananPAMI17SegnetDeepConvEncoderDecoder} & SegNet & FCN & FCN &  - &  - & AD &  keep maxpool locations\\

\citet{PohlenCVPR17FullResolutionResidualNetworksSegmentation} & FRRN & FCN & DCN  & -  & - & AD & high-res stream \\

\color{blue}{Chaurasia et al. (2017)}
& LinkNet & FCN & DCN &- &-&     AD & skip connections  \\

\citet{FourureBMVC17ResidualConvDeconvGridNetwork4SiS} & GridNet  & FCN & DCN & - &  - & AD &  multiple interconnected streams\\

\citet{LinCVPR17RefineNetMultiPathRefinementNetworks}  & RefineNet & FCN & RCU & - &  C2FR & IP & Chained Residual Pooling \\

\citet{ZhaoCVPR17PyramidSceneParsingNetwork} & PSPNet & RbFCN &  SPP  & - &   - & IP  & auxiliary loss\\


\citet{ChenECCV18EncoderDecoderAtrousSeparableConvSemSegm}  & Deeplab+ &  DeepLab & FCN & - & -  & Obj/AD &  depthwise
separable convolution \\ 

\citet{LiBMVC18PyramidAttentionNetworkSemSegm} & PAN  & RbFCN & GPA & FPA+GPA &  - & Obj/AD  &  spatial pyramid attention\\

\citet{WangWACV18UnderstandingConvolutionSemSegm} & DUC-HDC & HDC & DUC &  - & -  & Obj/AD & hybrid dFCN/dense upsampling\\

\cite{XiaoECCV18UnifiedPerceptualParsingSceneUnderstanding}  & UperNet &  ResNet & FPN &  - & SPP & Obj/Mat  & multi-task framework  \\

\citet{YangCVPR18DenseASPPSemSegmStreetScenes} & DenseASPP & denseNet & - & - & ASPP & AD &  densely connected ASPP \\

\citet{ZhaoECCV18PSANetPointWiseSpatialAttentionNetworkSceneParsing} & PSANet &  RbFCN & - & PSA & - & IP & bi-direction information propagation \\


\cite{FuTIP19StackedDeconvolutionalNetwork4SiS} & SDN & DenseN & sDCN & - & - & Obj &  hierarchical supervision   \\ 
  
\cite{FuCVPR19DualAttentionNetworkSceneSegm} & DANet &  dFCN & - & PAM+CAM & - & IP & dual self-attention \\

\cite{HeCVPR19AdaptivePyramidContextNetworkSemSegm} & APCNet & RbFCN & - &-  &   MSP  &  Obj &  Adaptive Context Module \\

\color{blue}{Teichmann et al. (2019)} & ConvCRF & FCN &   - & & dCRF  & Obj &  CRF inference as convolutions\\

%


\citet{WangECCV20AxialDeepLabStandAloneAxialAttPanopticSegm} & Axial-DLab & DeepLab & - & AAL & - & AD/PanS &  position-sensitive SelfAtt  \\

\cite{YuanECCV20ObjectContextualReprSemSegm} & OCR & HRNEt & MLP & selfAtt &  ObjP & IP & object-contextual representation\\


\citet{AliNIPS21XCiTCrossCovImageTransformers} & XCiT  & ViT & UperNet & XCA & -& 
IP & cross-covariance image transformers \\
\citet{ChuNIPS21TwinsRevisitingDesignSpatialAttVisTransformers} & 
 Twins  &  HTr &  SFPN &    LGA+GSA & - & IP &   spatially separable self-attention \\
 
\citet{GuoICCV21SOTRSegmentingObjectsWithTransformers} & SOTR &  FPN+sTrL & PUP & HTwinT & - &  Obj & twin (column/row) attention \\

\citet{JainX21SeMaskSemanticallyMaskedTransformersSiS} & SeMask & HTr  & SFPN  & SwT+SMA & - & IP & semantic priors through attention \\

\citet{LiuICCV21SwinTransformerHierarchicalViTShiftedWindows} & 
 SwinTr &  UperNet & SFPN & SwT & - & Obj/InstS & self-att within local windows\\

\citet{StrudelICCV21SegmenterTransformerForSemSegm} & Segmenter & ViT & MaskT &  MHSA  & - & IP & class embeddings\\

\citet{RanftlICCV21VisionTransformers4DensePrediction} & DPT & sTrL & RCU  & MHSA  & - & IP &   ViT-Hybrid architecture \\ 

\citet{WangICCV21PyramidVisionTransformerDensePredWithoutConvolutions} & PVT & sTrL+FPN  & SFPN & SRA & - & IP/Inst & progressively shrinking pyramid \\

\citet{XieNIPS21SegFormerSemSegmTransformers} & SegFormer & HTr & MLP & SelfAtt &  -&  IP & positional-encoding-free \\

\citet{ZhengCVPR21RethinkingSiSSeq2SeqPerspTransformers} & SETR & sTrL & FPN & MHSA &  & IP & sequence-to-sequence prediction
\\

\bottomrule
\end{NiceTabular}
} 
\end{center}
\caption{Summary of the state-of-the-art methods, schematized according to their characteristics. 
\textbf{Encoder}:  FCN (fully convolutional network), RbFCN (Resnet based FCN), dFCN (dilated FCN), HDC (hybrid dilated convolutions), ViT (Visual Transformer), HTr (hierarchical transformer), sTrL (stacked transformer layers).
\textbf{Decoder}: DCN (deconvolution network), sDCN (stacked DCN)  MLP (multi layer perceptron), DUC (dense upsampling convolutions), MaskT (Mask Transformer), FPN (feature pyramid networks), SFPN (semantic FPN), RCU (residual convolutional units),  PUP (progressive upsampling).
\textbf{Attention}: PFw (positional feature weight), GPA (global attention upsample), FPA (feature pyramid attention), PAM (position attention module), CAM (channel attention module), PSA (point-wise spatial attention module), AAL (axial attention layer), MHSA (multiheaded self-attention), SwT (Swin Transformer Blocks), HTwinT (hybrid twin transformer layer), SRA (spatial-reduction attention  layer), SAB (SeMask Attention Block), LGA (locally-grouped
attention),  GSA (global sub-sampled attention), cross-covariance attention (XCA).
\textbf{High-resolution (HR) Segmentation}: SP (superpixels), CRF (dense conditional random field), dCRF (dense CRF),  ObjP (object/region proposals), MSP (multi-scale/multi-resolution processing), C2FR (coarse-to-fine refinement), SPP (spatial pyramid pooling), ASPP (atrous SPP).
\textbf{Data type}: Obj (object segmentation such as Pascal VOC), IP (image parsing of \textit{stuff} and \textit{things} such as ADEK20), AD (similar to IP, but focused on autonomous driving datasets such as Cityscapes or Kitti), InstS (instance segmentation), Mat (material segmentation), PanS (panoptic segmentation).
\textbf{Specificity} reports important elements that are not covered by the previously described common characteristics.}
\label{tab:DeepSiS_methods}
\end{table*}


In this section we describe the main types of deep learning-based SiS pipelines, grouping the corresponding methods, similarly to~\citep{MinaeeX20ImageSegmUsingDeepLearningSurvey}, based on their main principles. Concretely, in \secref{deepFeatures} we present a few works where classical models have been revisited with deep features and in~\secref{GraphModSiS} we discuss how deep networks were combined with graphical models. Methods based on Fully Convolutional Networks are surveyed in \secref{FCNSiS} and those using decoders or deconvolutional networks are presented in~\secref{EncodeDecodeSiS}. Several models using Recurrent Neural Networks are briefly reviewed in \secref{ReccNetSiS} and those having pyramidal architectures in \secref{PyramidSiS}. Dilated convolutions allowing to easily aggregate multi-scale contextual information are discussed in \secref{DilatedSiS}; attention 
mechanisms exploited in SiS are addressed in \secref{AttBasedSiS}. Finally, we end the section with very recent transformer-based models reviewed in \secref{tranfSiS}. In addition, we propose \tabl{DeepSiS_methods}, where most deep SiS methods are briefly summarized according to their main characteristics, the type of data they are evaluated on, as well as the specificity of each method compared to the others. 

\subsection{Deep features used in classical models}
\label{sec:deepFeatures}

Following the preliminary work of~\citet{GrangierICML09DeepConvolutionalNetworks4SceneParsing} who show that convolutional neural networks (CNNs) fed with raw pixels can be trained to perform scene parsing with decent accuracy, several methods have been proposed to replace hand crafted appearance representations (see Section~\ref{sec:apperance}) with representations obtained from deep convolutional networks. For example, \citet{FarabetICML12SceneParsingMultiscaleFeatureLearningPurityTrees} train a multi-scale CNN to learn good features for region classification which are used to represent nodes in a segmentation tree. Instead, in  \citep{FarabetPAMI13LearningHierarchicalFeaturesSceneLabeling}, they consider a deep dense feature extractor that produces class predictions for each pixel independently from its neighbors and use these predictions as unary potentials in a CRF graph. They also propose an alternative, where an image is represented first as a hierarchy of super-pixels and the average class distribution for each node is computed from the pixel-level predictions of the network.

\citet{Mostajabi2015FeedforwardSemanticSegmentation} propose a purely feed-forward architecture for semantic segmentation, where they concatenate the super-pixel representations with deep features extracted from a sequence of nested regions of increasing extent. These context regions are obtained by ”zooming out” from the super-pixels all the way to scene-level resolution. \citet{HariharanCVPR15HypercolumnsObjectSegmentation} propose to concatenate features from intermediate layers into so called \textit{hyper-column} representation used as pixel descriptors for object segmentation.

\subsection{Graphical models}
\label{sec:GraphModSiS}

Inspired by the shallow image segmentation methods that integrate local and global context (see Section~\ref{sec:globloc_consistency}), several works propose to combine the strengths of CNNs with CRFs by training them jointly in an end-to-end manner.
As such, the model proposed in~\citep{ChandraECCV16FastExactMultiScaleInferenceSemSegmDeepGaussianCRFs} relies on Gaussian-CRF modules that collect the unary and pairwise terms from the network and propose image hypothesis (scores); these scores are then converted into probabilities using the softmax function and thresholded to obtain the segmentation. In addition, they introduce a multi-resolution architecture to couple information across different scales in a joint optimization framework showing that this yields systematic improvements.

\citet{SchwingX15FullyConnectedDeepStructuredNetworks} propose to jointly train the parameters of a CNN used to define the unary potentials as well as the smoothness terms, taking into account the dependencies between the random variables.  \citet{ChenICLR15SemanticImgSegmFullyConnectedCRFs} treat every pixel as a CRF node and exploit long-range dependencies using CRF inference to directly optimize a deep CNN driven cost function.~\citet{TeichmannBMVC19ConvolutionalCRFsSemSegm} reformulate the CRF inference in terms of convolutions; this allows them to improve the efficiency of the CRF, which is known for being hard to optimize and slow at inference time.

\begin{figure}[ttt]
\begin{center}
\includegraphics[width=0.85\textwidth]{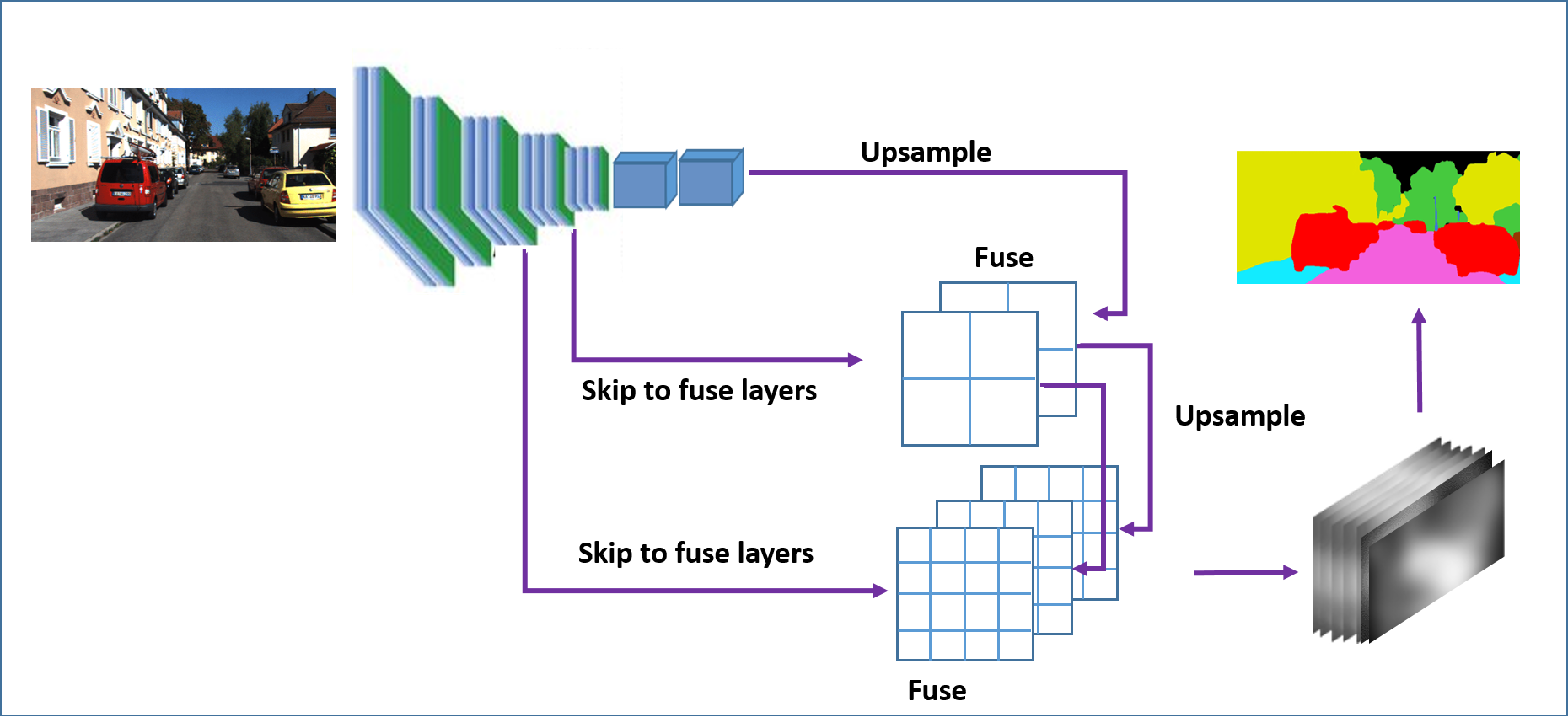}
\caption{The FCN~\citep{LongCVPR15FullyConvolutionalNetworksSegmentation}
transforms the fully connected layers of the classification network to produce class presence heatmaps. The model takes coarse, high layer information at low resolution and upsample it using billinearly initialized deconvolutions. At each upsampling stage, the prediction maps are refined by fusing them with coarser but higher resolution feature maps. The model hence can take an arbitrary size image and produce the same size output, 
suitable for spatially dense prediction tasks such as SiS.}
\label{fig:FCN}
\end{center}
\end{figure}

\subsection{Fully convolutional networks (FCNs)}
\label{sec:FCNSiS}

The preliminary multi-scale convolutional network developed in
\citep{FarabetPAMI13LearningHierarchicalFeaturesSceneLabeling} learns to produce a fairly accurate segmentation map. Still, the model needs to act on a multi-scale pyramid of image windows. Instead, the Fully Convolutional Network (FCN) proposed by \citet{LongCVPR15FullyConvolutionalNetworksSegmentation}, transforming the fully connected layers into convolution layers, enables the net to predict  directly a dense high resolution output (class presence heatmaps) from arbitrary-sized inputs. To further improve the segmentation quality, they propose to obtain such prediction maps at different levels of the network, where the lower resolution outputs are upsampled using billinearly initialized deconvolutions and fused with the coarser but higher resolution feature maps (see illustration in~\fig{FCN}).

\citet{LiuICLRW16ParseNetLookingWiderToSeeBetter} extend FCN by adding a global context vector obtained by global pooling of the feature map, showing that it reduces local confusion. This image-level information is appended to each local feature and the combined feature map is sent to the subsequent layer of the network.

The idea of using only fully convolution layers has been largely adopted as encoder for many SiS models (see Table \ref{tab:DeepSiS_methods}).

\subsection{Encoder-decoder networks}
\label{sec:EncodeDecodeSiS}

\begin{figure}[ttt]
\begin{center}
\includegraphics[width=\textwidth]{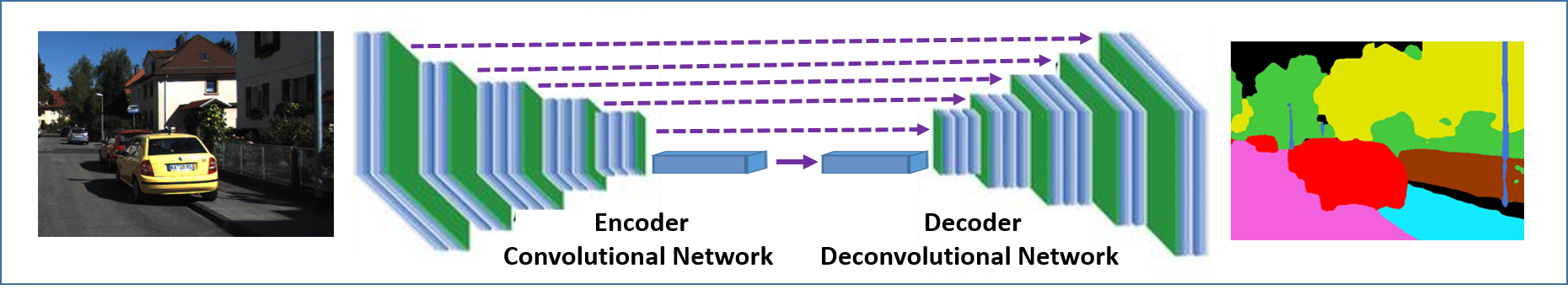}
\caption{DeConvNet~\citep{NohICCV15LearningDeconvolutionNNSegmentation} is composed of a multi-layer convolutional network as encoder and a multi-layer deconvolution network as decoder. The latter is built on the top of the output of the convolutional network, where a series of unpooling, deconvolution and rectification operations are applied  yielding a dense pixel-wise class prediction map. There are connections between the corresponding encoder and decoder layers (dashed lines). In the case of DeConvNet~\citep{NohICCV15LearningDeconvolutionNNSegmentation} the maxpooling locations in the encoder are kept at each level and sent to the unpooling operators in the corresponding level. UNet and LinkNet extend DeConvNet by skip connections; in UNet~\citep{RonnebergerMICCAI15UNetSegmentation} the corresponding features maps from encoder are copied and concatenated to the layers obtained by up-convolutions, and in LinkNet~\citep{ChaurasiaVCIP17LinkNetExploitingEncoderRepr4EfficientSiS} the input of each encoder layer is bypassed to the output of the corresponding decoder.}
\label{fig:DeconvNet}
\end{center}
\end{figure}

SiS models based on encoder-decoder architectures are composed of an encoder -- where the input image is compressed into a latent-space representation that captures the underlying semantic information -- and a decoder that generates a predicted output from this latent representation. In general there are connections between the corresponding encoder and decoder layers allowing the spatial information to be used by the decoder and its upsampling operations (see~\fig{DeconvNet}).   DeConvNet~\citep{NohICCV15LearningDeconvolutionNNSegmentation}, SegNet~\citep{BadrinarayananPAMI17SegnetDeepConvEncoderDecoder}, UNeT~\citep{RonnebergerMICCAI15UNetSegmentation},   and LinkNet~\citet{ChaurasiaVCIP17LinkNetExploitingEncoderRepr4EfficientSiS} (see Figure~\ref{fig:DeconvNet}). 
One such model is DeConvNet~\citep{NohICCV15LearningDeconvolutionNNSegmentation} where the encoder computes low-dimensional feature representations via a sequence of pooling and convolution operations, while the decoder, stacked on top of the encoder, learns to upscale these low-dimensional features via subsequent unpooling and deconvolution operations; the maxpooling locations are kept during encoding and sent to the unpooling operators in the corresponding level. The trained network, applied to a set of candidate object proposals, aggregates them to produce the semantic segmentation of the whole image.  

Another popular method is SegNet~\citep{BadrinarayananPAMI17SegnetDeepConvEncoderDecoder}, where the encoder is similarly composed of consecutive convolutions, followed by  max-pooling sub-sampling layers to increase the spatial context for pixel labelling. However, instead of deconvolution operations, trainable convolutional filters are used by the decoder and combined with the unpooling operations. 

Instead of unpooling, in  UNet~\citep{RonnebergerMICCAI15UNetSegmentation}, the corresponding features maps from the encoder are copied and concatenated to the layers obtained by up-convolutions, and in LinkNet~\citep{ChaurasiaVCIP17LinkNetExploitingEncoderRepr4EfficientSiS}, 
the input of each encoder layer is bypassed to the output of the corresponding decoder layer. In addition, since the decoder is sharing knowledge learned by the encoder at every layer, the decoder can use fewer parameters yielding a more efficient network. 

\citet{PohlenCVPR17FullResolutionResidualNetworksSegmentation} propose a Full Resolution Residual Network (FRRN) that has two processing streams: the residual one which stays at the full image resolution and a Conv-DeconvNet  which undergoes a sequence of pooling and unpooling operations. The two processing streams are coupled using full-resolution residual units. 

\citet{FuICIP17DenselyConnectedDeconvNetSiS} stack multiple shallow deconvolutional networks to improve accurate boundary localization which is extended in \citep{FuTIP19StackedDeconvolutionalNetwork4SiS}  by redesigning the deconvolutional network with intra-unit and inter-unit connections -- to generate more refined recovery of the spatial resolution -- and by training it with hierarchical supervision.

To maintain high-resolution representations through the encoding process, 
\citet{SunX19HighResolutionReprLabelingPixelsRegions} and \citet{YuanECCV20ObjectContextualReprSemSegm}  consider using HRNet~\citep{SunCVPR19DeepHighResolutionReprHumanPoseEstimation} as backbone instead of ResNet or VGG, since it enables connecting the high-to-low resolution convolution streams in parallel.

\begin{figure}[ttt]
\begin{center}
\includegraphics[width=\textwidth]{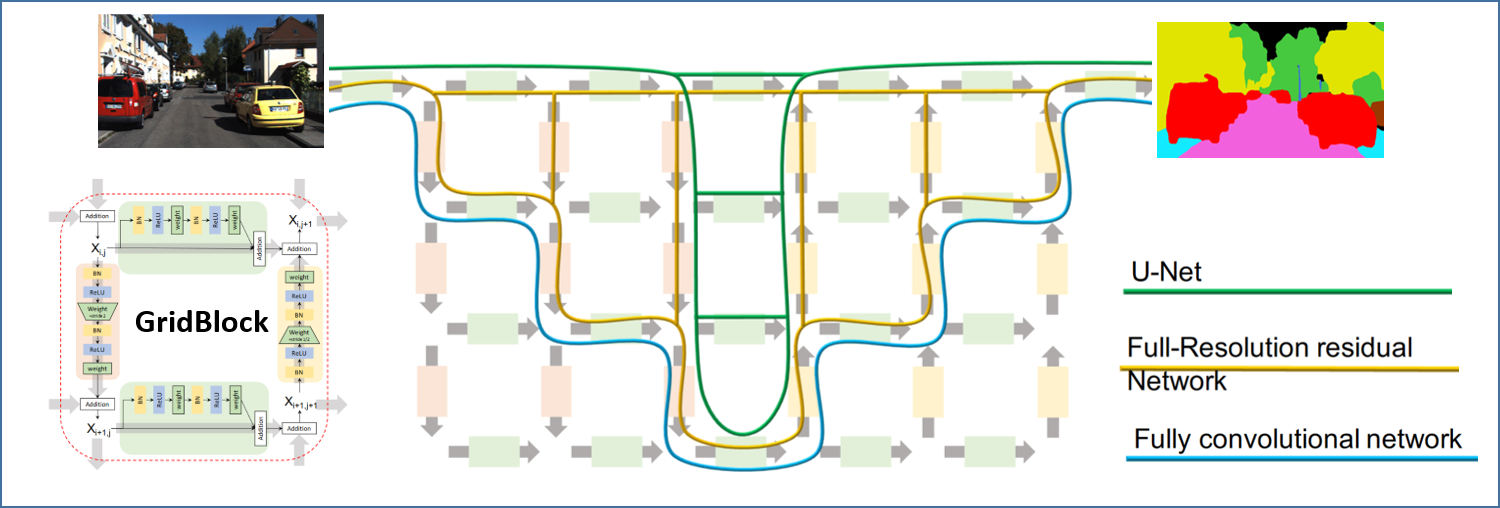}
\caption{\citet{FourureBMVC17ResidualConvDeconvGridNetwork4SiS} propose the GridNet architecture showing that it generalizes several encoder-decoder networks such as DeConvNet~\citep{NohICCV15LearningDeconvolutionNNSegmentation}(blue connections), UNet~\citep{RonnebergerMICCAI15UNetSegmentation} (green connections) or FRRN~\citet{PohlenCVPR17FullResolutionResidualNetworksSegmentation} (yellow connections) (Figure based on \citep{FourureBMVC17ResidualConvDeconvGridNetwork4SiS}) .}
\label{fig:GridNet}
\end{center}
\end{figure}

The GridNet~\citep{FourureBMVC17ResidualConvDeconvGridNetwork4SiS} architecture  follows a grid pattern which is composed of multiple paths called {\it streams} from the input image to the output prediction. The streams are interconnected with convolutional and deconvolutional units, where information from low and high resolutions can be shared. The two-dimensional grid structure allows information to flow horizontally in a residual resolution-preserving way or vertically through down- and up-sampling layers. The authors show that this architecture generalizes several encoder-decoder networks such as DeConvNet~\citep{NohICCV15LearningDeconvolutionNNSegmentation}, UNet~\citep{RonnebergerMICCAI15UNetSegmentation} or FRRN~\citet{PohlenCVPR17FullResolutionResidualNetworksSegmentation} (see \fig{GridNet}).

\subsection{Recurrent neural networks}
\label{sec:ReccNetSiS}

Another group of methods consider using recurrent neural network (RNN) instead of CNNs \citep{PinheiroICML14RecurrentConvolutionalNN4SceneParsing,GattaCVPRWS14UnrollingLoopyTopDownSemanticFeedbackCNN}; they show that modeling the long distance dependencies among pixels is beneficial to improve the segmentation quality. \citet{PinheiroICML14RecurrentConvolutionalNN4SceneParsing} is the first to use recurrent network for SiS exploiting the fact that RNN allows to consider a large input context with limited capacity of the model. Their model is trained with different input patch sizes (the instances) recurrently to learn increasingly large contexts for each pixel, whilst ensuring that the larger context is coherent with the smaller ones. 
Similarly,~\citet{GattaCVPRWS14UnrollingLoopyTopDownSemanticFeedbackCNN} propose  unrolled CNNs through different time steps to include semantic feedback connections. However, in contrast to classical RNNs, the architecture is replicated without sharing the weights and each network is fed with the posterior probabilities generated by the previous softmax classifier. Local and global features are learned in an unsupervised manner and combined.

\begin{figure}[ttt]
\begin{center}
\includegraphics[width=\textwidth]{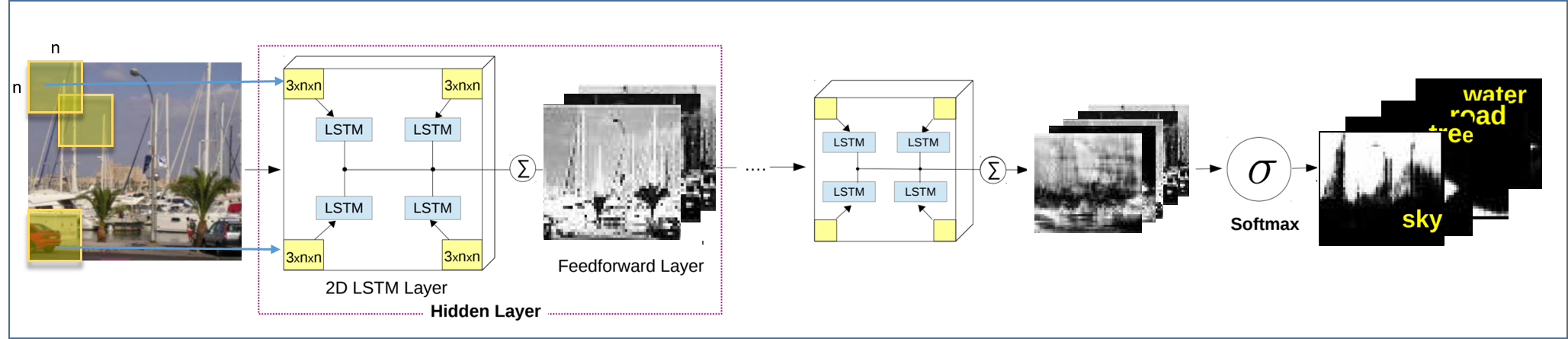}
\caption{In the 2D LSTM model~\citep{ByeonCVPR15SceneLabelingLSTM}, the input image is divided into non-overlapping windows of size $n\times n$ which are are fed into four separate LSTM memory blocks. The current window of LSTM block is connected to its surrounding directions and the output of each LSTM block is then passed to the feed-forward layer where all directions are summed. At the last layer, the outputs of the final LSTM blocks are summed and sent to the softmax layer (Figure based on~\citep{ByeonCVPR15SceneLabelingLSTM}).}
\label{fig:LSTM}
\end{center}
\end{figure}

\citet{VisinCVPRWS16ReSegRNNSemSegm} propose the ReSeg structured prediction architecture which exploits the local generic features extracted by CNN and the capacity of RNN to retrieve distant dependencies. The model is a sequence of ReNet layers composed of four RNNs that sweep the image horizontally and vertically in both directions providing relevant global information and are followed by upsampling layers to recover the original image resolution in the final predictions. ReNet layers are stacked on top of pre-trained convolution layers, benefiting from generic local features.
\citet{ZhengICCV15ConditionalRandomFieldsAsRNN} propose RNNs to perform inference on the CRFs with Gaussian pairwise potentials where a mean-field iteration is modeled as a stack of CNN layers. 

\citet{ByeonCVPR15SceneLabelingLSTM} introduce two-dimensional Long Short Term Memory (LSTM) networks, which consist of 4 LSTM blocks scanning all directions of an image (see \fig{LSTM}). This allows the model to take into account complex spatial dependencies between labels, where each local prediction is implicitly affected by the global contextual information of the image. \citet{LiangECCV16SemanticObjectParsingGraphLSTM} develop the Graph LSTM model, which considers an arbitrary-shaped super-pixel as a semantically consistent node of the graph and spatial relations between the super-pixels as its edges.

\subsection{Pyramidal architectures}
\label{sec:PyramidSiS}

While deep CNNs can capture rich scene information with multi-layer convolutions and nonlinear pooling, local convolutional features have limited receptive fields. 
Different categories may share similar local textures, \eg ``road'' and ``sidewalk'', hence it is important to take into account the context at multiple scales to remove the ambiguity caused by local regions. Therefore several works have been proposed to solve this with pyramidal architectures, which furthermore help to obtain more precise segmentation boundaries. Amongst them, the work by ~\citet{FarabetPAMI13LearningHierarchicalFeaturesSceneLabeling} who transform the input image through a Laplacian pyramid where different scale inputs are fed into a pyramid of CNNs and the feature maps obtained from different scales are then combined. \citet{GhiasiECCV16LaplacianPyramidReconstructionRefinementSemSegm} develop a multi-resolution reconstruction architecture based on a Laplacian pyramid that uses skip connections from higher resolution feature maps and multiplicative confidence-weighted gating to successively refine segment boundaries reconstructed from lower-resolution maps.

The main idea behind RefineNet~\citep{LinCVPR17RefineNetMultiPathRefinementNetworks} is similarly to refine coarse resolution predictions with finer-grained ones in a recursive manner. This is achieved by short-range and long-range residual connections with identity mappings which enable effective end-to-end training of the whole system. Furthermore a chained residual pooling allows the network to capture background context from large image regions.

\begin{figure}[ttt]
\begin{center}
\includegraphics[width=0.9\textwidth]{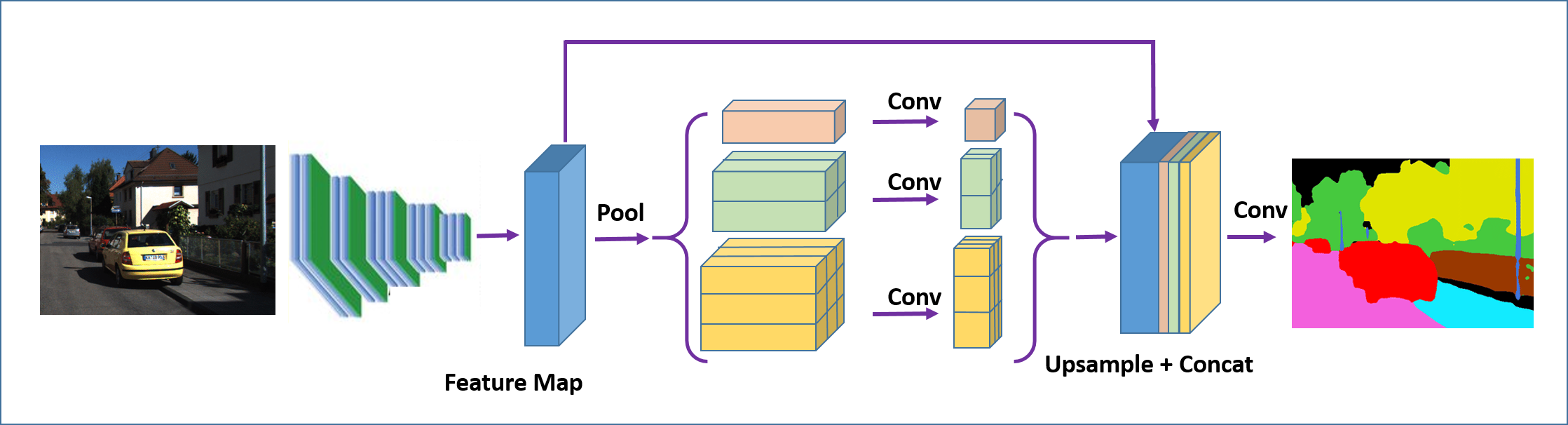}
\caption{The PSPNet~\citep{ZhaoCVPR17PyramidSceneParsingNetwork} gets the feature map of the last convolution layer of the encoder on which  a pyramid parsing module is  applied to harvest different sub-region representations. These representations are further upsampled and concatenated to the initial feature map to form the final feature representation. In this way local and global clues are fused together to make the final prediction more reliable (Figure based on~\citep{ZhaoCVPR17PyramidSceneParsingNetwork}.)}
\label{fig:PSPNet}
\end{center}
\end{figure}

The pyramid scene parsing network (PSPNet)~\citep{ZhaoCVPR17PyramidSceneParsingNetwork} extends Spatial Pyramid Pooling (SPP), proposed by~\citet{HeECCV14SpatialPyramidPoolingCNNVisRec}, to semantic segmentation. The pyramid parsing module is applied to harvest different sub-region representations, followed by up-sampling and concatenation layers to form the final feature representation, which -- carrying both local and global context information -- is fed into a convolution layer to get the final per-pixel prediction (see~\fig{PSPNet}). 
 
\citet{XiaoECCV18UnifiedPerceptualParsingSceneUnderstanding}
propose a Unified Perceptual Parsing framework (UperNet) which combines the Feature Pyramid Network (FPN) \citep{LinCVPR17FeaturePyramidNetworksObjDet} with a Pyramid Pooling Module (PPM)~\citep{ZhaoCVPR17PyramidSceneParsingNetwork}. The model is trained in a multi-task manner with image-level (scenes, textures) and pixel-level (objects, object parts, materials) annotations.

\subsection{Dilated convolutions}
\label{sec:DilatedSiS}

Dilated convolution-based networks~\citep{ChenICLR15SemanticImgSegmFullyConnectedCRFs,YuX15MultiScaleContextAggregationDilatedConvolutions} aggregate multi-scale contextual information where, instead of sub-sampling, dilated convolutions are used as they support exponential expansion of the receptive field without loss of resolution nor coverage. 

Many recent SiS methods adopt the dilated convolutions. For example, \citet{PaszkeX16ENetDeepNNForRealTimeSemSegm} extend SegNet~\citep{BadrinarayananPAMI17SegnetDeepConvEncoderDecoder} with dilated convolutions and make it asymmetric, using a large encoder and a small decoder. The encoder-decoder network of~\citet{WangWACV18UnderstandingConvolutionSemSegm} uses hybrid dilated convolutions in the encoding phase and dense upsampling convolutions to generate pixel-level prediction. 
\citet{ChenPAMI17DeeplabSemanticImgSegmentationDeepFullyConnectedCRF} propose to perform Spatial Pyramid Pooling with dilated convolutions where parallel atrous convolution layers with different rates capture multi-scale information.~\citet{YangCVPR18DenseASPPSemSegmStreetScenes} densely connect ASSP layers where the output of each dilated convolution layer is concatenated with input feature map and then fed into the next dilated layer. \citet{HeCVPR19AdaptivePyramidContextNetworkSemSegm} introduce multi-scale contextual representations with multiple adaptive context modules, where each of such modules uses a global representation to guide the local affinity estimation for each sub-region. The model then concatenates context vectors from different scales with the original features for predicting the semantic labels of the input pixels.

\begin{figure}[ttt]
\begin{center}
\includegraphics[width=\textwidth]{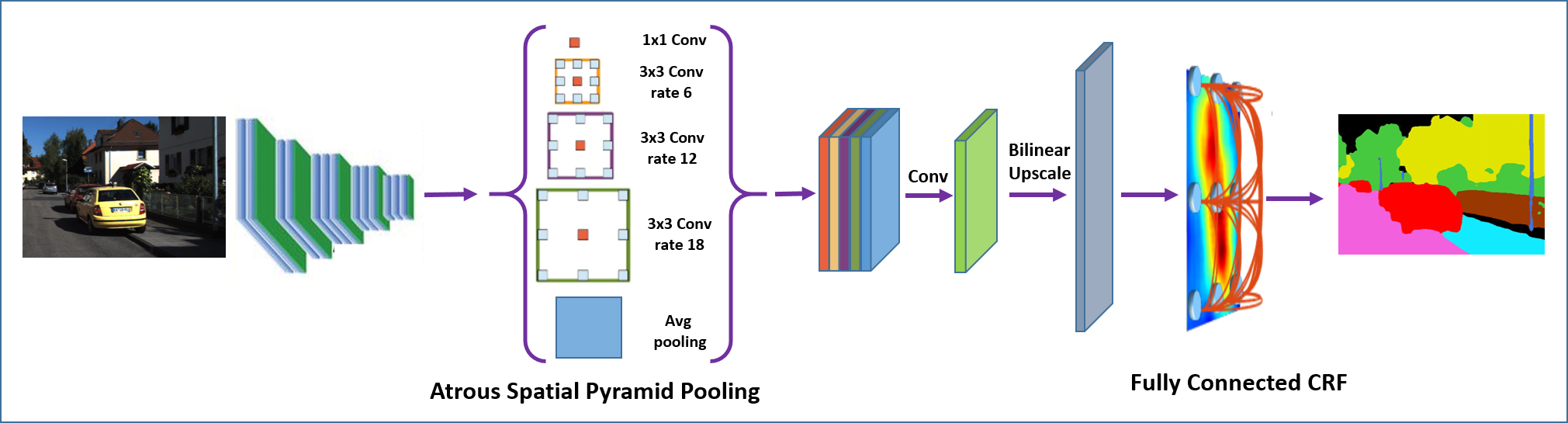}
\caption{The DeepLab model~\citep{ChenPAMI17DeeplabSemanticImgSegmentationDeepFullyConnectedCRF} relies on a deep CNN with atrous convolutions to reduce the degree of signal downsampling and a bilinear interpolation stage that enlarges the feature maps to the original image resolution. A fully connected CRF is finally applied to refine the segmentation result and to better capture the object boundaries (Figure based on~\citep{ChenPAMI17DeeplabSemanticImgSegmentationDeepFullyConnectedCRF}).}
\label{fig:DeepLab}
\end{center}
\end{figure}

The architecture of the popular DeepLab family~\citep{ChenPAMI17DeeplabSemanticImgSegmentationDeepFullyConnectedCRF,ChenX17RethinkingAtrousConvolutionSemSegm,ChenICLR15SemanticImgSegmFullyConnectedCRFs}  combines several ingredients including dilated convolution to address the decreasing resolution, ASPP to capture objects as well as image context at multiple scales, and CRFs to improve the segmentation boundaries (see \fig{DeepLab}).~\citet{ChenECCV18EncoderDecoderAtrousSeparableConvSemSegm} use the DeepLabv3~\citep{ChenX17RethinkingAtrousConvolutionSemSegm} framework as encoder in an encoder-decoder architecture.

\subsection{Attention mechanism}
\label{sec:AttBasedSiS}

Attention and self-attention mechanism is widely used for many visual tasks. Amongst the methods for SiS, we can mention the work of~\citet{ChenCVPR16AttentionToScaleScalAwarSemSegm} who propose a simple attention mechanism that weighs multi-scale features at each pixel location. These spatial attention weights reflect the importance of a feature at a given position and scale. 

\citet{LiBMVC18PyramidAttentionNetworkSemSegm} propose a Pyramid Attention Network (PAN) where Feature Pyramid Attention modules are used to embed context features from different scales and Global Attention Upsample  modules on each decoder layer to provide global context as a guidance during global average pooling to select category localization details. 

\begin{figure}[ttt]
\begin{center}
\includegraphics[width=\textwidth]{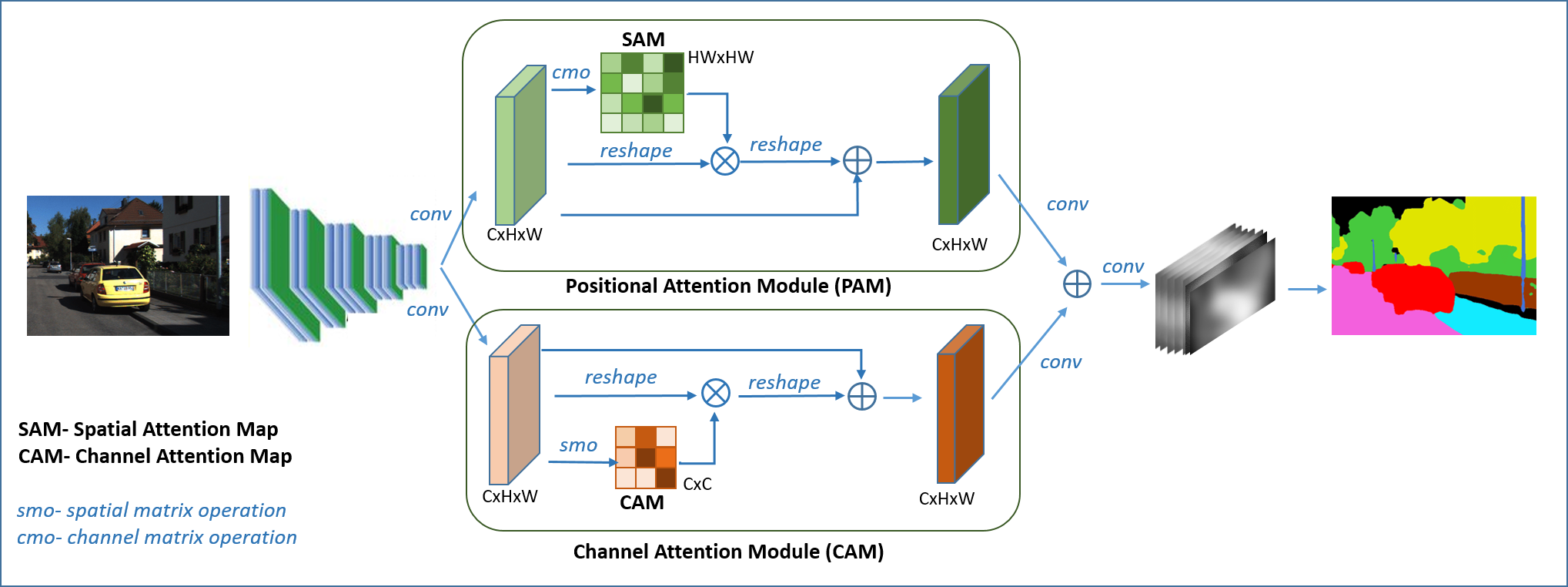}
\caption{The Dual Attention Networks~\citep{FuCVPR19DualAttentionNetworkSceneSegm} aggregates  the output of the Position Attention Module (PAM) that aims at capturing the spatial dependencies between any two positions of the feature maps with the channel attention module (CAM) exploiting the inter-dependencies between channel maps. Specifically, the outputs of the two attention modules are transformed by a convolution layer before fused by an element-wise sum followed by a convolution layer to generate the final prediction maps (Figure based on~\citep{FuCVPR19DualAttentionNetworkSceneSegm}).}
\label{fig:DANet}
\end{center}
\end{figure}

\citet{FuCVPR19DualAttentionNetworkSceneSegm} introduce Dual Attention Networks (DANs) to adaptively integrate local features with their global dependencies. This is achieved by two self-attention mechanisms, the Position Attention Module (PAM) capturing the spatial dependencies between any two positions of the feature maps, and the Channel Attention Module (CAM) that exploits dependencies between channel maps. The outputs of the two attention modules are fused to enhance the feature representations (see illustration in~\fig{DANet}).

To aggregate long-range contextual information in a flexible and adaptive manner,  \citet{ZhaoECCV18PSANetPointWiseSpatialAttentionNetworkSceneParsing} propose the Point-wise Spatial Attention Network (PSANet) where each position in the feature map is connected with all the other ones learning self-adaptive attention masks  which are sensitive to location and category information. The contextual information collected with a bi-directional information propagation path is fused with local features to form the final representation of complex scenes. 

\citet{WangECCV20AxialDeepLabStandAloneAxialAttPanopticSegm} propose an axial-attention block (AAL) which factorizes 2D self-attention into two 1D ones. It consists of two axial-attention layers operating along height-axis and width-axis sequentially. 

\subsection{Transformer-based models}
\label{sec:tranfSiS}

\begin{figure}[ttt]
\begin{center}
\includegraphics[width=\textwidth]{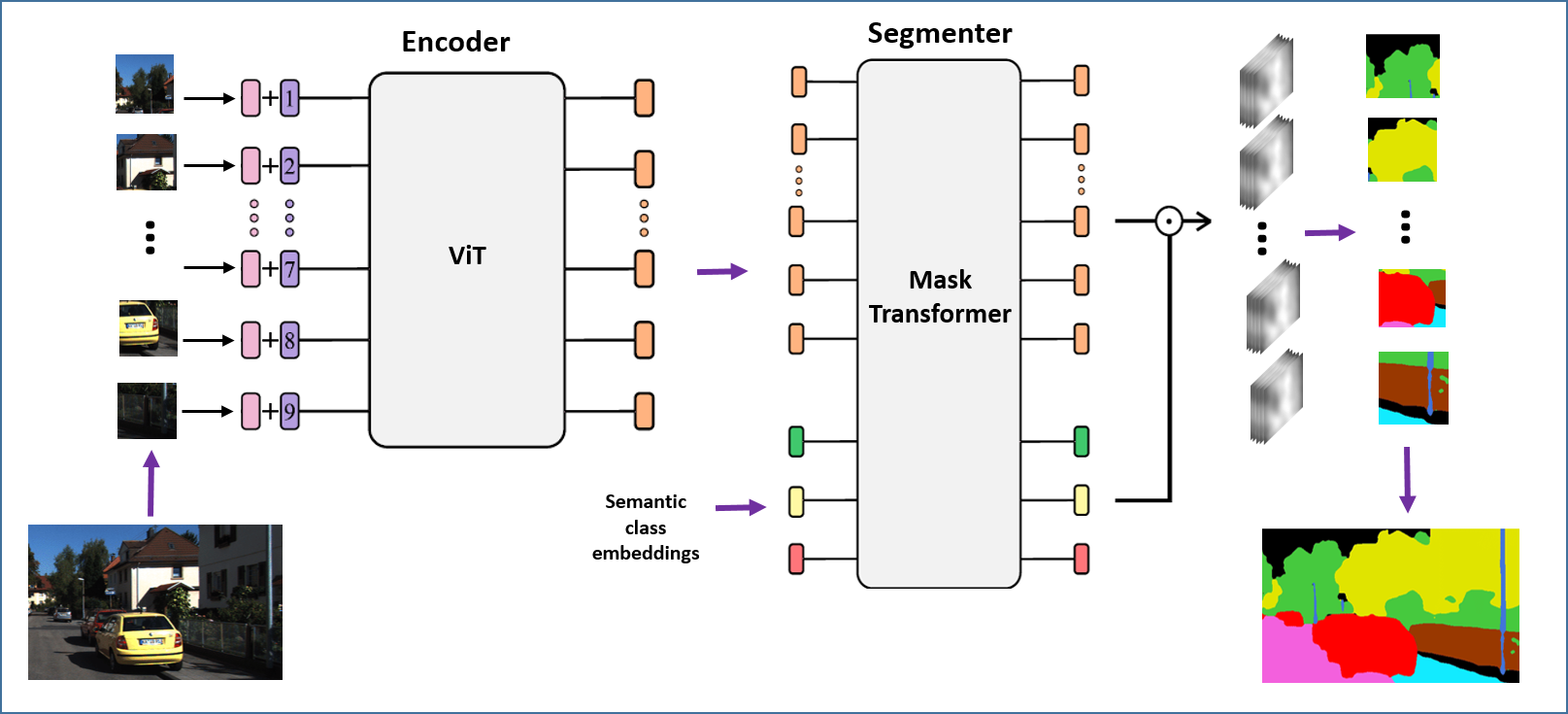}
\caption{The Segmenter model~\citep{StrudelICCV21SegmenterTransformerForSemSegm} 
projects image patches to a sequence of embeddings encoded with the Vision Transformer (ViT)~\citep{DosovitskiyICLR21AnImageIsWorth16x16WordsTransformersAtScale} and the Mask Transformer takes the output of the encoder and semantic class embeddings to predict segmentation class masks for each image patch; these masks are then combined to get the full semantic prediction map (Figure based on \citep{StrudelICCV21SegmenterTransformerForSemSegm}).}
\label{fig:Segmenter}
\end{center}
\end{figure}

Transformer-based models belong to the most recent networks that rely on self-attention, aimed to capture global image context and to address the segmentation ambiguity at the level of image patches. Amongst the first,~\citet{StrudelICCV21SegmenterTransformerForSemSegm} extend the recent Vision Transformer (ViT) model~\citep{DosovitskiyICLR21AnImageIsWorth16x16WordsTransformersAtScale} to handle semantic segmentation problems. In contrast to convolution-based approaches, ViT allows to model global context starting from the first layer and throughout all the network. The model relies on the output embeddings corresponding to image patches and obtains class labels from these embeddings with a point-wise linear decoder or a mask transformer decoder (see~\fig{Segmenter}). 

The pyramid Vision Transformer (PVT)~\citep{WangICCV21PyramidVisionTransformerDensePredWithoutConvolutions} is another extension of ViT where 
incorporating the pyramid structure from CNNs allows the model to better handle dense predictions. 
\citet{RanftlICCV21VisionTransformers4DensePrediction} introduce a transformer for dense prediction, including depth estimation and semantic segmentation. The model takes region-wise output of convolutional networks augmented with positional embedding, assembles tokens from various stages of the ViT into image-like representations at various resolutions, and progressively combines them into full-resolution predictions using a convolutional decoder.

\citet{XieNIPS21SegFormerSemSegmTransformers} combine hierarchical transformer-based encoders to extract coarse and fine features with lightweight multi-layer perceptron decoders to aggregate information from different layers, thus combining both local and global attentions to render a more powerful representation. \citet{GuoICCV21SOTRSegmentingObjectsWithTransformers} also follow the hierarchical approach; they use Feature Pyramid Network (FPN) to generate multi-scale feature maps, which are then fed into a transformer -- to acquire global dependencies and to predict per-instance category -- and into a multi-level upsampling module to dynamically generate segmentation masks guided by the transformer output.

\citet{LiuICCV21SwinTransformerHierarchicalViTShiftedWindows} introduce the Swin Transformer for constructing hierarchical feature maps and promote it as a general-purpose backbone for major downstream vision tasks. The key idea is a hierarchy of shifted window based multi-headed self-attention layers,  where each layer contains a Swin Attention Block (SwT) followed by a SeMask Attention Blocks (SAB)  to capture the semantic context in the encoder network. The Semantic-FPN like decoder ensures the connections between windows of consecutive layers.  

\citet{JainX21SeMaskSemanticallyMaskedTransformersSiS} incorporate semantic information into the encoder with the help of a semantic attention operation. This is achieved by adding Semantic Layers composed of SeMask Attention Blocks after the Swin Transformer Layer to capture the semantic context in a hierarchical encoder network. At each stage, the semantic maps, decoded from the Semantic Layers, are aggregated and passed through a weighted cross entropy loss to supervise the semantic context.

\citet{ChuNIPS21TwinsRevisitingDesignSpatialAttVisTransformers} propose an architecture with two Twin Transformers; the first one, called Twins-PCPVT, replaces the positional encoding in PVT by positional encodings generated by a position generator. The second one, called Twins-SVT,  interleaves locally-grouped attention (LGA) -- capturing the fine-grained and short-distance information -- with global sub-sampled attention (GSA), which deals with the long-distance and global information.

\citet{AliNIPS21XCiTCrossCovImageTransformers} propose a novel way of computing self-attention, where attention matrices are computed over feature channels rather than on input tokens. The resulting Cross-covariance Image Transformer (XCiT) model, hence, has the intriguing property of scaling linearly with respect to the input size -- in terms of computation and memory. This allows training on higher-resolution images and/or smaller patches.

\subsection{SiS specific losses}
\label{sec:SiS_losses}

The most common loss used in deep SiS methods is the standard~\textit{pixel-wise cross-entropy} loss, where the aim is to minimize the difference between the class predictions and the ground-truth annotations for all pixels: 
\begin{align*}
\Loss_{ce}  =  - \mathbb{E}_{
(\calX, \calY)}
 \left[ \sum_{h,w} {\bfy^{(h,w)} \cdot \log(p(F(\bfx^{(h,w)})))} \right],
\end{align*}
where $F$ is the segmentation model, $p(F(\bfx^{(h,w)}))$ is a vector of class probabilities at pixel $\bfx^{(h,w)}$ and $\bfy^{(h,w)}$ is a one-hot vector with 1 at the position 
of the pixel's true class and 0 elsewhere.

As SiS data is compiled in real world environments, most of them are often imbalanced, with dominant portions of data assigned to a few majority classes while the rest belonging to minority classes, thus forming under-represented categories. As consequence, deep SiS methods trained with the conventional cross-entropy loss tend to be biased towards the majority classes during inference~\citep{RahmanISVC17OptimizingIntersectionOverUnionDeepNNSiS,BuloCVPR17LossMaxPoolingSiS}.

To mitigate this class-imbalance problem, SiS datasets can undergo the {\it resampling} step, by over-sampling minority classes and/or under-sampling majority classes. However, such approaches change the underlying data distributions and may result in sub-optimal exploitation of available data, increasing the risk of over-fitting when repeatedly visiting the same samples from minority classes.

An alternative to resampling is the {\it cost-sensitive} learning procedure, which introduces class-specific weights, often derived from the original data statistics. They use statically-defined cost matrices~\citep{Caesar2015JointCalibration,Mostajabi2015FeedforwardSemanticSegmentation} or introduce additional parameter learning steps~\citep{Khan2018CostSensitiveLearning}. Due to the spatial arrangement and strong correlations of classes between adjacent pixels, cost-sensitive learning techniques outperform resampling methods. However, the increasing complexity and a large number of minority in the recent SiS datasets make cumbersome the accurate estimation of the cost matrices.

Some approaches to the class-imbalance problem take in account the SiS specificity. One such method introduces a generalized max-pooling operator acting at the pixel-loss level~\citep{BuloCVPR17LossMaxPoolingSiS}. It provides an adaptive re-weighting of contributions of each pixel, based on the loss they actually exhibit. Image pixels that incur higher losses during training are weighted more than pixels with a lower loss, thus indirectly compensating potential inter-class and intra-class imbalances within the dataset. 

Another approach is to revise the standard cross-entropy loss, which optimizes the network for overall accuracy, and to address the intersection-over-union (IoU) measure instead. As described in Section~\ref{sec:evaluate}, the IoU measure gives the similarity between the prediction and the ground-truth for every segment present in the image; it is defined as the intersection over the union of the labeled segments, averaged over all classes. Methods that optimize the IoU measure proceed either by direct optimization~\citep{RahmanISVC17OptimizingIntersectionOverUnionDeepNNSiS} or by deploying the convex surrogates 
of sub-modular losses~\citep{BermanCVPR18TheLovaszSoftmaxLossTractableSurrogateIoU}.

Another group of SiS specific losses is driven by the observation that segmentation prediction errors are more likely to occur near the segmentation boundaries. \citet{Borse2021InverseForm} introduce a boundary distance-based measure and include it into the standard segmentation loss. They use an inverse transformation network to model the distance between boundary maps, which can learn the degree of parametric transformations between local spatial regions.

\section{Beyond Classical SiS}
\label{sec:beyond_sis}

In this section, we present SiS approaches that go beyond the classical methods described above. A substantial part is focused on solutions that address the shortage of pixel-wise image annotation: in \secref{SSLearn} we detail methods which exploit unlabeled samples; in \secref{WSSiS} we describe approaches that rely on weak labels such as image-level annotations or bounding boxes, instead of ground-truth segmentation maps. Furthermore, \secref{currLearn} considers the case when the training is decomposed into easier-to-harder tasks learned sequentially; \secref{class-inc} reviews methods where the model's underlying knowledge is incrementally extended to new classes; finally, \secref{SelfLearn} focuses on the effects of self-supervised visual pre-training on SiS.

\subsection{Semi-supervised SiS}
\label{sec:SSLearn}

In conventional semi-supervised learning (SSL), to overcome the burden of costly annotations,  the model makes usage of a small number of labeled images and a large number of unlabeled ones. In the case of SiS, most semi-supervised methods exploit a small set of labeled images with pixel-level annotations and
s set of images with image-level annotation, like image class labels or object bounding boxes. 
Below we present semi-supervised extension of SiS models presented in Section~\ref{sec:deepSS}. 

Amongst the early semi-supervised SiS works, we first mention~\citep{HongNIPS15DecoupledDeepNNSegmentation}, that proposes an encoder-decoder framework where 
image-level annotations are used to train the encoder and pixel-wise ones are used to train the decoder.~\citet{OquabCVPR15IsObjectLocalizationForFree} rely on FCN and introduce a max-pooling layer that hypothesizes the possible location of an object in the image.~\citet{PathakICCV15ConstrainedCNNWeaklySegmentation} propose a constrained CNN where a set of linear constraints are optimized to enforce the model's output to follow a distribution over latent “ground-truth” labels as closely as possible.
\citet{PapandreouICCV15WeaklySemiSupervisedCNNSegmentation} develop an Expectation-Maximization (EM) method for training CNN semantic segmentation models under weakly and semi-supervised settings. The algorithm alternates between estimating the latent pixel labels, subject to the weak annotation constraints, and optimizing the CNN parameters using stochastic gradient descent (SGD).~\citet{HongCVPR16LearningTransferrableKnowledgeSemSegm} combine the encoder-decoder architecture with an attention model and exploit auxiliary segmentation maps available for different categories together with the image-level class labels. 

Another principle of semi-supervised learning, consistency regularization by data augmentation, has been also successfully applied to SiS~\citep{FrenchBMVC19SemiSupervisedSemSegmHighDimPerturbations,FrenchBMVC20SemiSupSiSNeedsStrongVariedPerturbations,OualiCVPR20SemiSupervisedSemSegmCrossConsistencytraining,LuoECCV20SemiSupervisedSemSegmStrongWeakDualBranchNetwork,OlssonWACV21ClassMixSegmentationBasedDataSugmentation4SemiSupLearning,ChenCVPR21SemiSupervisedSISCrossPseudoSupervision,YuanICCV21ASimpleBaselineSemiSupSemSegmStrongDataAugmentation} and extended in 
~\citep{LaiCVPR21SemiSupSiSDirectionalContextAwareConsistency} towards a directional context-aware consistency between pixels under different environments. 

To further improve 
the consistency regularization methods, contrastive learning is used 
1) by~\cite{ZhouICCV21C3SemiSegContrastiveSemiSupSegmCrossSetLearningDynamicClassBalancing} to decrease inter-class feature discrepancy and increase inter-class feature compactness across the dataset, 
2) by~\cite{ZhongICCV21PixelContrastiveConsistentSemiSupSiS} to simultaneously enforce the consistency in the label space and the contrastive property in the feature space, and 
3) by~\citet{AlonsoICCV21SemiSupeSiSPixelLevelContrastiveClassWiseMemoryBank} to align class-wise and per-pixel features from both labeled and unlabeled data stored in a memory bank.

\cite{YangCVPR22STppMakeSelfTrainingWorkBetterSemiSupSiS} show that re-training by injecting strong data augmentations on unlabeled images allows constructing strong baselines, but such strong augmentations might yield incorrect pseudo labels. To avoid the potential performance degradation incurred by incorrect pseudo labels, they perform selective re-training via prioritizing reliable unlabeled images based on holistic prediction-level stability in the entire training course.

\citet{HeCVPR21ReDistributingBiasedPseudoLabelsSemiSupSiSBaselineInvestigation} observe that semi-supervised SiS methods in the wild severely suffer from the long-tailed class distribution and  propose a \textit{distribution alignment} and \textit{random sampling} method to produce unbiased pseudo labels that match the true class distribution estimated from the labeled data. Similarly, to cope with long-tailed label distribution,~\citet{HuNIPS21SemiSupSiSViaAdaptiveEqualizationLearning} propose an adaptive equalization learning framework that adaptively balances the training of well and badly performed categories, with a confidence bank to dynamically track category-wise performance during training.

Finally, several methods 
use Generative Adversarial Networks (GANs)~\cite{GoodfellowNIPS14GenerativeAdversarialNets} to train a discriminator able to distinguish between confidence maps from labeled and unlabeled data predictions~\citep{HungBMVC18AdversarialLearning4SemiSupervisedSemSegm}, to refine low-level errors in the predictions through a discriminator that classifies between generated and ground-truth segmentation maps~\citep{MittalPAMI21SemiSupSemSegmHighLowLevelConsistency}, or 
to generate fake visual data forcing the discriminator to learn better features~\citep{SoulyICCV17SemiSupSemSegmGAN}.

\begin{figure}[ttt]
\begin{center}
\includegraphics[width=0.85\textwidth]{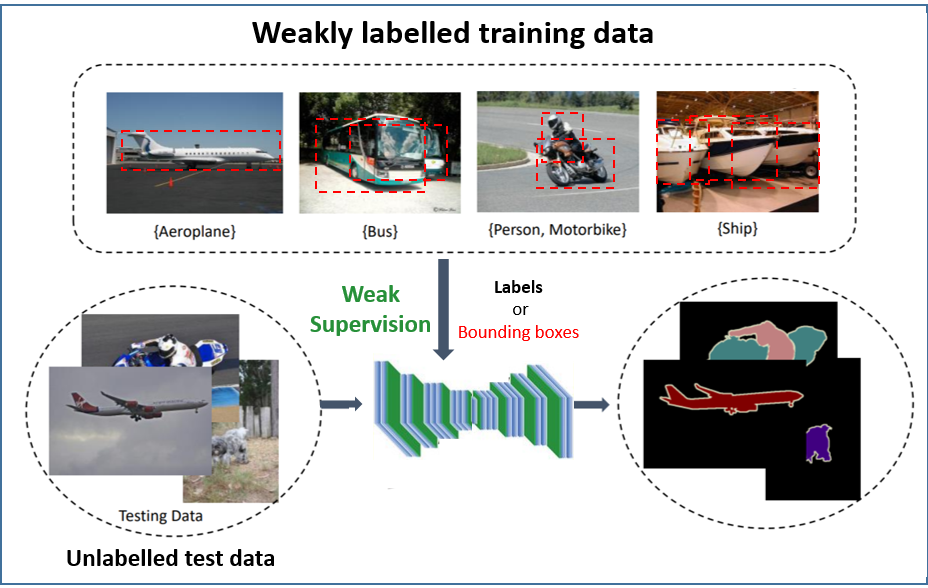}
\caption{Weakly-supervised SiS aims at using either the image-level or the bounding box annotations as supervision to learn a pixel-level image segmentation (Image courtesy of Xinggang Wang).}
\label{fig:WSSiS}
\end{center}
\end{figure}

Learning from partially labeled images, where some regions are labeled and others not, is a particular case of semi-supervised segmentation~\citep{VerbeekNIPS07SceneSegmCRFfromPartiallyLabeledImages,HeNIPS09LearningHybridModelsPartiallyLabeledData}.

\subsection{Weakly-supervised SiS}
\label{sec:WSSiS}

In contrast to semi-supervised learning, weakly-supervised SiS~\citep{BorensteinECCV04LearningToSegment} relies only on weak annotations such as image captions, bounding box or scrible annotations (see example in \fig{WSSiS}). 

Early methods using only image-level annotations in general rely on multiple instance learning where each image is viewed as a bag of patches or super-pixels, and the final prediction is accomplished by aggregation of the class predictions on these patches or super-pixels~\citep{GalleguillosECCV08WeaklySupObjLocalizationWithStableSegm,VezhnevetsCVPR10TowardsWeaklySupSiSMIL}. In contrast, CNN based weakly-supervised SiS models exploit the observation that CNNs have remarkable localization ability despite being trained on image-level labels~\citep{ZhouCVPR16LearningDeepFeaturesDiscriminativeLocalization}. These~\textit{Classification Activation Maps} (CAM) allow to select discriminative regions for each semantic class that can be used as pixel-level supervision for segmentation networks. 

To improve such initial CAMs, AffinityNet \citep{AhnCVPR18LearningPixelLevelSemanticAffinityWSSS} learns from data how to propagate local activations by performing a random walk to the entire object area, predicting semantic affinity between a pair of adjacent image coordinates. The seed-expand-constrain (SEC) model~\citep{KolesnikovECCV16SeedExpandConstrainWeaklySupervisedImgSegm} \textit{seeds} weak localization cues,  \textit{expands} them with image-level class predictions and  \textit{constrains} with a CRF the segmentation to coincide with object boundaries. A similar framework is used by~\citet{HuangCVPR18WeaklySupSiSDeepSeededRegionGrowing}, except that the region expansion to cover the whole objects is done with the Seeded Region Growing algorithm~\citep{AdamsPAMI94SeededRegionGrowing}. The method was extended by~\citet{LeeCVPR19FickleNetWeaklySemiSupeSiSStochasticInference} where, instead of CAM, they rely on Grad-CAM~\citep{SelvarajuICCV17GradCAMVisualExplanationsDeepNetworksViaGradientBasedLocalization} to generate and combine a variety of localization maps obtained with random combinations of hidden units.
\citet{RedondoCabreraTIP18Learning2Exploit2PriorNetworkKnowledgeWeaklySupSiS} combine two Siamese CAM modules to get activation masks that cover full objects and a segmenter network which learns to segment the images according to these activation maps. 

\citet{RoyCVPR17CombiningBottomUpTopDownSmoothnessCuesWSSiS} propose to train a CRF-RNN~\citep{ZhengICCV15ConditionalRandomFieldsAsRNN} where, for each object class, bottom-up segmentation maps -- obtained from the coarse heatmaps -- are combined with top-down attention maps and, to improve the object boundaries,  refined in the CRF-RNN over iterations.~\citet{WangCVPR18WeaklySupervisedSiSIterativelyMiningCommonObjectFeatures} mine common object features from the initial rough localizations and expand object regions with the mined features. To supplement non-discriminative regions, saliency maps are then considered to refine the object regions.

\citet{KwakAAAI17WeaklySupSiSSuperpixelPoolingNetwork} propose a Super-pixel Pooling Network, which utilizes super-pixel segmentation as a pooling layout to reflect low-level image structure, and use them  within deCoupledNet~\citep{HongNIPS15DecoupledDeepNNSegmentation} to learn semantic segmentation. The WILDCAT model~\citep{DurandCVPR17WILDCATDeepConvNets4WSSiS} is based on FCN, where all regions are encoded into multiple class modalities with a  multi-map transfer layer, and pooled separately for each classes to obtain class-specific heatmaps.~\citet{SunECCV20MiningCrossImageSemantics4WSSIS} propose two complementary neural co-attention models to capture the shared and unshared objects in paired training images.

Several methods consider adding a separate localization branch that performs the object detection and thus helps adjusting the output of the segmentation branch. In this spirit, \citet{QiECCV16AugmentedFeedbackSiSImageLevelSupervision} select positive and negative proposals from the predicted segmentation maps for training the object localization branch and uses an aggregated proposal to build pseudo labeled segmentation to train the segmentation branch. 

Another group of weakly-supervised SiS methods considers that the object bounding boxes in an image are available and obtained either manually, as much less costly than pixel-level annotation, or automatically by pretrained object detectors such as R-CNNs~\citep{GirshickCVPR14RichFeatureHierarchies}. As such,~\citet{XiaCVPR13SemanticSegmentationWithoutAnnotatingSegments} introduce a simple voting scheme to estimate the object's shape in  each bounding box  using a subsequent graph-cut-based figure-ground segmentation. Then, they aggregate the segmentation results in the bounding boxes to obtain the final segmentation result.

\citet{DaiICCV15BoxSupExploitingBoundingBoxesSuperviseSiS} iterate between 1) automatically generating segmentation masks and 2) training an FCN segmentation network under the supervision of these approximate masks. The segmentation masks are obtained with \textit{multi-scale combinatorial grouping} (MCG)  \citep{PontTusetPAMI16MultiscaleCombinatorialGroupingImSegmObjProposalGeneration} of unsupervised region proposals~\citep{ArbelaezPAMI11ContourDetectionHierarchicalSegmentation}.
A similar approach has been proposed by~\citet{KhorevaCVPR17SimpleDoesItWeakSupInstSemSegm} who combine MCG with a modified GrabCut~\citep{RotherTOG04GrabCutInteractiveForegroundExtractionIteratedGraphCuts} to train a DeepLab model.~\citet{JiBMVC21WeaklySupSiSBox2TagAndBack} train a per-class CNN using the bounding box annotations to learn the object appearance and to segment these bounding boxes into object-class versus background labels. These bounding box segments are then combined to get pseudo-labeled image segmentations which can be used to train a DeepLab model.

Instead, \citet{SongCVPR19BoxDrivenClassWiseRegionMaskingFilling} train an FCN model with a box-driven class-wise masking  model to generate class-aware masks, and rely on the mean filling rates of each class as prior cues. \citet{KulhariaECCV20Box2SegAttentionWeightedLossWSSiS} learn pixel embeddings to simultaneously optimize high intra-class feature affinity and increasing discrimination between features across different classes. The model uses per-class attention maps that saliently guides the per-pixel cross entropy loss to focus on foreground pixels and to refine the segmentation boundaries. 

More recent methods rely on the effectiveness of transformer networks (see also Section~\ref{sec:tranfSiS}) to generate high-quality localization for different semantic classes (class aware CAMs) that can be used to generate pseudo labels for supervising the segmentation network. However, as discussed in \citep{GaoICCV21TSCAMTokenSemanticCoupledAttentionMapWSObjLoc} the attention maps of visual transformers are in general semantic-agnostic (not distinguishable to object classes) and therefore are not competent to semantic-aware localization. They propose instead the Token Semantic Coupled Attention Map, that relies on a semantic coupling module which combines the semantic-aware tokens with the semantic-agnostic
attention map. Similarly,~\cite{XuCVPR22MultiClassTokenTransformer4WSSS} exploit class-specific transformer attentions and develop an effective framework to learn class-specific localization maps from the class-to-patch attention of different class tokens. Instead,~\citet{RuCVPR22LearningAffinityFromAttentionEnd2EndWSSSTransformers} propose an \textit{Affinity from Attention} module to learn semantic affinity from the multi-head self-attention and a \textit{Pixel-Adaptive Refinement} of the initial CAM based pseudo labeling via a random walk process.

Amongst other types of weak annotations, we can mention~\textit{scribble} supervision~ \citep{LinCVPR16ScribbleSupSemSegm,VernazaCVPR17LearningRandomWalkLabelPropagationWSSiS,TangECCV18OnRegularizedLossesWSSiS} and the even more constrained \textit{point supervision}~\citep{BearmanECCV16WhatsThePointSiSPointSupervision}, where a single pixel from each class is manually annotated in every image. \citet{XuCVPR15LearningToSegmentUnderVariousFormsOfWeakSupervision} design a unified framework to handle different types of weak supervision (image-level, bounding boxes and scribbles), formulating the problem as a max margin clustering, where supervision comes as additional constraints in the assignments of pixels to class labels.

Crawling the web is another source of weak image supervision.~\citet{JinCVPR17WeblySupervisedSiS} use images with simple background -- crawled from the web -- to train shallow CNNs to predict class-specific segmentation masks, which then are assembled into one deep CNN for end-to-end training.~\citet{ShenBMVC17WeaklySupSiSWebImageCoSegmentation} use a large scale co-segmentation framework to learn an initial dilated FCN segmentation model which is refined using  pseudo-labeled masks and image-level labels of webly crawled images.

\citet{HongCVPR17WeaklySupSemSegmWebCrawledVideo} propose to crawl the web for video sequences and to exploit relevant spatio-temporal volumes within the retrieved videos. In \citep{FanECCV18AssociatingInterImageSalientInstancesWSSiS}, images are fed into a salient instance detector to get proxy ground-truth data and to train DeepLab for segmentation,  and respectively,  Mask-RCNN~\citep{HeICCV17MaskRCNN} for instance segmentation.~\citet{ShenCVPR18BootstrappingThePerformanceWeblySupSiS} rely on two SEC (Seed-Expand-Constrain) models ~\citep{KolesnikovECCV16SeedExpandConstrainWeaklySupervisedImgSegm} where an initial SEC model -- trained on weakly labeled target data -- is used to filter images from the web, a second SEC model learns from these weakly labeled images. Note that many weakly supervised methods rely on Grab-Cut \citep{RotherTOG04GrabCutInteractiveForegroundExtractionIteratedGraphCuts} to improve bounding box binary segmentations, and on CRFs (Section~\ref{sec:globloc_consistency}) to refine the final segmentations. 

\subsection{Curriculum learning based SiS}
\label{sec:currLearn}

Curriculum learning~\citep{BengioICML09CurriculumLearning} refers to the practice where the training process first approaches~\textit{easier tasks} and then progressively solve the~\textit{harder tasks}. 
\citet{SovianyX21CurriculumLearningSurvey} classify the curriculum learning methods into data-level and model-level curriculum learning, where the former group ranks the training samples/tasks from easy to hard and a special module selects which examples/tasks should be considered at a given training step, while the latter group starts with a simpler model and increases progressively the model capacity. Curriculum based SiS methods belong in general to the former group. 

In this setup,~\citet{KumarICCV11LearningSpecificClassSegmDiverseData} propose to use self-paced learning (SPL) algorithm for object segmentation which chooses a set of easy images at each iteration to update the model. \citet{ZhangCVPR17SPFTNSelfPacedFineTuningSegmObjVideo} incorporate the SPL into a  fine-tuning process for object segmentation in videos. The model learns over iterations from easy to complex samples in a self-paced fashion thus allowing the model to cope with data ambiguity and complex scenarios.

\citet{FengX20DMTDynamicMutualTraining4SemiSupervisedLearning} propose an easy-to-hard curriculum self-training approach for semi-supervised SiS where the number of confident pseudo-labels selected from each class is progressively increased where   more difficult (lower confidence) samples are added at later phases.~\citet{JessonMICCAI16CASEDCurriculumAdaptiveSamplingExtremeDataImbalance} combine curriculum learning with hard negative mining for lung nodules segmentation where the model initially learns how to distinguish nodules from their immediate surroundings and then continuously increases the proportion of difficult-to-classify global context.

\subsection{Class-incremental learning for SiS}
\label{sec:class-inc} 

\begin{figure}[ttt]
\begin{center}
\includegraphics[width=\textwidth]{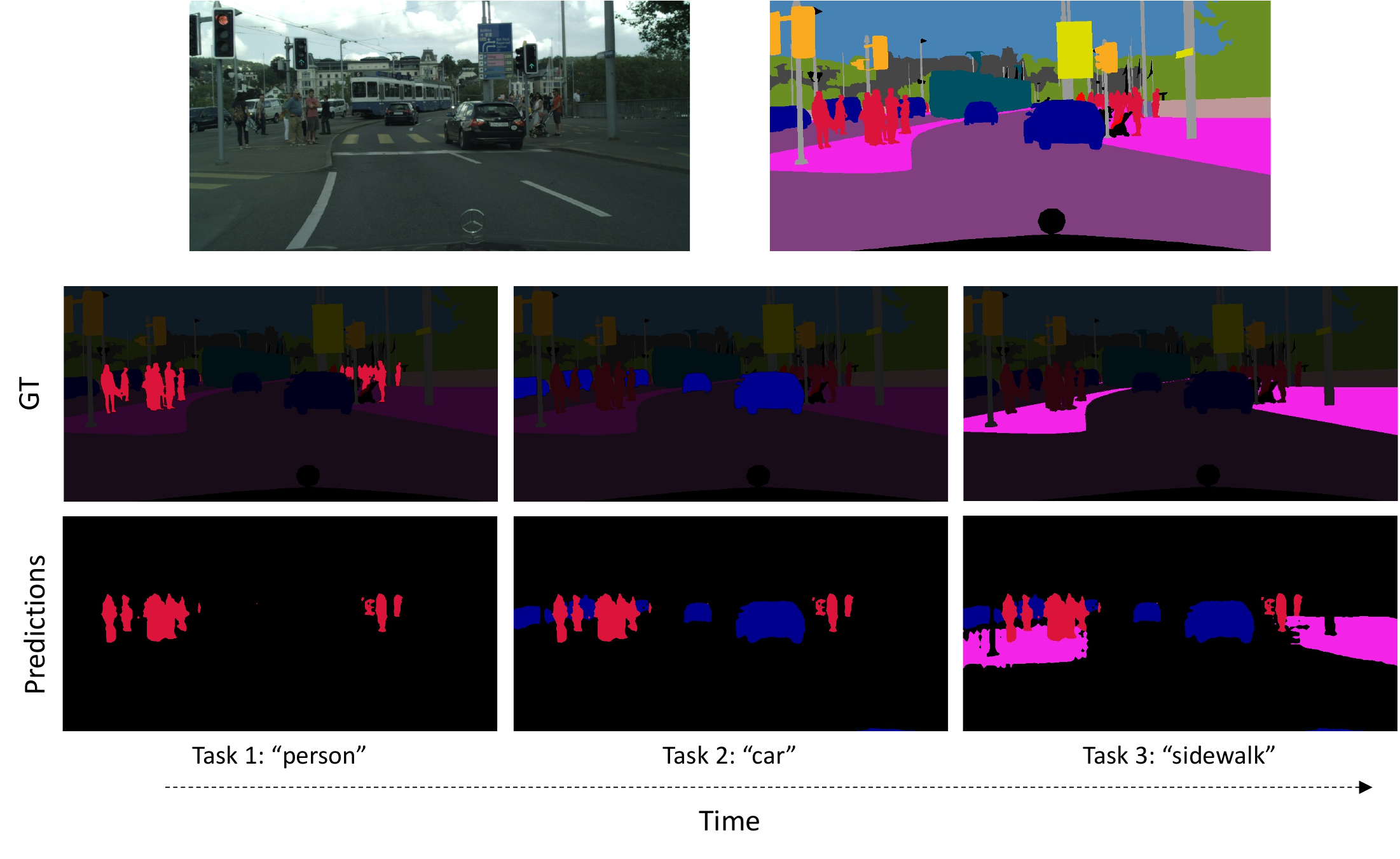}
\caption{Class-incremental learning for SiS. In standard semantic segmentation, models are
trained by relying on datasets where each image is annotated with a mask covering all the classes of interest \textit{(top)}. Instead, in class-incremental learning the goal is to extend the underlying knowledge of a model in a sequence of steps, where at each step only one/a few classes are annotated -- and the rest is ``background''. In the example in \textit{(middle)}, the model is extended with the classes ``person'', ``car'' and  ``sidewalk''  in three different steps. After each stage, the model can predict more classes \textit{(bottom)}.}
\label{fig:PLOP}
\end{center}
\end{figure}

Class-incremental learning is a branch of continual learning where the goal is to extend the underlying knowledge of a model to new classes. In general, the assumption is to have a model trained on an initial class-set 
which at different stages 
is fine-tuned on different new data, 
where images contain annotation for one or more new classes 
(see Figure~\ref{fig:PLOP}). 

The class-incremental learning problem has a long history in image classification; yet, this problem has been only recently addressed in the SIS context~\citep{CermelliCVPR20ModelingBackgroundIncrementalLearningSemSegm,MichieliCVIU21KnowledgeDistillationIncLearnSemSegm, DouillardCVPR21PLOPLearningWithoutForgettingComtSemSegm,MaracaniICCV21RECALLReplayBasedContinualLearningSemSegm,ChaNeurIPS21SSULSemSegmwithUnknownLabelforExamplarClassIncLearning}.
Most works focus on the scenario where the original dataset on which the model was trained is not available, and propose to fine-tune the model on samples available for the current new class~\citep{CermelliCVPR20ModelingBackgroundIncrementalLearningSemSegm,MichieliCVIU21KnowledgeDistillationIncLearnSemSegm,DouillardCVPR21PLOPLearningWithoutForgettingComtSemSegm,MaracaniICCV21RECALLReplayBasedContinualLearningSemSegm} -- without storing samples from the new classes over time. 
This is a realistic scenario under the assumption that the model has been trained by a third-party and therefore one does not have access to the original training set or the access is prohibited by 
the rights to use the original training samples for copyright issues. Recently, in contrast to the above methods,~\citet{ChaNeurIPS21SSULSemSegmwithUnknownLabelforExamplarClassIncLearning} have proposed a memory-based approach to class-incremental SiS.

In class-incremental learning -- and in continual learning in general -- a major challenge that the models need to face is to avoid \textit{catastrophic forgetting}. While learning on new samples --  where only the new classes are annotated (and the rest of the image is considered as background) -- the model may overfit on them and the performance on the previous ones may decrease. In the context of SiS, this is typically referred to as \textit{background shift}~\citep{CermelliCVPR20ModelingBackgroundIncrementalLearningSemSegm}, given that pixel annotation from previously learned classes are in general annotated as ``background'' in the new samples (see Figure~\ref{fig:PLOP}).

\citet{CermelliCVPR20ModelingBackgroundIncrementalLearningSemSegm} 
propose to tackle the background shift via distillation, in order to avoid forgetting previously learned categories. In addition, they propose an initialization strategy, devised \textit{ad hoc}, to mitigate the vulnerability of SiS class-incremental learners to the background shift issue.
\citet{DouillardCVPR21PLOPLearningWithoutForgettingComtSemSegm} mitigate the catastrophic forgetting 
by generating pseudo-labels for old classes where the confidence of the pseudo ground-truth is weighed via its entropy. They further rely on a distillation loss that preserves short- and long-distance relationships between different elements in the scene.

Instead,~\citet{MaracaniICCV21RECALLReplayBasedContinualLearningSemSegm} propose two strategies to recreate data that comprise the old classes; the first one relies on GANs and the second exploits images retrieved from the web. \citet{CermelliCVPR22IncrementalLearninginSemSegmfromImageLabels} propose a class-incremental learning pipeline for SiS where new classes are learned by relying on global image-level labels instead of pixel-level annotations, hence related to weekly supervised SiS  (see Section~\ref{sec:WSSiS}). 

Differently from previous approaches,
\citet{ChaNeurIPS21SSULSemSegmwithUnknownLabelforExamplarClassIncLearning} consider a memory bank in which  a few hundred past samples are stored to mitigate forgetting. Furthermore, they propose to model the ``unknown'' classes other than the ``background'' one, which further helps avoiding forgetting and preparing the model to learn future classes.

Finally, \citet{CermelliBMVC21PrototypeBasedIncFewShotSemSegm} introduce a
new task called Incremental Few-Shot Segmentation (iFSS), where the goal is class-incremental learning by relying on few samples for each new class. They propose a method that can learn from only a few samples while at the same time avoiding catastrophic forgetting. This is done by relying on a prototype-based distillation and on batch renormalization
\citep{IoffeNeurIPS17BatchRenormTowardsReducingMinibatchDependenceinBNModels} to handle~\textit{non-iid} data.

\subsection{Self-supervised SiS}
\label{sec:SelfLearn}

Under the shortage of human annotations, self-supervised learning represents another alternative to learn effective visual representations. The idea is to devise an auxiliary task, such as rotation prediction~\citep{GidarisICLR18UnsupervisedRepresentationLearningPredictImageRotations}, colorization~\citep{ZhangECCV16ColorfulImageColorization}, or contrastive learning~\citep{ChenICML20ASimpleFrameworkContrLearningVisRepr}, and to train a model for this task instead of a supervised one. 

\citet{ZhanAAAI17MixAndMatchTuningSelfSupervisedSiS} are the first to apply self-supervised learning in the SiS context; they propose \textit{Mix-and-Match}, where in the \textit{mix} stage sparsely sampled patches are mixed, and in the \textit{match} stage a class-wise connected graph is used to derive a strong triplet-based discriminative loss for fine-tuning the network. 

Most recently, motivated by the success of BERT~\citep{DevlinNAACL19BERTPreTrainingDeepBidirectionalTransformersLanguageUnderstanding} in NLP and by the introduction of Vision Transformers (ViT)~\citep{DosovitskiyICLR21AnImageIsWorth16x16WordsTransformersAtScale}, a variety of masked image models 
for self-supervised pre-training 
has been proposed. Aiming to reconstruct masked pixels \citep{ElNoubyX21AreLargeScaleDatasetsNecessarySelfSupervisedPreTraining,HeCVPR22MaskedAutoencodersAreScalableVisionLearners,XieCVPR22SimMIMSimpleFrameworkMaskedImageModeling}, discrete tokens~\citep{BaoICLR22BEiTBERTPreTrainingImageTransf,ZhouICLR22IBOTImageBERTPreRrainingWithOnlineTokenizer} or deep features~\citep{BaevskiX22data2vecAGeneralFrameworkSelfSupervisedLearningSpeechVisionLanguage,WeiX21MaskedFeaturePrediction4SelfSupervisedVisualPreTraining}, these methods have demonstrated the ability to scale to large datasets and models and achieve state-of-the-art results on various downstream tasks, including SiS. In particular, the masked autoencoder (MAE)~\citep{HeCVPR22MaskedAutoencodersAreScalableVisionLearners} accelerates pre-training by using an asymmetric architecture that consists of a large encoder that operates only on unmasked patches followed by a lightweight decoder that reconstructs the masked patches from the latent representation and mask tokens.
MultiMAE~\citep{BachmannX22MultiMAEMultiModalMultiTaskMaskedAutoencoders} leverages the efficiency of the MAE approach and extends it to multi-modal and multitask settings. Rather than masking input tokens randomly, the Masked Self-Supervised Transformer model (MST)~\citep{LiNIPS21MSTMaskedSelfSupTransformer} proposes to rely on the attention maps produced by a teacher network, to dynamically mask low response regions of the input, and a student network is then trained to reconstruct it.

Instead,~\citet{FangX22CorruptedImageModeling4SelfSupVisualPreTraining} propose the Corrupted Image Modeling (CIM) for self-supervised visual pre-training. CIM uses an auxiliary generator to corrupt the input image where some patches are randomly selected and replaced with plausible alternatives. Given such a corrupted image, an enhancer network learns to either recover all the original image pixels, and to predict whether a visual token is replaced by a generator sample or not. After pre-training, the enhancer can be used as a high-capacity visual encoder that achieves compelling results in image classification and semantic segmentation. 

The model fine-tuning in 
self-supervised SiS makes a transition towards the domain adaptation that we present in detail in the next chapter. We consider both fine-tuning and domain adaptation as instances of transfer learning. 
Fine-tuning uses the pre-trained models to initialize the target model parameters and update these parameters during training. In general, it requires labeled data from the target domain and does not use data from the source domain. Also the target task is often different from the source task.
Instead, domain adaptation is the process of adapting model(s) trained on source domain(s), by transferring information to improve model performance on the target domain(s). In general, labels are not available in the target set but source and target are associated with the same task. However, note that -- as we will see in \secref{beyond_dass} -- most recent DA models go beyond classical DA, 
thus making the distinction between the two approaches more subtle.

\chapter{Domain Adaptation for SiS (DASiS)}
\label{c:DAsemsegm}

The success of deep learning methods for SiS discussed in~\secref{deepSS} typically depends on the availability of large amounts of annotated training data. Manual annotation of images with pixel-wise semantic labels is an extremely tedious and time consuming process. 
Progress in computer graphics and modern high-level generic graphics platforms, such as game engines, enable the generation of photo-realistic virtual worlds with diverse, realistic, and physically plausible events and actions. The computer vision and machine learning communities realized that such tools can be used to generate datasets for training deep learning models~\citep{RichterECCV16PlayingForData}. Indeed, such synthetic rendering pipelines can produce a virtually unlimited amount of labeled data, leading to good performance when deploying models on real data, due to constantly increasing photorealism of the rendered datasets. Furthermore, it becomes easy to diversify data generation; for example, when generating scenes that simulate driving conditions, one can simulate seasonal, weather, daylight or architectural style changes, making such data generation pipeline suitable to support the design and training of computer vision models for diverse tasks, such as SiS.

While some SiS models trained on simulated images can already perform relatively well on real images, their performance can be further improved by domain adaptation (DA) -- and in particular {\it unsupervised domain adaptation} (UDA) -- by bridging the gap caused by the domain shift between the synthetic and real images. For the aforementioned reasons, sim-to-real adaptation represents one of the leading benchmarks to assess the effectiveness of domain adaptation for semantic image segmentation.

The main goal of DASiS is to ensure that SiS models trained on synthetic images perform well on real target data, by leveraging annotated synthetic and non-annotated real data. A classical DASiS framework relies on either SYNTHIA~\citep{RosCVPR16SYNTHIADataset} or GTA~\citep{RichterECCV16PlayingForData} dataset as a source, and the real-world Cityscapes~\citep{CordtsCVPR16CityscapesDataset} dataset as a target. Some known exceptions include domain adaptation between medical images~\citep{BermudezChaconISBI18ADomainAdaptiveUNetElectronMicroscopy,PeroneNI19UDAMedicalImSegmSelfEnsembling}, aerial images~\citep{LeeICCV21SelfMutatingNetworkDASSAerialImages}, weather and seasonal condition changes of outdoor real images~\citep{WulfmeierIROS17AddressingAppearanceChangeAdvDA}, and adaptation between different Field of View (FoV) images~\citep{GuICCV21PITPositionInvariantTransformForCrossFoVDA}.

Early DASiS methods have been directly inspired by adaptation methods originally designed for image classification~\citep{CsurkaX20DeepVisualDomainAdaptation,WangNC18DeepVisualDASurvey}.
However SiS is a more complex task, as predictions are carried out at the pixel level, where neighbouring pixels are strongly related (as discussed in Chapter \ref{c:semsegm}). DA methods for image classification commonly embed entire images in some latent space and then align source and target data distributions. Directly applying such a strategy to SiS models is sub-optimal, due to the higher dimensionality and complexity of the output space. To address this complexity, most DASiS methods take into account the spatial structure and the local image context, act at multiple levels of the segmentation pipeline and often combine multiple techniques.

Therefore, to overview these methods, we step away from grouping the DA methods into big clearly distinguishable families,  as it is done in recent surveys on image classification~\citep{CsurkaX20DeepVisualDomainAdaptation,WangNC18DeepVisualDASurvey}.
We instead identify a number of critical characteristics of existing DASiS pipelines and categorize the most prominent methods according to them. From this point of view,~\tabl{semsegm_methods} is one of our major contributions. It is detailed in Section~\ref{sec:dass_siamese}, where we describe the different domain alignment techniques that are applied at input image, feature and output prediction levels. In Section~\ref{sec:dass_improvements}, we describe complementary machine learning strategies that can empower domain alignment and improve the performance of a segmentation model on target images. Before presenting these different methods, in the next section we first formalize the UDA problem and list the most popular domain alignment losses optimized by a large majority of  DA approaches.

\section{Brief Introduction into UDA}
\label{sec:UDA}

Let $\calD_\calS = \calX_\calS \times \calY_\calS$ be a set of paired sample images with their ground-truth annotated segmentation maps ($\calX_\calS=\{\bfx_i\}_{i=1}^M$ and $\calY_\calS=\{\bfy_i\}_{i=1}^M$, respectively), drawn from a source distribution $P_\calS(\calX,\calY)$. In the SiS context, $\bfx$ and $\bfy$ represent images and their pixel-wise annotations, respectively, $\bfx \in \mathcal{R}^{H \times W \times 3}$ and $\bfy \in \mathcal{R}^{H \times W \times C}$, where $(H, W)$ is the image size and $C$ is the number of semantic categories. Let $\calD_\calT = \calX_\calT = \{\bfx_i\}_{i=1}^N$ be a set of unlabeled samples drawn from a target distribution $P_\calT$, such that $P_\calS \neq P_\calT$ due to the domain shift. In the UDA setup, both sets are available at training time ($\calD = \calD_\calS \cup \calD_\calT$) and  the goal is to learn a model performing well on samples from the target distribution.

\begin{figure}[ttt]
\begin{center}
\includegraphics[width=0.75\textwidth]{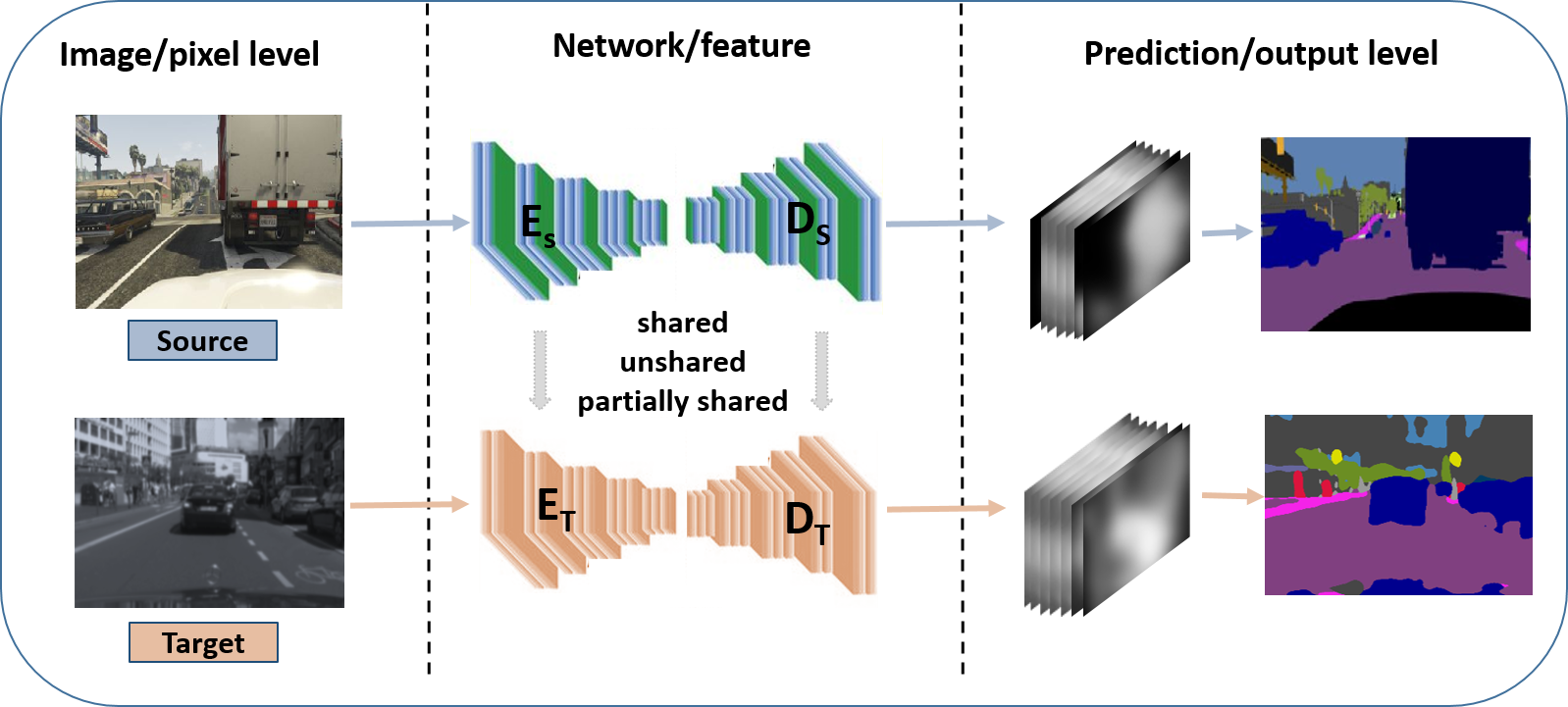}
\caption{Several DASiS models adopt a Siamese architecture with one branch per domain, where the alignment is carried out at different levels of the pipeline: at {\it pixel level}, by transferring the image style from one domain to another; at {\it network level}, by aligning the image features, derived from activation layers and sharing (full or partial) the network parameters; and finally at the {\it output level}, by aligning the class prediction maps.}
\label{fig:AdaptLevels}
\end{center}
\end{figure}

Often, the segmentation network used in DASiS has an encoder-decoder structure (see \fig{AdaptLevels}) and the domain alignment can happen at different levels of the segmentation network, including the output of the encoder, at various level of the decoder, or even considering the label predictions as features (as discussed later in this section). Hence, the features used to align the domains can be extracted at image level, region level or pixel-level. Therefore, when we use the notation of $F_S$ and $F_T$ to refer to any of the above source respectively target feature generator. Note that it is frequent that the feature encoders $F_S$ and $F_T$ share their parameters $\theta_{F_S}$ and $\theta_{F_T}$ -- in this case, we simply refer to them as $F$ and $\theta_{F}$.

In the following, we will cover some basic components and commonly used losses that constitute the foundation of most UDA and DASiS approaches.

\myparagraph{Distribution Discrepancy Minimization}
In image classification, one popular approach to address UDA is to minimize the distribution discrepancy between source and target domains in some latent feature space -- that is, the space of the learned visual representation. This space often corresponds to some activation layers; most often the last layer the classifier is trained on, but other layers can be considered as well.
One popular measure is the empirical Maximum Mean Discrepancy (MMD)~\citep{BorgwardtBI06IntegratingKernelMaximumMeanDiscrepancy}, that is written as
\begin{align*}
\Loss_{mmd} = 
\left\| \frac{1}{M}  \sum_{ \bfx_s\in \calX_\calS} \phi(F_S(\bfx_s)) - \frac{1}{N}  \sum_{ \bfx_t\in \calX_\calT} \phi(F_T(\bfx_t)) \right\|\;,
\end{align*}
where $\phi$ is the mapping function corresponding to a Reproducing Kernel Hilbert Space (RKHS) kernel defined as a mixture of Gaussian kernels.

\myparagraph{Adversarial Training}
An alternative to minimizing the distribution discrepancy between source and target domains is given by adversarial training~\citep{GoodfellowNIPS14GenerativeAdversarialNets}.
Multiple studies have shown that domain alignment can be achieved by learning a domain classifier $C_{Disc}$ (the {\it discriminator}) with the parameters $\theta_D$ to distinguish between the feature vectors from source and target distributions and by using an adversarial loss to increase {\it domain confusion}~\citep{GaninJMLR16DomainAdversarialNN,TzengICCV15SimultaneousDeepTL,TzengCVPR17AdversarialDiscriminativeADDA}. The main, task-specific deep network -- in our case SiS --  aims to learn a representation that fools the domain classifier, encouraging encoders to produce domain-invariant features. Such features can then be used by the final classifier, trained on the source data, to make predictions on the target data. Amongst the typical adversarial losses, we mention the \textit{min-max game} proposed by \citet{GaninJMLR16DomainAdversarialNN}
\begin{align*}
  \Loss_{adv} = \min_{\theta_{F},\theta_{C}} \max_{\theta_{D}} \, &\{ \Exp_{\bfx_s \in \calX_\calS} [\mathcal{L}_{Task}(F(\bfx_s), \bfy_s)] \\
  &- \lambda \cdot \Exp_{\bfx \in \calX_\calS \cup \calX_\calT} [\mathcal{L}_{Disc}(F(\bfx), \bfy_d)] \},
\end{align*}
where $\mathcal{L}_{Task}$ is the loss associated with the task of interest (it depends on both the feature encoder parameters $\theta_F$ and the final classifier's parameters $\theta_C$), $\mathcal{L}_{Disc}$ is a loss measuring how well a discriminator model parametrized by $\theta_D$ can distinguish whether a feature belongs to source ($\bfy_d=1$) or to target domain ($\bfy_d=0$), and $\lambda$ is a trade-off parameter.
By alternatively training the discriminator $C_{Disc}$ to distinguish between domains and the feature encoder $F$ 
to \textit{fool} it, one can learn domain agnostic features. Also, training the encoder and the final classifier $C_{task}$ for the task of interest, guarantees that such features are not simply domain-invariant, but also discriminative.

An effective way to approach this minimax problem consists in introducing a Gradient Reversal Layer (GRL)~\citep{GaninJMLR16DomainAdversarialNN} which reverses the gradient direction during the backward pass in backpropagation (in the forward pass, it is inactive). The GRL allows to train the discriminator and the encoder at the same time.

A related but different approach by \citet{TzengCVPR17AdversarialDiscriminativeADDA} brings adversarial training for UDA closer to the original GAN formulation~\citep{GoodfellowNIPS14GenerativeAdversarialNets}. It splits the training procedure into two different phases: a fully discriminative one, where a module is trained on source samples, and a fully generative one, where a GAN loss is used to learn features for the target domain that mimic the source ones -- or, more formally, that are projected into the same feature space, on which the original classifier is learned. This second step can be carried out by approaching the following minimax game
\begin{align*}
  \Loss_{GAN} = \min_{\theta_{F_\calT},\theta_{F_S}} \max_{\theta_{D}} \, \left\{ \Exp_{\bfx_s \in \calX_\calS} [\log(C_{Disc}(F_S(\bfx_s)))] \right.
  \\
    +  \left. \Exp_{\bfx_t \in \calX_\calT} [\log(1-C_{Disc}(F_T(\bfx_t)))] \right\},
\end{align*}
\noindent
where $C_{Disc}$ is the discriminator, and both $F_S$ and $F_T$ are initialized with the weights pre-trained by supervised learning on the source data.

\begin{table*}
\begin{center}
{
\tiny
\setlength{\tabcolsep}{4pt}
\begin{NiceTabular}{|l|c|c|c|c|c|c|c|c|}[code-before = 
\rowcolor{gray!10}{4}
\rowcolor{gray!10}{6}
\rowcolor{gray!10}{8}
\rowcolor{gray!10}{10}
\rowcolor{gray!10}{12}
\rowcolor{gray!10}{14}
\rowcolor{gray!10}{16}
\rowcolor{gray!10}{18}
\rowcolor{gray!10}{20}
\rowcolor{gray!10}{22}
\rowcolor{gray!10}{24}
\rowcolor{gray!10}{26}
\rowcolor{gray!10}{28}
\rowcolor{gray!10}{30}
\rowcolor{gray!10}{32}
\rowcolor{gray!10}{34}
\rowcolor{gray!10}{36}
\rowcolor{gray!10}{38}
\rowcolor{gray!10}{40}
\rowcolor{gray!10}{42}
\rowcolor{gray!10}{44}
\rowcolor{gray!10}{46}
\rowcolor{gray!10}{48}
\rowcolor{gray!10}{50}
]
\toprule

\textbf{Citation} & \textbf{Segm.} & \textbf{ Image} 
  & \textbf{Net} &  \textbf{Sha-} 
    & \textbf{CW}  & \textbf{Output} & \textbf{Comple-} & \textbf{Specificity}\\
  & \textbf{Net} & \textbf{Level} & \textbf{Level} 
         & \textbf{red} & \textbf{feat.} & \textbf{Level}  & \textbf{mentary} & \\
\midrule

\cite{HoffmanX16FCNsInTheWildPixelLevelAdversarialDA} & FCNs & - &  mDC   & \checkmark & \checkmark & mInstLoss & - &  class-size hist transfer\\
%
\cite{ChenICCV17NoMoreDiscriminationCrossCityAdaptationRoadSceneSegmenters} & dFCN & - & DC &  \checkmark & \checkmark &- & SelfT & static obj prior \\
\cite{PeroneNI19UDAMedicalImSegmSelfEnsembling} &  UNet & Aug & EMA &  - & - & SemCons & SelfEns & Dice loss/medical \\
%
\cite{ChenCVPR18RoadRealityOrientedDASemSegmUrbanScenes} & DLab/PSPN  &  - & DC  & - & - & - & Distill & spatial aware adaptation \\ 
\cite{HuangECCV18DomainTransferThroughDeepActivationMatching} & ENet &- & DC &  - & - & - & - & Jensen-Shannon divergence\\
\cite{HongCVPR18ConditionalGANStructuredDA} & FCNs & - & DC &  \checkmark & \checkmark & - & - &  CGAN/target-like features \\
\cite{HoffmanICML18CyCADACycleConsistentAdversarialDA} & FCN & S$\leftrightarrow$T & DC  & - &- & SemCons  &   - &  CycleGAN \\
\cite{LiBMVC18SemanticAwareGradGANVirtualToRealUrbanSceneAdaption}& UNet& S$\leftrightarrow$T & - & - & \checkmark & - & - & PatchGAN/semantic-aware gradient \\
\cite{MurezCVPR18ImageToImageTranslationDA} & dFCN  & S$\leftrightarrow$T & DC    &- & - & SemCons &- & dilated DenseNet\\
\cite{SaitoCVPR18MaximumClassifierDiscrepancyUDA}   & FCN & - &  - &  \checkmark & - & MCD &  CoT  &  minimize/maximize  discrepancy\\
\cite{SaitoICLR18AdversarialDropoutRegularization} & FCN &  - & -   &  \checkmark & - & MCD &  CoT & adversarial drop-out  \\
%
\cite{TsaiCVPR18LearningToAdaptStructuredOutputSpaceSemSegm} & DeepLab & - & -  & \checkmark & - &  mDC  &  - & multi-level predictions\\ 
\cite{WuECCV18DCANDualChannelWiseAlignmentNetworksUDA} & FCN &  S$\rightarrow$T  & DM &  \checkmark & - & -  & -  & channel-wise Gramm-matrix align \\
\cite{ZhuECCV18PenalizingTopPerformersConservativeLossSemSegmDA} & FCN &  S$\leftrightarrow$T  & - & \checkmark & - &  -& - & conservative loss\\
\cite{ZouECCV18UnsupervisedDASemSegmClassBalancedSelfTraining}&   FCN &  -  & - & \checkmark &  - & -  & SelfT & self-paced curriculum/spacial priors\\
%
\cite{ChangCVPR19AllAboutStructureDASemSeg} & DeepLab & S$\leftrightarrow$T & -   & - &  - &   DC &- & domain-specific encoders/percL  \\
\cite{ChenCVPR19CrDoCoPixelLevelDomainTransferCrossDomainConsistency} & FCN/DRN  & S$\leftrightarrow$T & DC & - &  - &   SemCons & -  & KL cross-domain consistency \\ 
\cite{ChenICCV19DASemSegmentationMaximumSquaresLoss} &  DeepLab & - & - &  \checkmark & - & - & SelfT & max squares loss/img-wise weights \\
\cite{ChoiICCV19SelfEnsemblingGANBasedDataAugmentationDASemSegm} &ASPP  & S$\rightarrow$T  & EMA & - & - & SemCons & SelfEns &  target-guided+cycle-free augm. \\
\cite{DuICCV19SSFDANSeparatedSemanticFeatureDASemSegm}
& FCN &  -& DC &  \checkmark & \checkmark & - & PL  &class-wise adversarial reweighting \\
\cite{LeeCVPR19SlicedWassersteinDiscrepancyUDA} &  PSPNet & - & - & \checkmark  & -  & MCD & CoT &   sliced Wasserstein discrepancy \\
\cite{LuoCVPR19TakingACloserLookDomainShiftSemanticsConsistentDA} & FCN/DLab & - & -  &   \checkmark   &   - &  MCD & CoT &  local consistency/self-adaptive weight  \\
\cite{LiCVPR19BidirectionalLearningDASemSegm} & DeepLab & S$\leftrightarrow$T &  -  & \checkmark &  -   &  DC & SelfT  & perceptual loss (percL) \\ 
\cite{LianICCV19ConstructingSelfMotivatedPyramidCurriculumsCrossDomainSemSegm} & FCN/PSPN  & - & - &  \checkmark  & -  & - & CurrL & self-motivated pyramid curriculum \\
\cite{LuoICCV19SignificanceAwareInformationBottleneckDASemSegm} & FCN/DLab & - & DC  &   \checkmark & - & - & - & signif.-aware information bottleneck\\
\cite{ShenX19RegularizingProxiesMultiAdversarialTrainingUDASS} & ASPP & - & mDC &   \checkmark & \checkmark  & ConfMap  & SelfT  &   cls+adv-conf./class-balance weighs  \\
\cite{XuAAAI19SelfEnsemblingAttentionNetworksSemSegm} & DeepLab & Aug & EMA &  - & - & SemCons & SelfEns & 
self-ensembling attention maps \\
\cite{VuCVPR19ADVENTAdversarialEntropyMinimizationDASemSegm} & DLab/ASPP & - & - &\checkmark &  - &  DC  & TEM &  entropy map align./class-ratio priors\\
\cite{HuangECCV20ContextualRelationConsistentDASemSegm} &  DeepLab & - & - & \checkmark & \checkmark &  DC & TEM & local contextual-relation \\
%
\cite{LvCVPR20CrossDomainSemSegmentationInteractiveRelationTransfer} & FCN/DLab &  - &  - & \checkmark  & - & SemCons & SelfT & course-to-fine segm. interaction \\
\color{blue}{Musto et al. (2020)}
& FCN/DLab & S$\leftrightarrow$T & -  & \checkmark & - &  SemCons&  SelfT & spatially-adaptive normalization  \\
\cite{PanCVPR20UnsupervisedIntraDASemSegmSelfSupervision}& DeepLab & - & -  &   - &  - & DC & CurrL & align entropy maps/easy-hard split \\
\cite{ToldoIVC20UDAMobileSemSegmCycleConsistencyFeatureAlignment} & DeepLab & S$\leftrightarrow$T & DC  & \checkmark & - & SemCons &    - &  MobileNet \\
\cite{WangCVPR20DifferentialTreatmentStuffThingsDASemSegm} & ASPP & - & DC &  \checkmark &  \checkmark & - & SelfT &
disentangled things and stuff \\
\cite{YangCVPR20PhaseConsistentEcologicalDA}& Unet  &S$\leftrightarrow$T  & - &  \checkmark & - & -  &  SelfT & phase cons./cond. prior network\\
%
\color{blue}{Yang et al. (2020)}
& FCN/DL &  S$\rightarrow$T & - &  \checkmark & -   & - & SelfT & Fourier transform/low-freq. swap \\
\cite{YangAAAI20AnAdversarialPerturbationOrientedDASS} & DeepLab  & - & advF  & \checkmark & - & 
DC & TEM & adv. attack/feature perturbation\\ %
\cite{YangECCV20LabelDrivenReconstructionDASemSegm}& FCN/DLab & T$\rightarrow$S & - &  \checkmark   & - & DC & SelfT & reconstruction from predictions\\
\cite{ZhangCVPR20TransferringRegularizingPredictionSemSegm} &  FCN &  - & DC &  \checkmark & - & LocCons & - &
patch+cluster+spatial consistency\\
%
\color{blue}{Zheng et al. (2020)}
& PSPNet & - & -  & \checkmark & - & MCD  & CoT &  memory regularization (KL) \\
\color{blue}{Araslanov et al. (2021)} & DeepLab & Aug & EMA & - &  & SemCons  & SelfT & self-sup/imp. sampling/focal loss\\
\cite{ChengICCV21DualPathLearning4DASS} &
DLab/FCN &S$\leftrightarrow$T &   - & - & - & SemCons & SelfT & dual perceptual loss/dual path DASS\\
\cite{GuoCVPR21MetaCorrectionDomainAwareMetaLossCorrectionDASS}  & DeepLab  & - & - &  \checkmark & - & - & SelfT &  meta-learning/meta-loss correction \\
\cite{ToldoWACV21UDASemSegmOrthogonalClusteredEmbeddings}   &  DeepLab & - & Clust. &  \checkmark &  \checkmark & &  TEM & discriminative clustering \\
\cite{TruongICCV21BiMaLBijectiveMaximumLikelihoodDASS}  & DeepLab & - & -&  \checkmark & - &  - & TEM & bij. max. likelihood/local consistency
\\
%
\cite{WangICCV21UncertaintyAwarePseudoLabelRefineryDASS} & DeepLab &  S$\rightarrow$T & -& \checkmark  & - & DC  & SelfT & target-guided uncertainty rectifying\\
\cite{WangAAAI21ConsistencyRegularizationSourceGuidedPerturbationUDASemSegm} & FCN & S$\leftrightarrow$T & EMA & - & - & - & SelfEns  & AdaIn/class-balanced
reweighting \\
%
\cite{YangICCV21ExploringRobustnessDASS} &
DeepLab & S$\rightarrow$T & - & \checkmark  & - & contrL & SelfT &
adv. attack/adv. self-supervised loss  \\  
 \cite{ChenX22SmoothingMattersMomentumTransformerDASIS} & SwinTr & - & DC & \checkmark  &  \checkmark & - & Distill & Momentum Transformer \\

\bottomrule
\end{NiceTabular}
} 
\end{center}
\caption{
Summary of the DASiS methods, according to their characteristics.
\textbf{Segmentation Network}: The neural network used as a backbone, 
\textbf{Image Level}: alignment at image level (by using style transfer), from source to target S$\rightarrow$T, target to source S$\leftarrow$T or both  S$\leftrightarrow$T; Aug stands for specific data augmentation strategies. 
\textbf{Network Level}: alignment at feature level 
where DC stands for adversarial domain confusion at single or  multiple (mDC) feature levels,  DM is discrepancy minimization between feature distributions, Clust stands for feature level clustering, AdvF for adversarial features.  Alternatively, EMA (exponential moving average) is model parameter adaptation
from  student to a teacher model or inversely.
\textbf{Shared}: the  parameters of the segmentation network are shared (\checkmark ) or  at least partially domain specific (-). 
\textbf{CW}:  class-wise feature alignment or clustering.
\textbf{Output Level}: alignment 
or regularization at output level: DC/DM  with  features  extracted from the predicted maps
or confusion maps (ConfMap),
MCD  stands for adversarial training of multi-classifier discrepancy. Further notations:  contrL (contrastive loss),
SemCons (semantic consistency loss between  predicted segmentation  maps), LocCons (local consistency),  mInstLoss (multi-instance loss).
\textbf{Complementary} indicates which complementary techniques are used, such as
exploiting pseudo labels (PL), 
self-training (SelfT), curriculum learning (CurrL), target (conditional) entropy minimisation (TEM),
CoT (co-training),
self-ensembling (SelfEns), 
model distillation (Distill).
\textbf{Specificity} reports important elements that are not covered by the previously described common characteristics.
}
\label{tab:semsegm_methods}
\end{table*}


\section{Adapting SiS between Domains}
\label{sec:dass_siamese}

Since the advent of representation learning solutions in most of machine learning applications, UDA research has witnessed a shift towards end-to-end solutions to learn models that may perform well on target samples. In image classification, a very successful idea has been to learn a representation where the source and target samples get {\it aligned} -- that is, the source and target distributions are close in the feature space under some statistical metrics.

This~\textit{alignment} is often achieved by means of a Siamese
architecture~\citep{Bromley93IJPRAISignatureVerificationTimeDelayNN} with two streams,
each corresponding to a semantic segmentation model: one stream is aimed at processing source samples and the other at processing the target ones (as shown in~\fig{AdaptLevels}). 
The parameters of the two streams can be shared, partially shared or domain specific; generally, the backbone architectures of both streams are initialized with weights that are pre-trained on the source set. The Siamese network is typically trained
with a loss comprising two terms. For what concerns SiS, one term is the standard~\textit{pixel-wise cross-entropy} loss 
(referred in this paper also as $\Loss_{Task}$), measuring performance on source samples for which the ground-truth annotations are available
\begin{align*}
\Loss_{ce}  =  - \mathbb{E}_{
(\calX_\calS, \calY_\calS) }
 \left[  \sum_{h,w,c}{\bfy_s^{(h,w,c)} \cdot  \log(p^{(h,w,c)}(F_S(\bfx_s))}) \right],
\end{align*}
where $p^{(h,w,c)}(F_S(\bfx_s))$ is a probability of class $c$ at pixel $\bfx_s^{(h,w)}$ and $\bfy_s^{(h,w,c)}$ is  1 if $c$ is the pixel's true class and 0 otherwise.

The second term is a~\textit{domain alignment} loss that measures the distance between source and target samples. The alignment can be addressed at different levels of the pipeline, as illustrated in~\fig{AdaptLevels}, namely, network (feature), at image (pixel) and output (prediction) levels, as detailed in Sections~\ref{sec:feature-level},~\ref{sec:image-level} and~\ref{sec:output-level}, respectively. Note that -- as shown in Table~\ref{tab:semsegm_methods} -- many approaches apply alignment at multiple levels.

While aligning the marginal feature distributions tends to reduce the domain gap, it can be sub-optimal as it does not explicitly take the specific task of interest (in this case, SiS) into account during the domain alignment as discussed for example in~\citep{ZhaoICML19OnLearningInvariantReprDA}. To overcome these weaknesses, several works have been proposed to leverage the class predictions during the alignment, what we call output level alignment (see \secref{output-level}). Furthermore, there is a growing  interest for adaptation at pixel-level (see \secref{image-level}), Indeed, the shift between two domains is often strongly related to visual appearance variations such as day \vs night, seasonal changes, synthetic \vs real. By exploiting the progress of image-to-image translation and style transfer brought by deep learning-based techniques~\citep{HuangICCV17ArbitraryStyleTransferRealTimeAdaptiveInstanceNormalization,ZhuICCV17UnpairedI2ICycleConsistentAdversarialNetworks}, several DASiS methods have been proposed to explicitly account for such stylistic domain shifts by performing an alignment at image level.

In the following, we discuss in details alignment solutions between source and target at various levels of the segmentation pipeline.

\subsection{Feature-level adaptation} 
\label{sec:feature-level}

Generic DA solutions proposed for image classification perform domain alignment in a latent space by minimizing some distance metrics, -- such as the maximum mean discrepancy (MMD)~\citep{LongICML15LearningTransferableFeaturesDAN} between feature distributions of source and target data, -- or by adversarially training a domain discriminator to increase {\it domain confusion}~\citep{GaninJMLR16DomainAdversarialNN,TzengICCV15SimultaneousDeepTL,TzengCVPR17AdversarialDiscriminativeADDA}. Both approaches scale up to semantic segmentation problems. In particular, adversarial training has been largely and successfully applied and combined with other techniques.

\begin{figure*}[ttt]
\begin{center}
\includegraphics[width=0.85\textwidth]{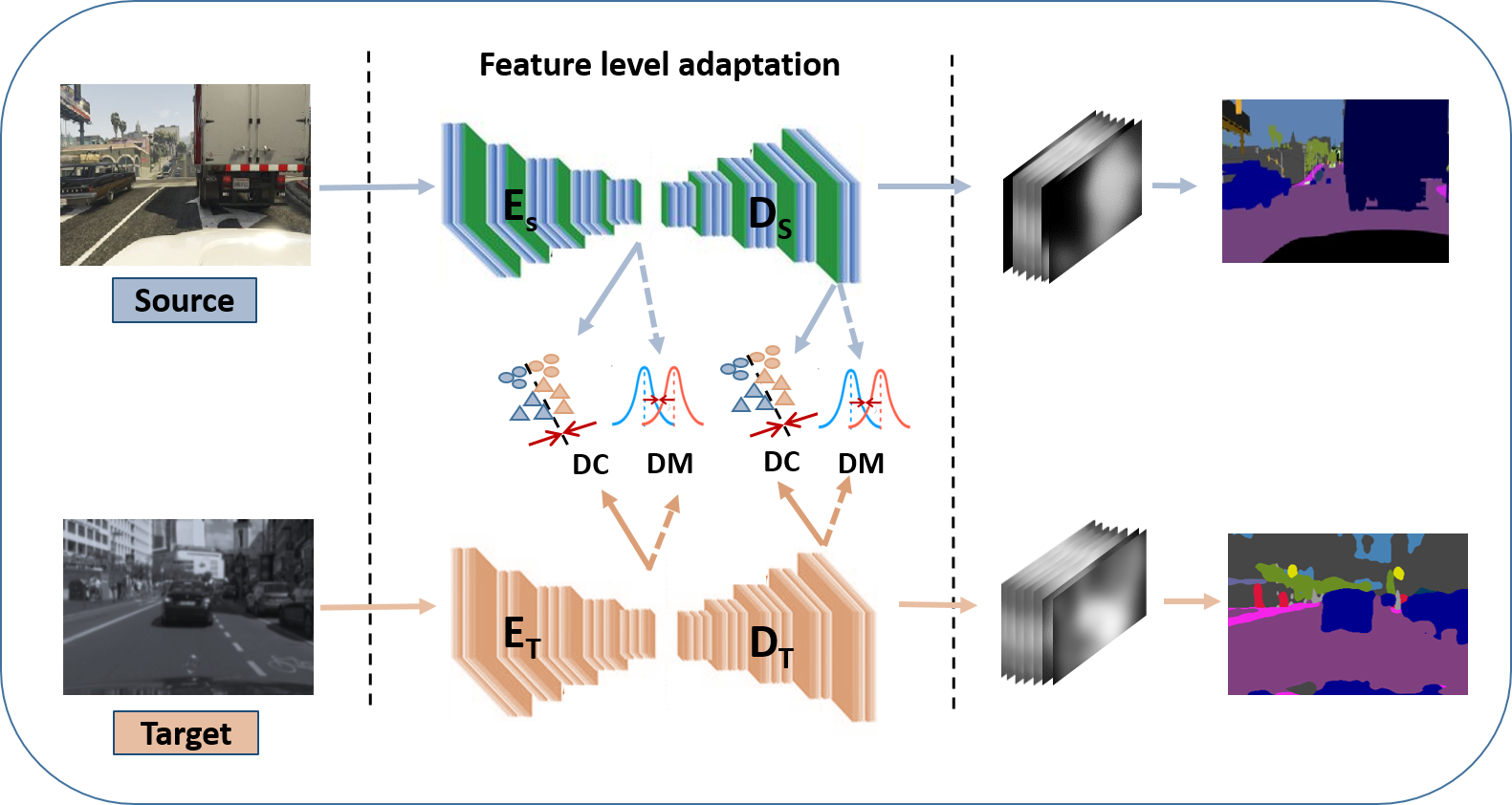}
\caption{In generic DA, domain alignment is often performed in a single latent representation space.
In DASiS, the alignment is often done at multiple layers, by discrepancy minimization between feature distributions or by adversarial learning relying on a domain classifier (DC) to increase domain confusion. Encoders and decoders of the segmentation network are often shared: $E_S=E_T$, $D_S=D_T$.}
\label{fig:FLAdapt}
\end{center}
\end{figure*}

In DASiS, we consider more complex models to tackle the SiS task.
We recall that adaptation in SiS is more challenging than in image classification, due to the structural complexity and the scale factor of the task. Indeed it is rather difficult and sub-optimal to fully capture and handle the DASiS problem by simple alignment of the latent global representations (activation layers) between domains. Therefore, the domain alignment is often carried out at different layers of the network (see Figure~\ref{fig:FLAdapt}). The alignment is done either by minimizing the feature distribution discrepancy~\citep{BermudezChaconISBI18ADomainAdaptiveUNetElectronMicroscopy} or by adversarial training via a domain classifier to increase domain confusion~\citep{HoffmanX16FCNsInTheWildPixelLevelAdversarialDA,HuangECCV18DomainTransferThroughDeepActivationMatching,ShenX19RegularizingProxiesMultiAdversarialTrainingUDASS}. 

While some works consider for the alignment simply a global representation of the image~\citep{HuangECCV18DomainTransferThroughDeepActivationMatching} -- by flattening or pooling the activation map -- most often pixel-wise~\citep{HoffmanX16FCNsInTheWildPixelLevelAdversarialDA}, grid-wise~\citep{ChenICCV17NoMoreDiscriminationCrossCityAdaptationRoadSceneSegmenters} or region-wise~\citep{ZhangCVPR20TransferringRegularizingPredictionSemSegm} representations are used. Furthermore, to improve the model performance on the target data, such methods are often combined with some prior knowledge or specific tools as discussed below (and also in \secref{dass_improvements}).

In their seminal work,~\citet{HoffmanX16FCNsInTheWildPixelLevelAdversarialDA} combine the distribution alignment with the class-aware constrained multiple instance loss used to transfer the spatial layout.
\citet{ChenICCV17NoMoreDiscriminationCrossCityAdaptationRoadSceneSegmenters} consider global and class-wise domain alignment and address it via adversarial training. In particular, they rely on local class-wise domain predictions over image grids assuming that the composition/proportion of object classes across domains -- different urban environments in their case -- is similar.

\citet{HongCVPR18ConditionalGANStructuredDA} rely on a conditional generator that transforms the source features into target-like features, using a multi-layer perceptron as domain discriminator. Assuming that decoding these target-like feature maps preserve the semantics, they are used with the corresponding source labels within an additional cross-entropy loss to make the model more suitable  for the target data.

\begin{figure*}[ttt]
\begin{center}
\includegraphics[width=\textwidth]{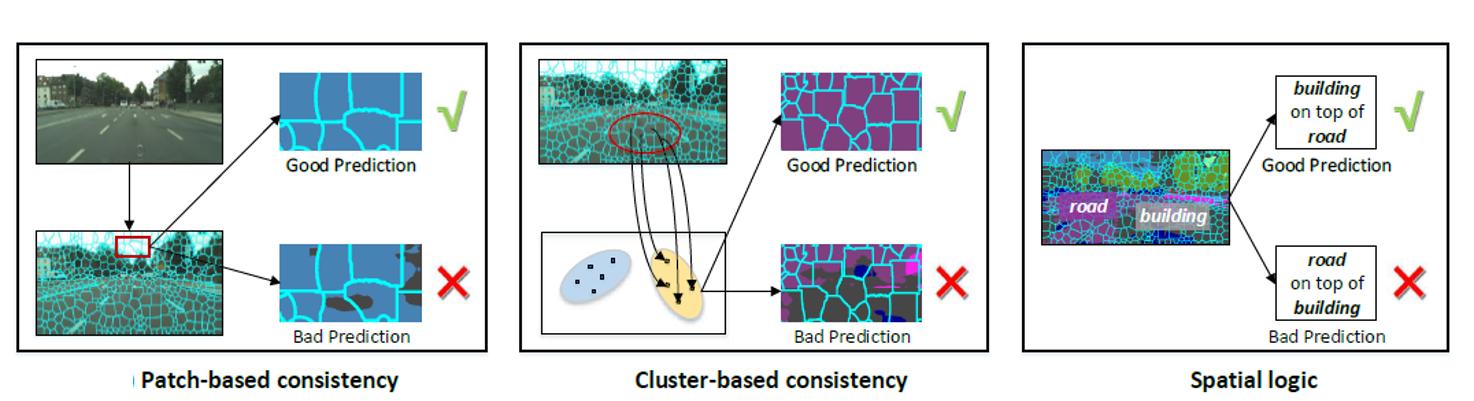}
\caption{To improve the domain alignment, \citet{ZhangCVPR20TransferringRegularizingPredictionSemSegm} propose to reduce patch-level, cluster-level and context-level inconsistencies
(Figure based on~\citep{ZhangCVPR20TransferringRegularizingPredictionSemSegm}).}
\label{fig:ZhangRegularize}
\end{center}
\end{figure*}

The  Pivot Interaction Transfer ~\citep{LvCVPR20CrossDomainSemSegmentationInteractiveRelationTransfer}
consists in optimizing a semantic consistency loss between image-level and  pixel-level semantic information. This is achieved by training the model with both a fine-grained component producing pixel-level segmentation and coarse-grained components generating class activation maps obtained by multiple region expansion units,  trained with image-level category information independently.
\citet{ZhangCVPR20TransferringRegularizingPredictionSemSegm}, to improve alignment, explore three label-free constraints as model regularizer,  enforcing \textit{patch-level}, \textit{cluster-level} and \textit{context-level} semantic prediction consistencies at different levels of image formation (see~\fig{ZhangRegularize}).

\subsection{Image-level adaptation}
\label{sec:image-level}

\begin{figure*}[ttt]
\begin{center}
\includegraphics[width=0.95\textwidth]{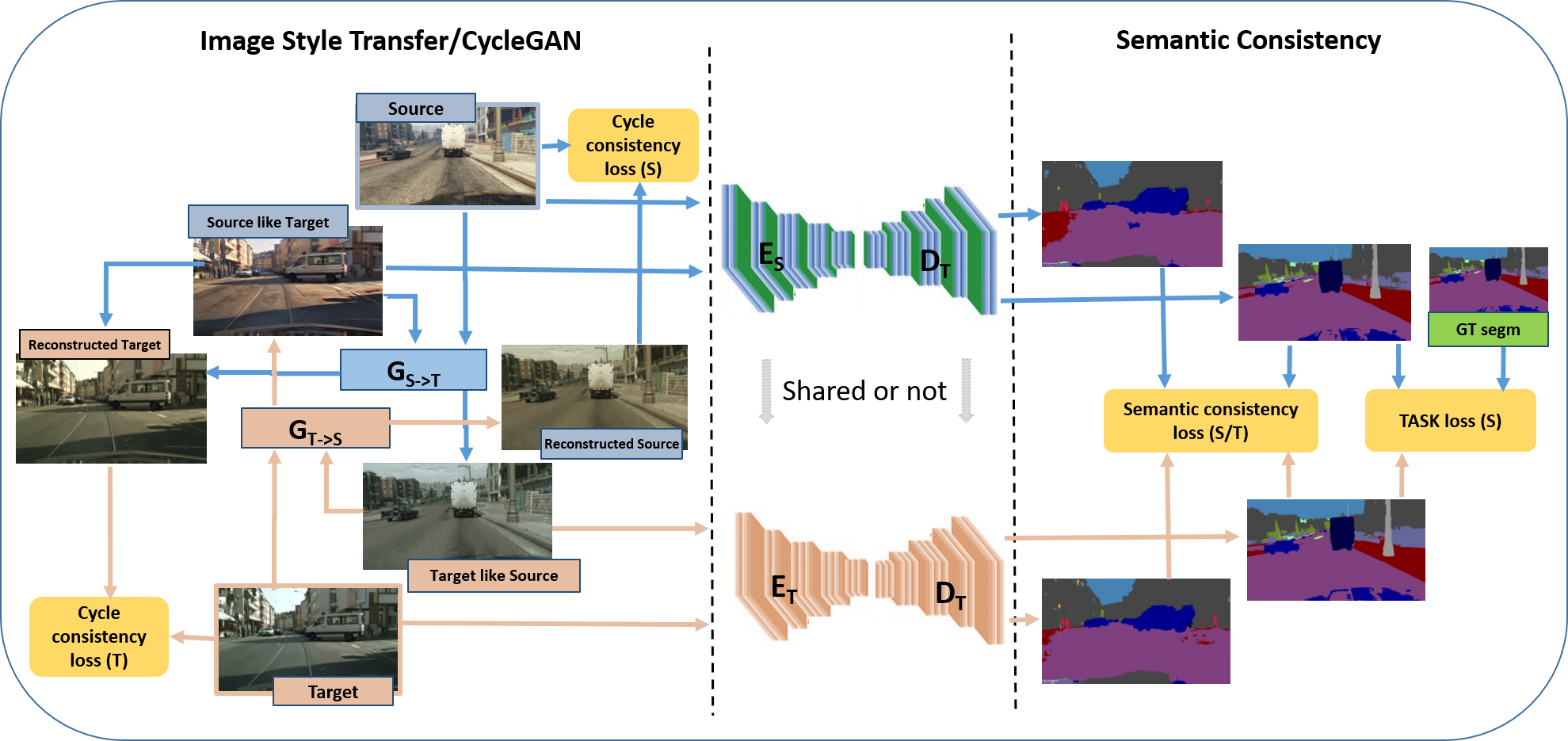}
\caption{In image-level adaptation that relies on image style transfer (IST), the main idea is to translate the \textit{style} of the target domain to the source data and/or the source style to the target domain. In order to improve the style transfer, often the \textit{style transferred} image is translated back to the original domain allowing to use a \textit{cyclic consistency} reconstruction loss. The style transferred source images \textit{inherit} the semantic content of the source and thus its pixel-level labeling, that allows the segmentation network to learn a model suitable for the target domain. On the other hand, the target and the \textit{source-like} target image share the content and therefore imposing that their predicted segmentation should match, -- using the \textit{semantic consistency} loss as a regularization, --  which  helps improving the model performance in the target domain.}
\label{fig:IST}
\end{center}
\end{figure*}

This class of methods relies on image style transfer (IST) methods, where the main idea is to transfer the \textit{domain  style} (appearance) from target to source, from source to target, or considering both (see illustration in~\fig{IST}).
The \textit{style transferred} source images maintain the semantic content of the source, and therefore its pixel-level labeling too, while their appearance results more similar to the target style -- helping the network to learn a model more suitable for the target domain~\citep{CsurkaTASKCV17DiscrepancyBasedNetworksUnsupervisedDAComparativeStudy,ThomasACCV19ArtisticObjectRecognitionUnsupervisedStyleAdaptation}.

Image-to-image translation for UDA has been pioneered within the context of image classification~\citep{BousmalisCVPR17UnsupervisedPixelLevelDAwGANs,LiuNIPS16CoupledGenerativeAdversarialNetworks,TaigmanICLR17UnsupervisedCrossDomainImageGeneration}; typically, such methods employ GANs \citep{GoodfellowNIPS14GenerativeAdversarialNets} to transfer the target images' style into one that resembles the source style. This approach has been proved to be a prominent strategy also within DASiS~\citep{ChangCVPR19AllAboutStructureDASemSeg,ChenCVPR19CrDoCoPixelLevelDomainTransferCrossDomainConsistency,HoffmanICML18CyCADACycleConsistentAdversarialDA,MurezCVPR18ImageToImageTranslationDA,ToldoIVC20UDAMobileSemSegmCycleConsistencyFeatureAlignment,SankaranarayananCVPR18LearningSyntheticDataDomainShiftSemSegm,WuECCV18DCANDualChannelWiseAlignmentNetworksUDA,YangCVPR20PhaseConsistentEcologicalDA}. Still, as in the case of feature alignment, for a better adaptation most methods combine the image translation with other ingredients (see also Table~\ref{tab:semsegm_methods}),  most often with self-training and different consistency regularization terms (detailed in Section \ref{sec:dass_improvements}).

The most used regularization terms in IST based DASiS are the \textit{cycle consistency} loss and the \textit{semantic consistency} loss proposed by \citet{HoffmanICML18CyCADACycleConsistentAdversarialDA}.
The proposed CyCADA is one of the first model that adopted image-to-image translation -- and in particular the consistency losses pioneered by Cycle-GAN~\citep{ZhuICCV17UnpairedI2ICycleConsistentAdversarialNetworks} -- for the DASiS problem.
The cycle consistency loss is defined as follows
\begin{align*}
	\begin{split}
	\Loss_{cycle} &  =
	 \Exp_{\bfx_s \sim \calX_\calS} \big[ \| G_{T \rightarrow S} \left( G_{S \rightarrow T} \left(\bfx_s\right) \right) - \bfx_s \|_k\big] \\
	& + \Exp_{\bfx_T\sim \calX_\calT}  \big[ \| G_{S \rightarrow T} \left( G_{T \rightarrow S} \left(\bfx_t\right) \right) - \bfx_t \|_k\big].
	\end{split}
\end{align*}
where $G_{S \rightarrow T}$ and $G_{T \rightarrow S}$ are the image generators  that learn to map the style from source to target and target to source, respectively and $\|\cdot \|_k$ is the L$k$ loss, where most often the L1 or the L2 loss is used. In short, this loss encourages the preservation of structural properties during the style transfer, while the semantic consistency loss
\begin{align*}
	\begin{split}
  \Loss_{SemCons} & =
  \mathcal{L}_{Task} \left( F_S(G_{S \rightarrow T}(\bfx_s)), p(F_S(\bfx_s)) \right) \\
& + \mathcal{L}_{Task} \left( F_\calS(G_{T \rightarrow S}(\bfx_t)), p(F_S(\bfx_t))  \right),
	\end{split}
\end{align*}
enforces an image to be labeled identically before and after translation. The task loss  $\mathcal{L}_{Task}$ here is the source pixel-wise cross-entropy, but instead of using GT label maps, it is used with the pseudo-labeled predicted maps $p(F_S(\bfx_s)) =\argmax (F_S(\bfx_s))$  and $p(F_S(\bfx_t)) =\argmax (F_S(\bfx_t))$, respectively.

Inspired by CyCADA, several approaches tried to refine IST for the DASiS problem. \citet{MurezCVPR18ImageToImageTranslationDA} propose a method that simultaneously learns domain specific reconstruction with cycle consistency and domain agnostic feature extraction, and learn to predict the segmentation from these agnostic features. In the IST based method proposed in~\citep{ZhuECCV18PenalizingTopPerformersConservativeLossSemSegmDA} the classical cross-entropy loss is replaced by a so-called Conservative Loss that penalizes the extreme cases, -- for which performance is very good or very bad -- enabling the network to find an equilibrium between its discriminative power and its domain-invariance.

\citet{ToldoIVC20UDAMobileSemSegmCycleConsistencyFeatureAlignment} perform image-level domain adaptation with Cycle-GAN~\citep{ZhuICCV17UnpairedI2ICycleConsistentAdversarialNetworks} and feature-level adaptation with a consistency loss between the semantic maps. Furthermore, they consider as backbone a lightweight MobileNet-v2 architecture which allows the model's deployment on devices with limited computational resources such as the ones used in autonomous vehicles.

\citet{LiBMVC18SemanticAwareGradGANVirtualToRealUrbanSceneAdaption} propose a semantic-aware Grad-GAN that aims at transferring personalized styles for distinct semantic regions. This is achieved by a \textit{soft gradient-sensitive} objective for keeping semantic boundaries, and a \textit{semantic-aware discriminator} for validating the fidelity of personalized adaptions with respect to each semantic region.

The method introduced by \citet{WuECCV18DCANDualChannelWiseAlignmentNetworksUDA} jointly synthesizes images and, to preserve the spatial structure, performs segmentation by fusing channel-wise distribution alignment with semantic information in both the image generator and the segmentation network. In particular, the generator synthesizes new images \textit{on-the-fly} to appear target-like and the segmentation network refines the high level features before predicting semantic maps by leveraging feature statistics of sampled images from the target domain.

\citet{ChenCVPR19CrDoCoPixelLevelDomainTransferCrossDomainConsistency} rely on both image-level adversarial loss to learn image translation and feature-level adversarial loss to align feature distributions. Furthermore, they propose a bi-directional cross-domain consistency loss based on KL divergence, -- to provide additional supervisory signals for the network training, -- and show that this yields more accurate and consistent predictions in the target domain.

The Domain Invariant Structure Extraction (DISE) method~\citet{ChangCVPR19AllAboutStructureDASemSeg} combines image translation with the encoder-decoder based image reconstruction, where a set of shared and private encoders are used to disentangle high-level, \textit{domain-invariant} structure information from \textit{domain-specific} texture information. Domain adversarial losses and perceptual losses ensure the perceptual similarities between the translated images and their counterparts in the source or target domains. Furthermore, an adversarial loss in the output space ensures domain alignment and therefore generalization to the target.

The approach by \citet{LiCVPR19BidirectionalLearningDASemSegm} relies on \textit{bi-directional} learning, proposing to move from a sequential pipeline -- where the SiS model benefits from the image-to-image translation network -- to a closed loop, where the two modules help each other. Essentially, the idea is to propagate information from semantic segmentation back to the image transformation network as a semantic consistent regularization.

\citet{ChengICCV21DualPathLearning4DASS} consider two image translation and segmentation pipelines from opposite domains to alleviate visual inconsistencies raised by image translation and to promote each other in an interactive manner. The source path assists the target path to learn precise supervision from source data, while the target path guides the source path to generate high quality pseudo-labels for self-training the target segmentation network.
\citet{MustoBMVC20SemanticallyAdaptiveI2ITranslationDASemSegm} propose a source to target translation model guided by the source semantic map  using Spatially-Adaptive (De)normalization (SPADE)~\citep{ParkCVPR19SemanticImageSynthesiswithSpatiallyAdaptiveNormalization} and Instance Normalization layers~\citep{UlyanovX16InstanceNormalizationFastStylization}.

\citet{YangECCV20LabelDrivenReconstructionDASemSegm} introduce a reconstruction network that relies on conditional GANs, which learn to reconstruct the source or the source-like target image from their respective predicted semantic label map. Furthermore, a perceptual loss and a discriminator feature matching loss are used to enforce the semantic consistency between the reconstructed and the original image features.

Some recent works propose IST solutions that do not rely primarily on GANs for image translation. For instance, the innovative approach by~\citep{YangCVPR20FDAFourierDASemSegm} relies on the Fourier Transform and its inverse to map the target style  into that of the source images, by swapping the low-frequency component of the spectrum of the images from the two domains. The same research team proposes to exploit the phase of the Fourier transform within a consistency loss~\citep{YangCVPR20PhaseConsistentEcologicalDA}; this guarantees to have an image-to-image translation network that preserves semantics.

\begin{figure}[ttt]
\begin{center}
\includegraphics[width=.85\textwidth]{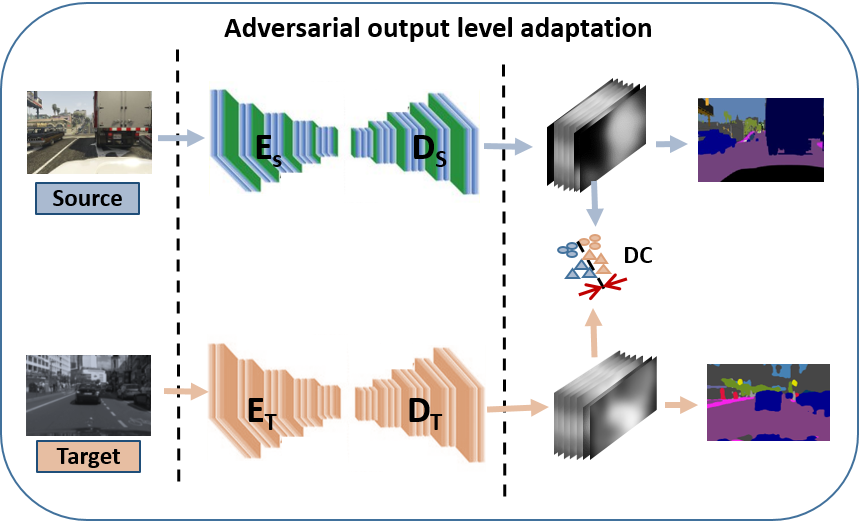}
\caption{Adversarial adaptation on the label prediction output space, where pixel-level representations are derived from the class-likelihood map and used to train the domain classifier.}
\label{fig:OutputLevelDA}
\end{center}
\end{figure}

\subsection{Output-level adaptation}
\label{sec:output-level}

To avoid the complexity of high-dimensional feature space adaptation, several papers propose to perform instead adversarial adaptation on the low-dimensional label prediction output space, -- defined by the class-likelihood maps  (see \fig{OutputLevelDA}). In this case,  the pixel-level representations corresponds to the class predictions (forming a $C$ dimensional vector), and in the derived feature space, -- similarly to the approaches described in \secref{feature-level}, -- domain confusion between the domains can be achieved by learning a corresponding domain discriminator. 
Such adversarial learning in the output space has been initially proposed by \citet{TsaiICCV19DAStructuredOutputDiscriminativePatchRepresentations} where they learn a discriminator to distinguish whether the segmentation predictions come from the source or from the target domain. To make the model adaptation more efficient, auxiliary pixel-level semantic and domain classifiers are added at multiple layer's of the network, and trained jointly.

\citet{VuCVPR19ADVENTAdversarialEntropyMinimizationDASemSegm} first derive the so called \textit{weighted self-information} maps (wSIM) defined as
\begin{align*}
I^{(h,w,c)}_x=-p^{(h,w,c)} \log p^{(h,w,c)}
\end{align*}
and perform adversarial adaptation on the features derived from these maps.
Furthermore, they show that minimizing the sum of these wSIMs is equivalent to direct entropy minimization and train the model jointly with these two complementary entropy-based losses (the direct entropy and the corresponding adversarial loss). \citet{PanCVPR20UnsupervisedIntraDASemSegmSelfSupervision} instead train a domain classifier on the entropy maps 
$E^{(h,w)}_x=-\sum_c I^{(h,w,c)}_x$ to reduce the distribution shift between the source and target data. 

Output level adversarial learning has often been used in combination with image-level style transfer and self-training~\citep{ChangCVPR19AllAboutStructureDASemSeg,LiCVPR19BidirectionalLearningDASemSegm,WangICCV21UncertaintyAwarePseudoLabelRefineryDASS} and curriculum learning~\citet{PanCVPR20UnsupervisedIntraDASemSegmSelfSupervision} (see also Table~\ref{tab:semsegm_methods}).

\section{Complementary Techniques}
\label{sec:dass_improvements}

In the previous section, we mainly focused on the core issue 
of domain alignment; in this section, we discuss other techniques 
that can be coupled with  the DASiS methods previously presented. Generally, they are not explicitly focused on domain alignment, but rather on improving the segmentation model accuracy on the target data. As a part of the transfer learning, DASiS -- and UDA in general -- possesses characteristics (domain separation, unlabeled target instances, etc.) that encourage researchers to integrate techniques from ensemble, semi- and self-supervised learning, often resulting in their mutual empowering. While being extensively used in UDA research, the methodologies detailed below
originated from other branches of machine learning; for example, {\it self-training with pseudo-labels}~\citep{LeeICMLW13PseudoLabelSimpleEfficientSSLMethodDNN} and {\it entropy minimization}~\citep{GrandvaletNIPS04SemiSupervisedLearningEntropyMinimization}
have been originally formulated for semi-supervised learning; {\it curriculum learning} has been devised as a stand-alone training paradigm~\citep{BengioICML09CurriculumLearning}; {\it model distillation} and {\it self-ensembling} are recent deep learning techniques that allow training more accurate models.

\begin{figure}[ttt]
\begin{center}
\includegraphics[width=0.85\textwidth]{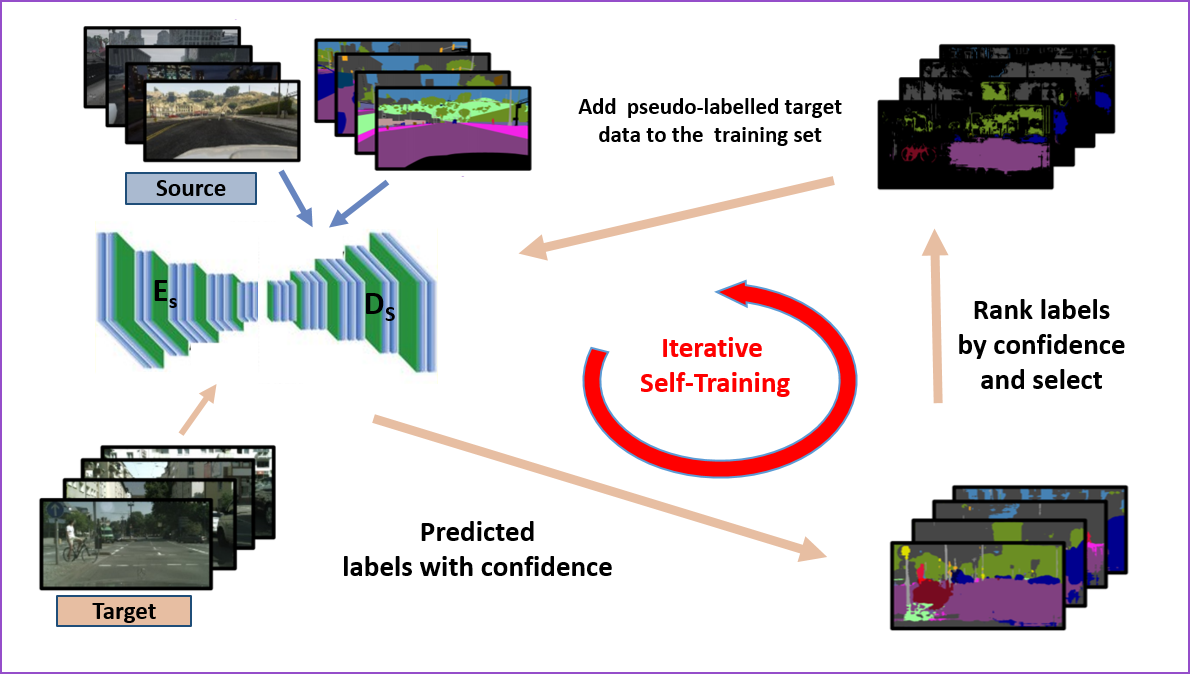}
\caption{Pseudo-labeling. Originated from semi-supervised learning, the idea is to generate pseudo-labels for the target data and to refine the model over iterations, by using the most confident labels from the target set.}
\label{fig:SelfTrainingSS}
\end{center}
\end{figure}

\subsection{Pseudo-labelling and self-training (SelfT)}

Originated from semi-supervised learning, the idea is to generate pseudo-labels for the target data and to refine (self-train) the model over iterations, by using the most confident labels from the target set~\citep{LiCVPR19BidirectionalLearningDASemSegm,ZouECCV18UnsupervisedDASemSegmClassBalancedSelfTraining,KimCVPR20LearningTextureInvariantRepresentationDASemSegm,LiCVPR19BidirectionalLearningDASemSegm} (see illustration in~\fig{SelfTrainingSS}). Indeed, pseudo-labels are often error-prone, so it is important to select the most reliable ones and to progressively increase the set of pseudo-labels as the training progresses. To this end, different works have been proposed that extrapolate the pseudo-label confidence relying on the maximum class probability (MCP) of the model output ~\citep{LiECCV20ContentConsistentMatchingDASemSegm,WangCVPR20DifferentialTreatmentStuffThingsDASemSegm,ZouECCV18UnsupervisedDASemSegmClassBalancedSelfTraining} or on the entropy of the softmax predictions~\citep{SaportaCVPRWS20ESLEntropyGuidedSelfSupDASiS}.
Pseudo-labels for which the prediction is above a certain threshold are assumed to be reliable; vice-versa, values below the threshold are not trusted. As shown in Table~\ref{tab:semsegm_methods}, self-training is one of the most popular complementary methods combined with domain alignment techniques.

In the following, we list a few of such methods with their particularities, aimed at further improving the effectiveness of self-learning. To avoid the gradual dominance of large classes on pseudo-label generation, \citet{ZouECCV18UnsupervisedDASemSegmClassBalancedSelfTraining} propose a class balanced self-training framework and introduce spatial priors to refine the generated labels. \citet{ChenICCV17NoMoreDiscriminationCrossCityAdaptationRoadSceneSegmenters} rely on static-object priors estimated for the city of interest by harvesting the Time-Machine of Google Street View to improve the soft pseudo-labels required by the proposed class-wise adversarial domain alignment. 
\citet{KimCVPR20LearningTextureInvariantRepresentationDASemSegm} introduce a texture-invariant pre-training phase; in particular, the method relies on image-to-image translation to learn a better performing model at a first stage, which is then adapted via the pseudo-labeling in a second stage.

In order to regularize the training procedure, \citet{ZhengIJCAI20UnsupervisedSceneAdaptationwithMemoryRegularizationInVivo}
average predictions of two different sets from the same network. Relatedly, \citet{ShenX19RegularizingProxiesMultiAdversarialTrainingUDASS} combine the output of different discriminators with the confidence of a segmentation classifier, in order to increase the reliability of the pseudo-labels. To rectify the pseudo-labels, \citet{ZhengIJCV21RectifyingPseudoLabelLearningUncertaintyEstimationDASS} propose to explicitly estimate the prediction uncertainty during training. They model the uncertainty via the prediction variance and integrate the uncertainty into the optimization objective.

\citet{CorbierePAMI21ConfidenceEstimationViaAuxiliaryModels} propose an auxiliary network to estimate the \textit{true-class probability map} for semantic segmentation and integrate it into an adversarial learning framework to cope with the fact that the predicted true-class probabilities might suffer from the domain shift. The confidence branch has a multi-scale architecture based on ASPP,  allowing the network to better cope with  semantic regions of variable size in the image.  

Differently, \citet{MeiX20InstanceAdaptiveSelfTrainingUDA} propose an \textit{instance adaptive}  framework where pseudo-labels are generated via an \textit{adaptive selector}, namely a confidence-based selection strategy with a confidence threshold that is adaptively updated throughout training. Regularization techniques are also used to respectively smooth and sharpen the pseudo-labeled and non-pseudo-labeled regions.

In order to make the self-training less sensitive to incorrect pseudo-labels,
\citet{ZouICCV19SConfidenceRegularizedSelfTraining} rely on \textit{soft} pseudo-labels in the model regularization, forcing the network output to be smooth. \citet{ShinECCV20TwoPhasePseudoLabelDensificationDA} propose a pseudo-label densification framework where a sliding window voting scheme is used to propagate confident neighbor predictions. In a second phase, a confidence-based easy-hard classifier selects images for self-training, while a hard-to-easy adversarial learning pushes hard samples to be like easy ones. 

\citet{ZhangNIPS19CategoryAnchorGuidedUDASS} propose a strategy where pseudo-labels are used in both a cross-entropy loss and a \textit{category-wise distance loss}, where class-dependent centroids are used to assign pseudo-labels to training samples.
\citet{LiECCV20ContentConsistentMatchingDASemSegm} select source images that are most similar to the target ones  via \textit{semantic layout matching} and to retain some pixels for the adaptation via \textit{pixel-wise similarity matching}. These pixels are used together with pseudo-labeled target samples to refine the model. Furthermore, entropy regularization is imposed on all the source and target images. 

To mitigate low-quality pseudo-labels arising from the domain shift, \citet{TranhedenWACV21DACSDACrossDomainMixedSampling} propose to \textit{mix} images from the two domains along with the corresponding labels and pseudo-labels. While training the model, they enforce consistency between predictions of images in the target domain and images mixed across domains.

\citet{GuoCVPR21MetaCorrectionDomainAwareMetaLossCorrectionDASS} propose to improve the reliability of pseudo-labels via a \textit{meta-correction}  framework; they model the noise distribution of the pseudo-labels by introducing a \textit{noise transaction matrix} that encodes inter-class noise transition relationship. The meta-correction loss is further exploited to improve the pseudo-labels via a meta-learning strategy to adaptively distill knowledge from all samples during the self-training process.

Alternatively, pseudo-labels can also be used to improve the model without necessarily using them in a self-training cross-entropy loss. For example, \citet{WangCVPR20DifferentialTreatmentStuffThingsDASemSegm} use pseudo-labels to disentangle source and target features by taking into account regions associated with~\textit{things} and \textit{stuff}~\citep{CaesarCVPR18COCOStuffThingsAndStuffClassesContext}.

\citet{DuICCV19SSFDANSeparatedSemanticFeatureDASemSegm} use separate semantic features according to the downsampled pseudo-labels to build \textit{class-wise confidence map} needed to reweigh the adversarial loss. A progressive confidence strategy is used to obtain reliable pseudo-labels and, in turn, class-wise confidence maps.

\subsection{Entropy minimization of target predictions (TEM)}

Originally devised for semi-supervised learning~\citep{GrandvaletNIPS04SemiSupervisedLearningEntropyMinimization},  entropy minimization has received a broad recognition as an alternative or complementary technique for domain alignment. Different DASiS/UDA methods extend simple entropy minimization on the target data by applying it jointly with adversarial losses~\citep{DuICCV19SSFDANSeparatedSemanticFeatureDASemSegm,VuCVPR19ADVENTAdversarialEntropyMinimizationDASemSegm} or square losses~\citep{ChenICCV19DASemSegmentationMaximumSquaresLoss,ToldoWACV21UDASemSegmOrthogonalClusteredEmbeddings}.

\citet{VuCVPR19ADVENTAdversarialEntropyMinimizationDASemSegm}
propose to enforce structural consistency across domains by minimizing both the conditional entropy of pixel-wise predictions and an adversarial loss that ensures the distribution matching in terms of weighted entropy maps (as discussed in \secref{output-level}). The main advantage of their approach is that computation of the pixel-wise entropy does not depend on any network and entails no overhead.

Similarly,~\citet{HuangECCV20ContextualRelationConsistentDASemSegm} design an entropy-based minimax adversarial learning scheme to align local contextual relations across domains. The model learns to enforce the prototypical local contextual relations explicitly in the feature space of a labeled source domain, while transferring them to an unlabeled target domain
via backpropagation-based adversarial learning using a Gradient Reversal Layer (GRL)~\citep{GaninJMLR16DomainAdversarialNN}.

\citet{ChenICCV19DASemSegmentationMaximumSquaresLoss} show that entropy minimization based UDA methods often suffer from the probability imbalance problem. To avoid the adaptation process to be dominated by the easiest to adapt samples, they propose instead a \textit{class-balanced weighted} maximum squares loss with a linear growth gradient.
Furthermore, they extend the model with self-training on low-level features guided by pseudo-labels obtained by averaging the output map at different levels of the network.  \citet{ToldoWACV21UDASemSegmOrthogonalClusteredEmbeddings} integrate this image-wise class-balanced entropy-minimization loss to regularize their feature clustering-based DASiS method. To further enhance the discriminative clustering performance,  they introduce an \textit{orthogonality loss} -- which force individual representations 
to be orthogonal, -- and a \textit{sparsity loss} to reduce class-wise the number of active feature channels.

The Bijective Maximum Likelihood (BiMaL) loss~\citet{TruongICCV21BiMaLBijectiveMaximumLikelihoodDASS} is a generalized form of the adversarial entropy minimization, without any assumption about pixel independence. The BiMaL loss is formed using a maximum-likelihood formulation to model the global structure of a segmentation input,  and a bijective function,  to map that segmentation structure to a deep latent space. Additionally, an \textit{unaligned domain score}  is introduced to measure the efficiency of the learned model on a target domain in an unsupervised fashion.

\subsection{Curriculum learning (CurrL)}
\label{sec:currDA}

Several papers apply {\it curriculum} strategy to DASiS; the main idea is to apply  simpler, intermediate tasks to determine certain properties of the target domain which allow to improve performance on the main segmentation task.
In this regard,
\citet{ZhangPAMI20CurriculumDASemSegmUrbanScenes} 
propose to use \textit{image-level label distribution} to guide the pixel-level target segmentation. Furthermore, they  use the \textit{label distributions of anchor super-pixels} to indicate the network where to update. Learning these easier tasks allows to improve the predicted pseudo-labels for the target samples and therefore can be used to effectively regularize the fine-tuning of the SiS network.

Similarly,~\citet{SakaridisICCV19GuidedCurriculumModelAdaptationSemSegm} propose a curriculum learning approach where models are adapted from \textit{day-to-night} learning with progressively increasing the level of darkness. They exploit the correspondences of images captured across different daytime to improve pixel predictions at inference time.  
\citet{LianICCV19ConstructingSelfMotivatedPyramidCurriculumsCrossDomainSemSegm} adopt the \textit{easy-to-hard} curriculum learning approach by predicting labels first at image level, then at region level and finally at pixel level (the main task).

To further improve model performance in the target domain, \citet{PanCVPR20UnsupervisedIntraDASemSegmSelfSupervision}, 
separate the target data into \textit{easy} and \textit{hard} samples -- relying on the entropy -- and try to diminish the gap between those predictions by so called intra-domain adversarial training on the corresponding entropy maps (see also \secref{output-level}).

\subsection{Co-training (CoT)}
\label{sec:cotraining}

Another set of UDA/DASiS methods has been inspired by {\it co-training}~\citep{ZhouTKDE05TriTrainingExploitingUnlabeledData} where the idea is to have two distinct classifiers enforced to be diverse, in order to capture different views of the data while predicting the same labels. The main idea behind such methods is that \textit{diversifying the classifiers} in terms of learned parameters -- while at the same time \textit{maximizing the consensus} on their predictions -- will encourage the model to output more discriminative feature maps for the target domain. The
rationale is that the target samples near the class boundaries are likely to be misclassified by the source classifier and using the disagreement of two classifiers on the prediction for target samples can implicitly detect such cases and, in turn, 
improve the class boundaries.

The first such UDA model maximizing the classifier discrepancy (MCD) has been proposed by \citet{SaitoCVPR18MaximumClassifierDiscrepancyUDA} where the adversarial model alternates between 1) maximizing the discrepancy between two classifiers on the target sample while keeping the feature generator fixed and 2) training the feature encoder to minimize discrepancy while keeping the classifiers fixed.
As an alternative, to encourage the encoder to output more discriminative features for the target domain, \citet{SaitoICLR18AdversarialDropoutRegularization} rely on adversarial dropout and \citet{LuoCVPR19TakingACloserLookDomainShiftSemanticsConsistentDA} enforce the weights of the two classifiers to be diverse while using self-adaptive weights in the adversarial loss to improve local semantic consistency.
Finally, \citet{LeeCVPR19SlicedWassersteinDiscrepancyUDA} consider the sliced Wasserstein discrepancy to capture the dissimilarity between the predicted probability measures that provides a geometrically meaningful guidance to detect target samples that lie far from the support of the source.

\subsection{Self-ensembling}
Another popular method for semi-supervised learning is to use an ensemble of models and to exploit the consistency between predictions under some perturbations.
While \citet{LaineX16TemporalEnsemblingSemisupervisedLearning} propose the temporal ensembling by taking the per-sample moving average of predictions, \citet{TarvainenNIPS17MeanTeachersBetterRoleModelsSSL} replace the averaging predictions with an \textit{exponential moving average} (EMA) of the model weights. In the latter case  the Mean Teacher framework is used, represented by a second, non-trainable model whose weights are updated with the EMA over the actual trainable weights.

\begin{figure}[ttt]
\begin{center}
\includegraphics[width=0.95\textwidth]{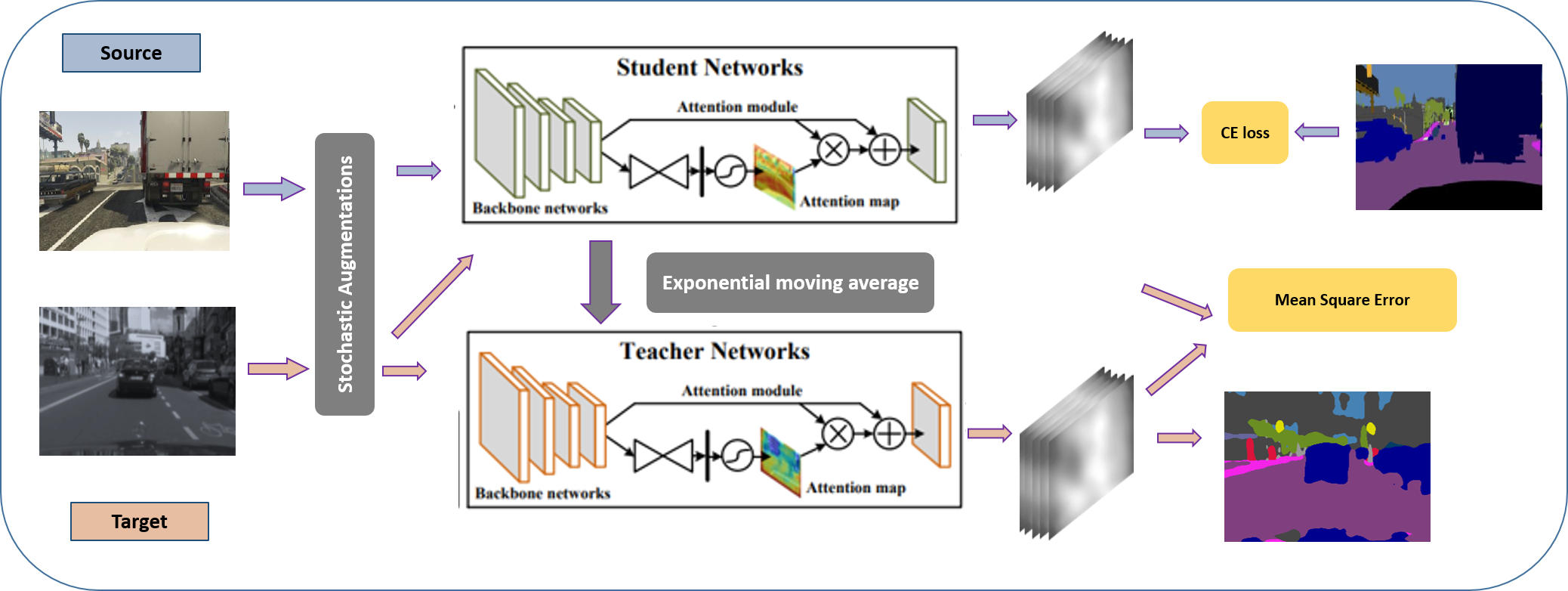}
\caption{The self-ensembling attention network~\citep{XuAAAI19SelfEnsemblingAttentionNetworksSemSegm} consists of a student network and a teacher network. The two networks share the same architecture with an embedded attention module. The \textit{student network} is jointly optimized with a supervised segmentation loss for the source domain and an unsupervised consistency loss for the target domain. The \textit{teacher network} is excluded from the back-propagation, it is updated with an exponential moving average. In the test phase, the target-domain images are sent to the teacher network to accomplish the SiS task (Image based on \citep{XuAAAI19SelfEnsemblingAttentionNetworksSemSegm}).}
\label{fig:self_ens}
\end{center}
\end{figure}

Such self-ensembling models can also be considered for UDA and DASiS, where the model is generally composed of a \textit{teacher} and a \textit{student} network, encouraged to produce consistent predictions. The teacher is often an ensembled model that averages the student’s weights and therefore the predictions from the teacher can be interpreted as pseudo-labels for the student model. Indeed,~\citet{FrenchICLR18SelfEnsemblingForVisualDA} extend the model proposed by \citet{TarvainenNIPS17MeanTeachersBetterRoleModelsSSL} to UDA considering a separate path for source and target, and sampling independent batches making the Batch Normalization (BN)~\citep{IoffeICML15BatchNormAcceleratingDeepNetsTrain} domain specific during the training process. \citet{PeroneNI19UDAMedicalImSegmSelfEnsembling} apply self-ensembling to adapt medical image segmentation. The Self-ensembling Attention Network of \citet{XuAAAI19SelfEnsemblingAttentionNetworksSemSegm} aims at extracting attention aware features for domain adaptation (see \fig{self_ens}).

In contrast to the above mentioned ensemble models, which are effective but require heavily-tuned manual data augmentation for successful domain alignment,~\citet{ChoiICCV19SelfEnsemblingGANBasedDataAugmentationDASemSegm} propose a self-ensembling framework which deploys a target-guided GAN-based data augmentation with spectral normalization. To produce semantically accurate prediction for the source and augmented samples, a semantic consistency loss is used. More recently, \citet{WangAAAI21ConsistencyRegularizationSourceGuidedPerturbationUDASemSegm} proposed a method that relies on AdaIN~\citep{HuangICCV17ArbitraryStyleTransferRealTimeAdaptiveInstanceNormalization} to convert the style of source images into that of the target images, and vice-versa. The stylized images are exploited in a training pipeline that exploits self-training, where pseudo-labels are improved via the usage of self-ensembling.

\subsection{Model distillation}
\label{sec:distillation}

In machine learning, \textit{model distillation}~\citep{HintonX15DistillingKnowledgeNN} has been introduced as a way to transfer the knowledge from a large model to a smaller one -- for example, compressing the discriminative power of an ensemble of models into a single, lighter one. In the context of DASiS, it has been exploited to guide the learning of more powerful features for the target domain, transferring the discriminative power gained on the source samples.

\begin{figure}[ttt]
\begin{center}
\includegraphics[width=0.95\textwidth]{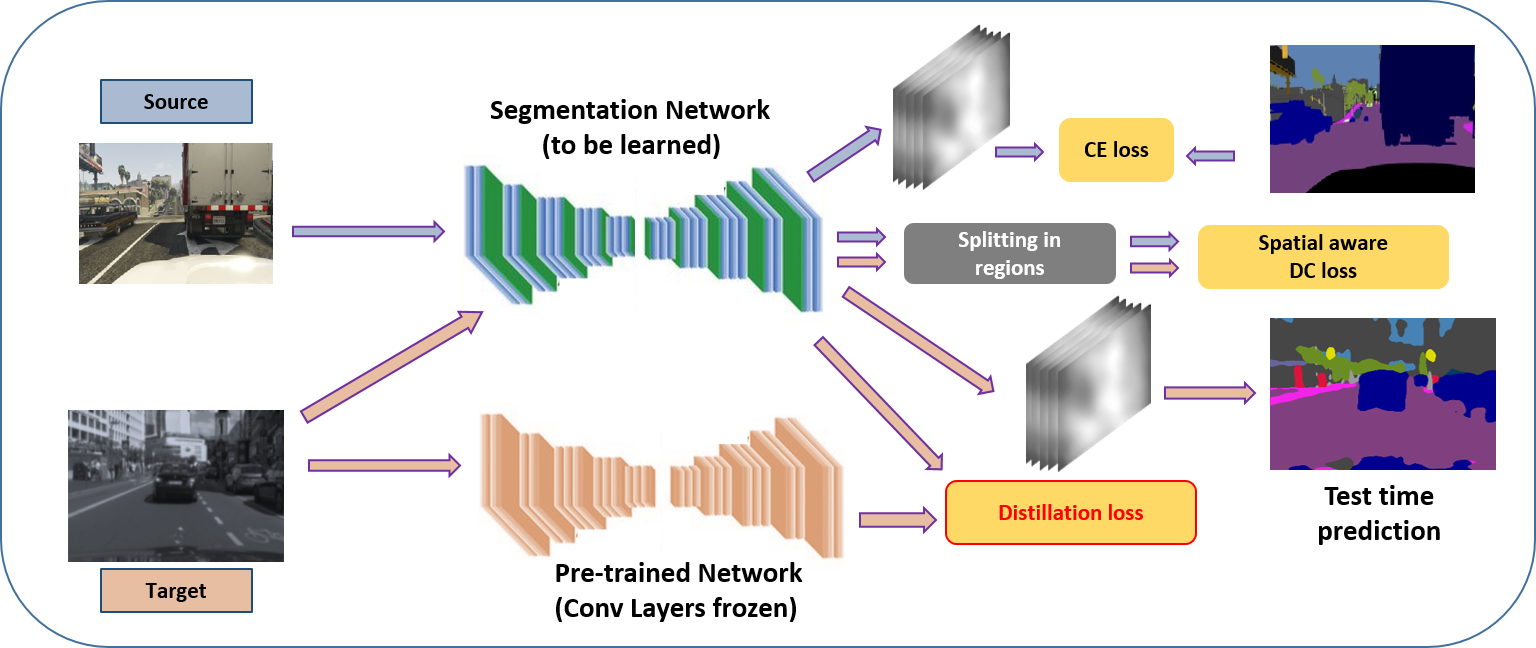}
\caption{\citet{ChenCVPR18RoadRealityOrientedDASemSegmUrbanScenes} propose to incorporate a target guided distillation module to learn the target (real) style convolutional filters from the (synthetic) source ones and to combine it with a spatial-aware distribution adaptation module 
(Figure based on~\citep{ChenCVPR18RoadRealityOrientedDASemSegmUrbanScenes}).} 
\label{fig:ROAD}
\end{center}
\end{figure}

\citet{ChenCVPR18RoadRealityOrientedDASemSegmUrbanScenes} propose to tackle the distribution alignment
in DASiS by using a distillation strategy to learn the target style convolutional filters (see \fig{ROAD}). Furthermore, taking advantage of the intrinsic spatial structure presented in urban scene images (that they focus on), they propose to partition the images into non-overlapping grids, and the domain alignment is performed on the pixel-level features from the same spatial region using GRL~\citep{GaninJMLR16DomainAdversarialNN}. 
The Domain Adaptive Knowledge Distillation model~\citep{KothandaramanWACV21DomainAdaptiveKnowledgeDistillationDrivingSceneSemSegm} consists in a multi-level strategy to effectively distill knowledge at different levels -- feature space and output space -- using a combination of KL divergence and MSE losses.

\citet{ChenX22SmoothingMattersMomentumTransformerDASIS} formalize the self-training as knowledge distillation where the target network is learned by knowledge distillation from the source teacher model. They analyze failures when adapting Swin Transformer~\citep{LiuICCV21SwinTransformerHierarchicalViTShiftedWindows} based segmentation model to new domains,
and suggest that 
these failures are due to the severe high-frequency components generated during both the pseudo-label construction and feature alignment for target domains.
As a solution, they introduce a low-pass filtering mechanism integrated into a {\it momentum network} which smooths the learning dynamics of target domain features and their pseudo labels. 
Then a dynamic adversarial training strategy is used to align the distributions, where the \textit{dynamic weights} are used to evaluate the importance of the samples. 
In a similar spirit,  \citet{HoyerCVPR22DAFormerImprovingNetworkArchitecturesTrainingStrategiesDASIS}  
exploit the strengths of the transformers in a knowledge distillation-based self-training framework. The proposed  DAFormer architecture is based on a Transformer encoder and
a context-aware fusion decoder. To overcome adaptation instability and overfitting to the source domain, they propose \textit{Rare Class Sampling}, which takes into account the long-tail distribution of the source domain. They further distill ImageNet knowledge through the \textit{Thing-Class ImageNet Feature Distance}.

\subsection{Adversarial attacks}
\label{sec:adversarial_attacks}

The aim of adversarial attacks~\citep{SzegedyICLR14IntriguingPropertiesNeuralNetworks} is \textit{to perturb} examples in a way that makes deep neural networks fail when processing them. The model trained with both clean and perturbed samples in an adversarial manner, have been shown to learn more robust models for the given task.
While the connection between adversarial robustness and generalization is not fully explained
yet~\citep{GilmerICML19AdvExamplesNaturalConsequenceTestErrNoise}, adversarial training has been successfully applied to achieve different goals than adversarial robustness; for instance, it has been used to mitigate over-fitting in supervised and semi-supervised
learning~\citep{ZhengCVPR16ImprovingRobustnessDeepNNStabilityTraining}, to tackle domain generalization tasks~\citep{VolpiNIPS19GeneralizingUnseenDomainsAdvDataAugm}, or to fill in the gap between the source and target domains by adapting the classification decision boundaries (as discussed in \secref{dass_siamese}).

Concerning adversarial attack in the case of DASiS, \citet{YangAAAI20AnAdversarialPerturbationOrientedDASS} propose pointwise perturbations to generate adversarial features that capture the vulnerability of the model -- for example  the tendency of the classifier to collapse into the classes that are more represented, in contrast with the long tail of the most under-represented ones -- and conduct adversarial training on the segmentation network to improve its robustness.

\citet{YangICCV21ExploringRobustnessDASS} study the adversarial vulnerability of existing DASiS methods and propose the adversarial self-supervision UDA, where the objective is to maximize -- by using a contrastive loss -- the proximity between clean images and their adversarial counterparts in the output space.~\citet{HuangICCV21RDARobustDAViaFourierAdversarialAttacking} propose a Fourier adversarial training method, where the pipeline is to generate adversarial samples -- by perturbing certain high frequency components that do not carry significant semantic information -- and  use them to train the model. This training technique allows reaching an area with a flat loss landscape, which yields a more robust domain adaptation model.

\subsection{Self-supervised learning}
\label{sec:SelfLearnDA}

Self-supervised learning approaches (see also~\secref{SelfLearn})
have found their place in UDA research~\citep{SunX19UDAThroughSSL,BucciPAMI21SelfSupervisedLearningAcrossDomains,XuIEEEA19SelfSupervisedDAForCVTasks}. For what concerns DASiS,
\citet{AraslanovCVPR21SelfSupervisedAugmentationConsistencyAdaptingSemSegm} propose a lightweight self-supervised training scheme, where the consistency of the semantic predictions across image transformations such as photometric noise, mirroring and scaling is ensured.
The model is trained end-to-end using co-evolving pseudo-labels -- using a momentum network, which is a copy of the original model that evolves slowly -- and maintaining an exponentially moving class prior. The latter is used to discount the confidence thresholds for classes with few samples, in order to  increase their relative contribution to the training loss. 

Similarly, \citet{YangICCV21ExploringRobustnessDASS} -- as mentioned in the previous paragraph -- exploit self-supervision in DASiS by minimizing the distance between clean and adversarial samples in the output space via a contrastive loss.

\section{Beyond Classical DASiS}\label{sec:beyond_dass}

\begin{figure*}[ttt]
\begin{center}
\includegraphics[width=0.9\textwidth]{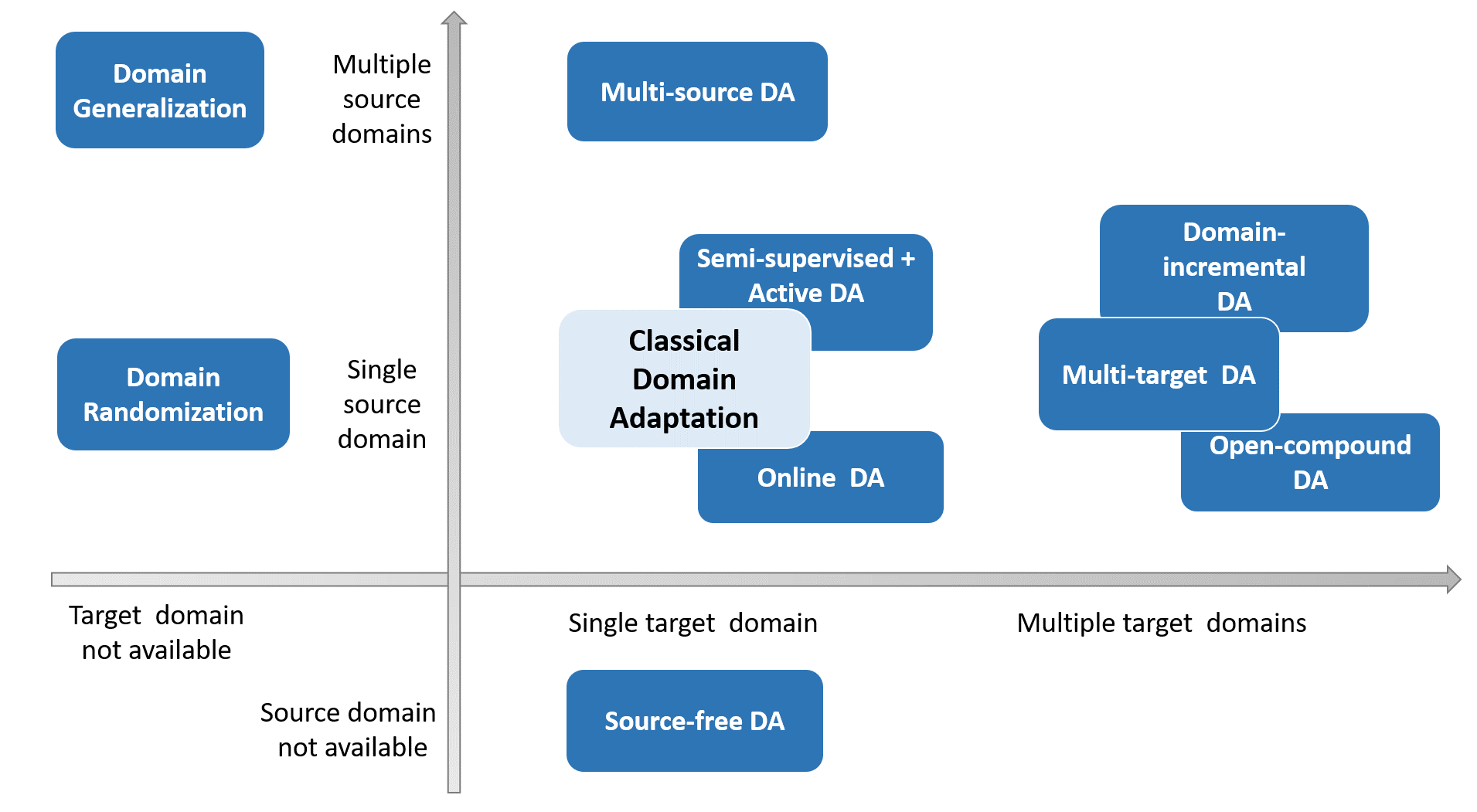}
\caption{Illustration of different scenarios based on source and target data availability.} 
\label{fig:beyondDA}
\end{center}
\end{figure*}

Typical DASiS methods assume that both source and target domains
consist of samples drawn from single data distributions, both available, and that there is a shift between the two distributions. Yet, these assumptions may not hold in the real world and therefore several methods have been proposed that tackle specific problem formulations where some of these assumptions are relaxed or additional  constraints added (see~\fig{beyondDA} for an illustration of different scenarios related to different data availability assumptions).

For instance, in {\it multi-source} domain adaptation (MSDA) the goal is learning from an arbitrary number of source domains (\secref{msdass}), and in {\it multi-target} domain adaptation (MTDA) the aim is to learn from a single source for several unlabeled target domains simultaneously (\secref{mtdass}).  Instead of having a well defined set of target domains, one can address the problem of a target distribution which is assumed to be a compound of multiple, unknown, homogeneous domains  (\secref{OpenCompoundDA}).  

Alternatively to adapting the model simultaneously to several new target domains, the learning can be done incrementally when the access to new domains is in a  sequential manner (\secref{domain-inc}), or considering a single  target domain, but the access to the target data is continuous and online  (\secref{oasis}).
One could  make the assumption that the source model is available, but the source data on which it was trained on is not -- the \textit{source-free domain adaptation} problem (\secref{SFDA});  

Another scenario is {\it domain generalization}, where the model learns from one or multiple source domains, but has not access to any target sample, nor hints on the target distribution (\secref{dg}). On the other end, different methods tackle the semi-supervised domain adaptation problem, where one even assumes that a few target samples are annotated (\secref{SSDA}), or can be actively annotated (\secref{act-DASS}).

Besides the number of domains and the amount of labeled/unlabeled samples available in the source/target domains, another important axis of variation for domain adaptation strategies is the overlap between source and target labels. Indeed, the class of semantic labels in the source and the target domains is not necessarily the same and, therefore, several methods have been proposed that address this issue (\secref{universal}).

\subsection{Multi-source DASiS}
\label{sec:msdass}

The simplest way to exploit multiple source domains is to combine all the available data in a single source and train a classical UDA model. While this can, in some cases,  provide a reasonable baseline, in other cases, it might yield poor results. This can be due to 1) the fact that there are several data distributions mixed in the combined source, making the adaptation process more difficult if this is not explicitly handled, and 2) in many cases this solution might yield to strong negative transfer as shown in \citep{MansourNIPS09DAMultipleSources}.
Alternatively, one can consider a weighted combination of multiple source domains for which theoretical analysis of error bounds has been proposed in~\citep{BenDavidML10ATheoryLearningDifferentDomains,CrammerJMLR08LearningFromMultipleSources}.  A such  algorithm with strong theoretical guarantees was propose by \citet{HoffmanNIPS18AlgorithmsTheoryMultipleSourceAdaptation}, where they  design a  distribution-weighted combination for the cross-entropy loss and other similar losses. \citet{CortesICML21ADiscriminativeTechniqueMSDA} propose instead a discriminative method which only needs conditional probabilities -- that can be accurately estimated  for the unlabeled target data, -- relying only on the access to the source predictors and not the labeled source data.~\citet{RussoICIAP19TowardsMultiSourceAdaptiveSemSegm} extend adversarial DASiS to deal with multiple sources and investigate such baselines, \ie comparing models trained on the union of the source domains versus weighted combination of adaptive adversarial models trained on individual source-target pairs.

Further methods proposed for image classification \citep{LiNIPS18ExtractingRelationshipByMultiDomainMatching,PengICCV19MomentMatchingMultiSourceDA,PengECCV20Domain2VecDomainEmbeddingUDA,YangECCV20CurriculumManagerForSourceSelectionMSDA,ZhaoNIPS18AdversarialMultipleSourceDA,ZhaoAAAI20MultiSourceDistillingDA,ZhouX20DomainAdaptiveEnsembleLearning,ZhuAAAI19AligningDomainSpecificDistributionMSDA,NguyenICCV21STEMMultiSourceDAWithGuarantees} show that when the relationship between different source domains is appropriately exploited, it is possible to train a target model to perform significantly better than using just the union of source data or a weighted combination of individual models' outputs. These deep multi-source DA (MSDA) approaches have often focused on learning a common domain-invariant feature extractor that achieves a small error on several source domains,  hoping that such representation  can generalize well to the target domain.

\begin{figure}[ttt]
\begin{center}
\includegraphics[width=0.95\textwidth]{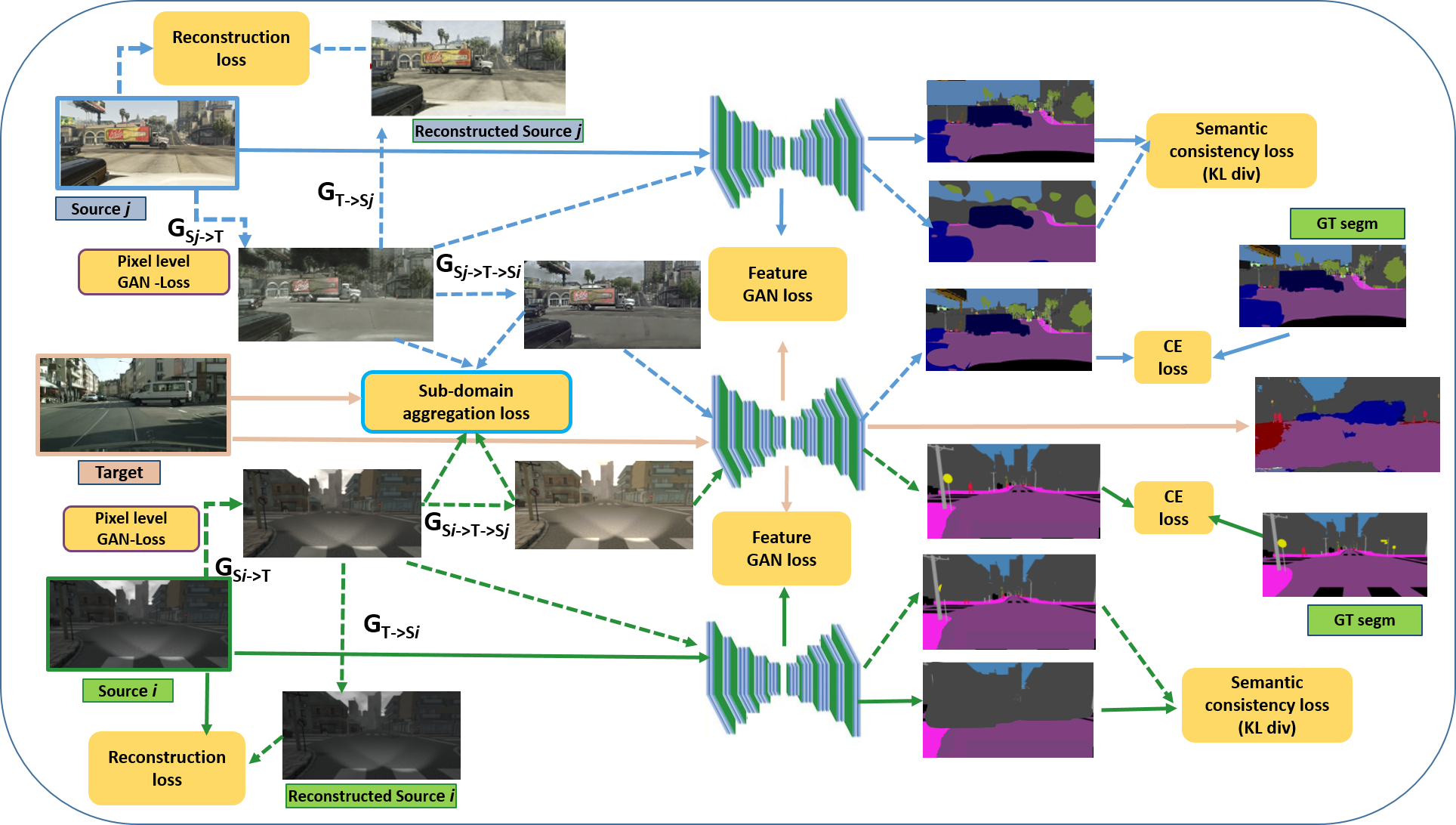}
\caption{The multi-source DASiS framework proposed by~\citet{ZhaoNIPS19MultiSourceDASemSegm}
consists in adversarial domain aggregation with two kinds of discriminators: a sub-domain aggregation discriminator, which is designed to directly make the different adapted domains indistinguishable, and a cross-domain cycle discriminator, discriminates between the images from each source and the images transferred from other sources (Figure based on \citep{ZhaoNIPS19MultiSourceDASemSegm}).}
\label{fig:MSDASiS}
\end{center}
\end{figure}

Inspired by these approaches, several methods have been proposed 
that extend MSDA solutions from classification to semantic image segmentation.  
As such, the Multi-source Adversarial Domain Aggregation Network~\citep{ZhaoNIPS19MultiSourceDASemSegm}
extends \citep{ZhaoNIPS18AdversarialMultipleSourceDA} by combining it with CyCADA~\citep{HoffmanICML18CyCADACycleConsistentAdversarialDA}. The model, trained end-to-end, generates for each source an adapted style transferred domain with dynamic semantic consistency loss between the source predictions of a pre-trained segmentation model and the adapted predictions of a dynamic segmentation model. To make these adapted domains indistinguishable, a \textit{sub-domain aggregation discriminator} and a \textit{cross-domain cycle discriminator} is learned in an adversarial manner (see~\fig{MSDASiS}).

Similarly, \citet{TasarCWPRWS20MSDASemSegmSatelliteImagesDataStandardization} propose  StandardGAN, a data standardization technique based on GANs (style transfer) for satellite image segmentation, whose goal is to standardize the visual appearance of the source and the target domain with adaptive instance normalization (AdIN)~\citep{HuangICCV17ArbitraryStyleTransferRealTimeAdaptiveInstanceNormalization} and Least-square GAN~\citep{MaoICCV17LeastSquaresGAN}
to effectively process target samples. Then, they extend the single-source StandardGAN to multi-source by multi-task learning where an auxiliary classifier is added on top of the discriminator.

In contrast, 
\citet{HeCVPR21MSDACollaborativeLearningSemSegm} propose a collaborative learning approach. They first translate source domain images to the target style by aligning the different distributions to the target domain in the LAB color space. Then, the SiS network for each source is trained in a supervised fashion by relying on the GT annotations and additional soft supervision coming from other models trained on a different source domains. Finally, the segmentation models associated with different sources collaborate with each other to generate more reliable pseudo-labels for the target domain,  used to refine the models.

\citet{GongICCV21mDALUMuSDALabelUnificationWithPartialDatasets}  consider the case where the aim is to learn from different source datasets with potentially different class sets, and formulate the task as a multi-source domain adaptation with \textit{label unification}. To approach this, they propose a two-step solution: first, the knowledge is transferred form the multiple sources to the target; second, a unified label space is created by exploiting pseudo-labels, and the knowledge is further transferred to this representation space. To address  -- in the first step -- the risk of making confident predictions for unlabeled samples in the source domains,
three novel modules are proposed: \textit{domain attention}, \textit{uncertainty maximization} and \textit{attention-guided adversarial alignment}.

\subsection{Multi-target DASiS}
\label{sec:mtdass}

In multi-target domain adaptation (MTDA) the goal is to learn from a single labeled source domain 
with the aim of performing well on multiple target domains at the same time. To tackle MTDA
within an image classification context, standard UDA approaches were directly extended to multiple targets~\citep{GholamiTIP20UnsupervisedMultiTargetDAInformationTheoreticApproach,ChenCVPR19BlendingTargetDAAdversarialMetaAdaptationNetworks,RoyCVPR21CurriculumGraphCoTeachingMTDA,NguyenMeidineWACV21UnsupervisedMultiTargetDAKnowledgeDistillation}.

Within the DASiS context, a different path was taken. 
\citet{IsobeCVPR21MTDAWithCollaborativeConsistencyLearning} propose 
to train an expert model for every target domain where the models are encouraged to \textit{collaborate via style transfer}. 
Such expert models are further exploited as teachers for a common student model that learns to imitate their output and serves as regularizer to bring the different experts closer to each other in the learned feature space. Instead, \citet{SaportaICCV21MultiTargetAdversarialFrameworksForDASS} propose to combine for each target domain $\calT_i$ two adversarial pipelines: one that learns to discriminate between the domain $\calT_i$ and the source, and one 
between $\calT_i$ and the union of the other target domains. Then, to reduce the instability that the multi-discriminator model training might cause, they propose a multi-target knowledge transfer by adopting a \textit{multi-teacher/single-student distillation mechanism}, which leads to a model that is agnostic to the target domains.

\subsection{Open-compound DASiS}
\label{sec:OpenCompoundDA}

\begin{figure}[ttt]
\begin{center}
\includegraphics[width=0.9\textwidth]{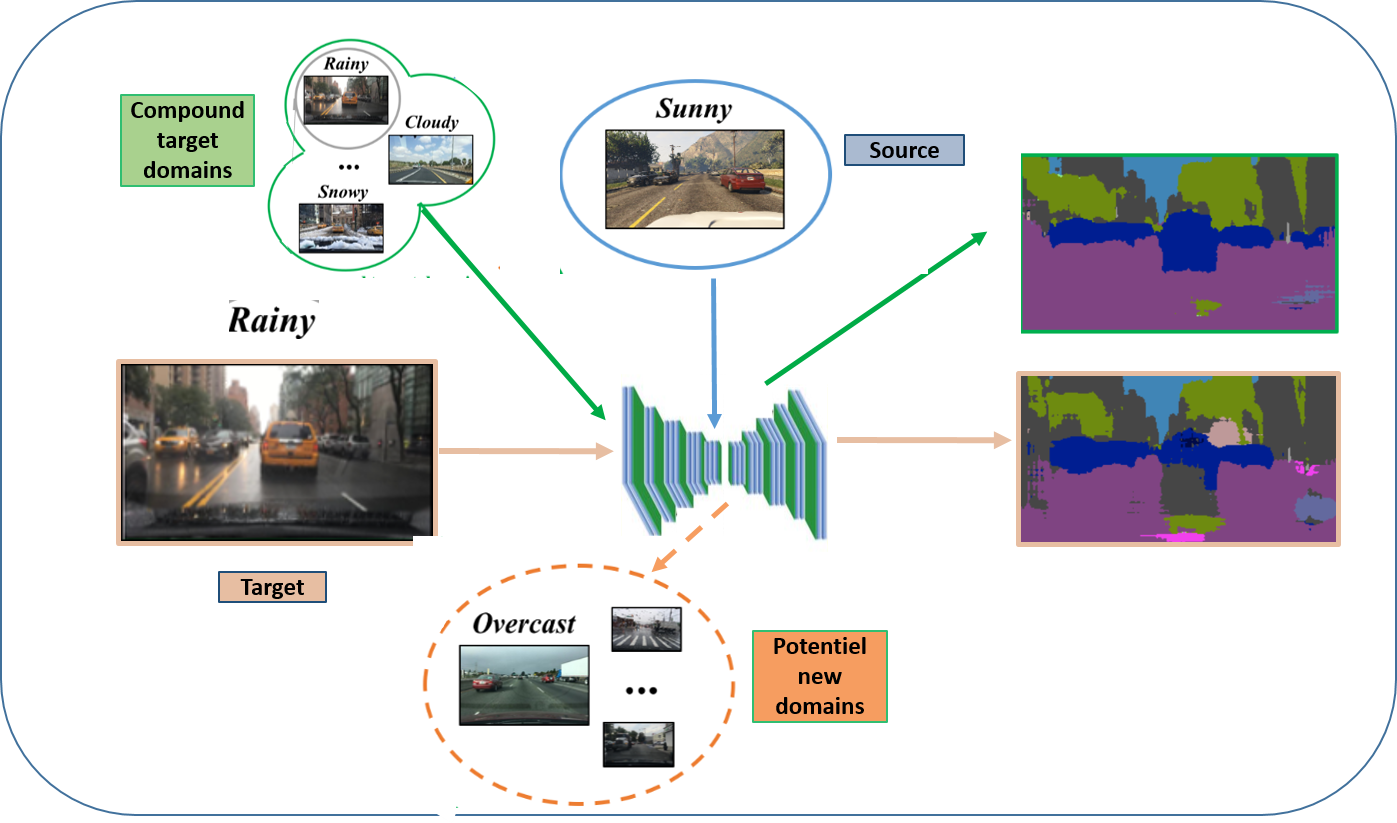}
\caption{In open-compound domain adaptation (OCDA) setting, the target distribution is assumed to be a compound of multiple, unknown, homogeneous domains 
(Figure based on~\citep{LiuCVPR20OpenCompoundDA}).}
\label{fig:OpenCompoundDA}
\end{center}
\end{figure}

The possibility of having multiple target domains is also addressed in the \textit{open-compound domain adaptation} (OCDA) setting, where the target distribution is assumed to be a compound of multiple, unknown, homogeneous domains (see \fig{OpenCompoundDA}). To face this problem, \citet{LiuCVPR20OpenCompoundDA} rely on a \textit{curriculum adaptive strategy}, where they schedule the learning of unlabeled instances in the compound target domain according to their individual gaps to the labeled source domain, approaching an incrementally harder and harder domain adaptation problem until the entire target domain is covered. The purpose is to learn a network that maintains its discriminative leverage on the classification or segmentation task at hand, while at the same time learning more robust features for the whole compound domain. To further prepare the model for open domains during inference, a \textit{memory module} is adopted to effectively augment the representations of an input when it is far away from the source. 

In contrast,~\citet{GongCVPR21ClusterSplitFuseUpdateOpenCompoundDASS}
propose a \textit{meta-learning-based framework} to approach OCDA.
First, the target domain is clustered in an unsupervised manner into multiple sub-domains by image styles; 
then, different sub-target domains are split into independent branches, for which domain-specific BN parameters are learned 
as in \citep{ChangCVPR19DomainSpecificBNforUDA}.  Finally, a meta-learner is deployed to learn to fuse sub-target domain-specific predictions, conditioned upon the style code, which  
is updated online by using the model-agnostic meta-learning 
algorithm that further improves its generalization ability.

\subsection{Domain-incremental SiS}
\label{sec:domain-inc}

\textit{Domain incremental learning} is a branch of continual learning, where the goal is extending the underlying knowledge of a machine learning system to new domains, in a sequence of different stages. These incremental learning stages can be either supervised or unsupervised, according to the available annotations. 

For what concerns general solutions under the assumption of \textit{supervised} adaptation,  where all data is labeled, a method, not specifically designed but successfully tested on SiS, was proposed by
\citet{VolpiCVPR21ContinualAdaptVisualReprDomainRandMetaLearn}. 
In order to learn visual representations that are robust against catastrophic forgetting, they propose a meta-learning solution where artificial meta-domains are crafted by relying on domain randomization techniques and they are exploited to learn models that are more robust when transferred to new conditions. The model can benefit from such solution also when only  a few samples are stored under the form of an episodic memory instead of access.

For what concerns methods designed \textit{ad hoc} for SiS, different works consider the problem of
learning different domains over the lifespan of a model~\citep{WuICCV19ACEAdaptChangingEnvironmentsSemSegm,PoravITSC19DontWorryAboutWeatherUnsupervisedConditionDependentDA,GargWACV22MultiDomainIncrementalLearningforSemSegm}. They face the domain-incremental problem by assuming that data from new domains come unlabeled -- and, therefore, they are more connected to the DASiS literature where the typical task is \textit{unsupervised} DA

\citet{WuICCV19ACEAdaptChangingEnvironmentsSemSegm} propose to generate data that resembles that of the current target domain and to update the model's parameters relying on such
samples. They further propose a \textit{memory bank} to store some domain-specific feature statistics, in order to quickly \textit{restore} domain-specific performance in case the need arises. This is done by deploying the model on a previously explored domain, on which the model has been previously adapted already.

\citet{PoravITSC19DontWorryAboutWeatherUnsupervisedConditionDependentDA} rely on a series of \textit{input adapters} to convert the images processed by the computer vision model when they come from a domain that significantly differ from the source one. They build their method by using GANs, and the proposed approach does not require domain-specific fine-tuning. 
Instead, \citet{GargWACV22MultiDomainIncrementalLearningforSemSegm}
learn domain-specific parameters for each new domain -- in their case corresponding to different geographical regions -- whereas other parameters are assumed to be domain-invariant.

\subsection{Online DASiS}
\label{sec:oasis}

In online learning~\citep{CesaBianchiB06PredictionLearningGames}, the goal is taking decisions and improving the underlying knowledge of the model sample by sample -- in contrast with offline learning, where typically one can process huge amount of data over multiple epochs.
The problem of \textit{online adaptation}, intimately connected to online learning, is essentially that of performing UDA as new samples arrive, in order to better perform on them. 
This problem has been recently re-branded as \textit{test-time adaptation}~\citep{SunICML20TestTimeTrainingSelfSupervisionGeneralizDistr,WangICLR21TENTFullyTestTimeAdaptEntropyMin,SchneiderNeurIPS2020ImprovingRobustnessCommonCorruptionsCovShiftAdapt}, where the main focus has been mainly on image
classification.

While online adaptation has been addressed in the case of  object detection more than a decade
ago~\citep{RothCVPR09ClassifierGridsForRobustAdaptObjectDetection}, it has been addressed only recently for SiS. In this context, \citet{VolpiCVPR22OnRoadOnlineAdaptSiS} propose a benchmark to tackle the problem of online adaptation of SiS models (the OASIS benchmark) where the goal is to adapt pre-trained models to new, unseen domains, in a frame-by-frame fashion, 
Such domain shifts can be  adversarial weather conditions met by an autonomous car (see examples  in the ACDC dataset~\citep{SakaridisICCV21ACDCAdverseConditionsDatasetSIS} and \secref{DAbench}).

Different approaches from the continual learning and the test-time adaptation literature have been tailored in \citep{VolpiCVPR22OnRoadOnlineAdaptSiS}
to face this problem, and empirically shown to be helpful, in particular, self-training via pseudo-labels~\citep{LeeICMLW13PseudoLabelSimpleEfficientSSLMethodDNN} and the application of the TENT algorithm~\citep{WangICLR21TENTFullyTestTimeAdaptEntropyMin}.
While adapting the model frame by frame, the main challenge is avoiding  catastrophic forgetting of the pre-trained model that is adapted to the new sequences. Therefore  two solutions are proposed to tackle this problem, The first one is  \textit{experience replay} where 
the test-time adaptation objective is regularized by optimizing a loss with respect to the original, labeled training samples. The second solution is  a reset strategy that allows resetting the model to its original weights when catastrophic forgetting is detected. 

Concurrently,~\citep{WangCVPR22ContinualTestTimeDA}  a continual Test-Time
Domain Adaptation (CoTTA) model to limit the error accumulation by using predictions computed via averaging different weights and augmented copies of an image,  which allows mitigating catastrophic forgetting. They propose to
stochastically restore a small amount of network units to their source pre-trained values at each iteration and, in turn, enforcing the adapted model to preserve source 
knowledge over time.

\subsection{Source-free domain adaptation} \label{sec:SFDA}

\begin{figure}[ttt]
\begin{center}
\includegraphics[width=\textwidth]{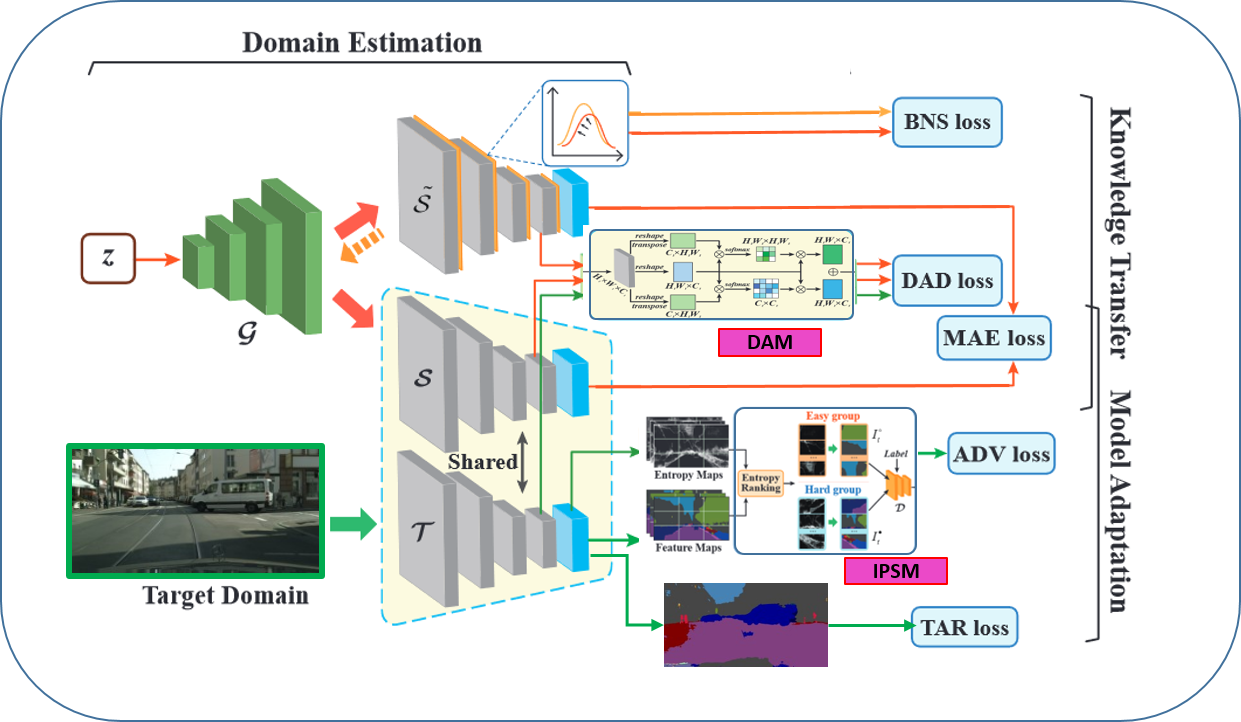}
\caption{Illustration of the source-free DASiS method proposed by  \citet{LiuCVPR21SourceFreeDASS} where during training only a well-trained source model and the unlabeled target domain set are accessible but not the source data. To tackle the problem, they propose a \textit{dual attention distillation mechanism} (DAM) to help the generator ${\cal G}$ to synthesize "source" samples with meaningful semantic context, beneficial to efficient pixel-level domain knowledge transfer. Furthermore,  an \textit{entropy-based intra-domain} patch-level supervision module (IPSM) leverages the correctly segmented patches  during the model adaptation stage (Figure based on~\citep{LiuCVPR21SourceFreeDASS}).}
\label{fig:SFDA}
\end{center}
\end{figure}

\textit{Source-free domain adaptation} constitutes the problem of adapting a given source model to a target domain, but without access to the original source dataset. It has been introduced by
\citet{ChidlovskiiKDD16DomainAdaptationAbsenceSourceDomainData}, who propose solutions for both supervised and unsupervised domain adaptation, testing them in a variety of machine learning problems (\eg, document analysis, object classification, product review classification).

More recently,~\citet{LiCVPR20ModelAdaptationUDAWithoutSourceData} propose to exploit the pre-trained source model as a starting component for an adversarial generative model that generates target-style samples, improving the classifier performance in the target domain, and in turn, improving the generation process. \citet{LiangICML20DoWeReallyNeedAccessSourceUDA} learn a target-specific feature extraction module by implicitly aligning target representations to the source hypothesis, with a method that exploits at the same time information maximization and self-training. 
\citet{KurmiWACV21DomainImpressionSourceDataFreeDA} treat the pre-trained source model as an energy-based function, in order to learn the joint distribution, and  train a GAN that generates annotated samples that are used throughout the adaptation procedure. 
\citet{XiaICCV21AdaptiveAdversarialNetworkSourceFreeDA} propose  a learnable target classifier that improves the recognition ability on source-dissimilar target features, and  perform adversarial domain-level alignment and contrastive matching  at category level.

For semantic segmentation, \citet{LiuCVPR21SourceFreeDASS} propose a \textit{dual attention distillation mechanism} to help the generator to synthesize samples with meaningful semantic context used to perform efficient pixel-level domain knowledge transfer. They  rely on  an \textit{entropy-based intra-domain}  
module to leverage the correctly segmented patches as supervision  during the model adaptation stage (see~\fig{SFDA}).

\citet{SivaprasadCVPR21UncertaintyReductionDASS} propose a solution where the uncertainty of the target domain samples' predictions is minimized, while the robustness against noise perturbations in the feature space is maximized.
\citet{KunduICCV21GeneralizeThenAdaptSourceFreeDASS} decompose the problem into performing first source-only domain generalization and then adapting the model to the target by self-training with reliable target pseudo-labels.

\subsection{Domain generalization}
\label{sec:dg}

In \textit{domain generalization} (DG), the goal is to generalize a model trained on one or several source domains to new, unseen target domains. Therefore, its main goal is learning {\it domain-agnostic representations}. As the target domain is unknown at training time, most DG methods aim to minimize the average risk over all possible target domains. According to \citet{WangX20GeneralizingToUnseenDomainsSurveyDG} and \citet{ZhouX20DomainGeneralizationSurvey}, such DG methods can be categorized into multiple groups, those relying on data randomization~\citep{TobinIROS17DomainRandomizationTransferringDNNSimToReal}, 
ensemble learning~\citep{XuECCV14ExploitingLowRankDomainGeneralization,SeoECCV20LearningToOptimizeDomainSpecificNormalizationDG}, meta-learning~\citep{BalajiNIPS19MetaRegTowardsDGMetaRegularization,LiICML19FeatureCriticNetworksHeterogeneousDG,RahmanPR20CorrelationAwareAdversarialDADG} domain-invariant representation learning~\citep{LiCVPR18DomainGeneralizationWithAdversarialFeatureLearning,MotiianICCV17UnifiedDeepSupervisedDADG}, feature disentanglement~\citep{ChattopadhyayECCV20LearningToBalanceSpecificityInvarianceDG}, self-supervised learning~\citep{CarlucciCVPR19DomainGeneralizationBySolvingJigsawPuzzles,WangECCV20LearningFromExtrinsicIntrinsicSupervisionsDG}, invariant risk minimization~\citep{ArjovskyX20InvariantRiskMinimization} and others.

While not devised \textit{ad hoc} for SiS, several data augmentation methods
have been empirically shown to be well performing for the segmentation task. Indeed,  \citet{VolpiICCV19AddressingModelVulnerabilityDistrShiftsImageTransf,VolpiNIPS19GeneralizingUnseenDomainsAdvDataAugm} and \citet{QiaoCVPR20LearningToLearnSingleDG} show that worst-case data augmentation strategies can improve robustness of segmentation models. \citet{VolpiNIPS19GeneralizingUnseenDomainsAdvDataAugm} propose to create \textit{fictitious visual domains} -- that are hard for the model at hand -- by leveraging adversarial training to augment the source domain, and use them to train the segmentation model. \citet{QiaoCVPR20LearningToLearnSingleDG} extend this idea by relying on meta-learning; to encourage \textit{out-of-domain augmentations}, the authors rely on a Wasserstein auto-encoder which is jointly learned with the segmentation and domain augmentation within a \textit{meta-learning framework}.
\citet{VolpiICCV19AddressingModelVulnerabilityDistrShiftsImageTransf} instead rely on standard image transformations, by using random and evolution search to find the worst-case perturbations that are further used as data augmentation rules.

Concerning DG methods  specifically designed
for SiS (DGSiS), different techniques have been  explored. 
\citet{GongCVPR19DLOWDomainFlowAdaptationGeneralization} propose to learn domain-invariant representations via \textit{domain flow generation}. The main idea is to generate  a continuous sequence of intermediate domains between the source and the target, in order to bridge the gap between them. To translate images from the source domain into an arbitrary intermediate domain, an adversarial loss is used to control how the intermediate domain is related to the two original ones (source and target). Several intermediate domains of this kind are generated, such that the discrepancy between the two domains is gradually reduced in a manifold space.

\citet{YueICCV19DomainRandomizationPyramidConsistencySimulationToReal} rely on domain randomization, where -- using auxiliary datasets -- the synthetic images are translated with multiple real image styles to effectively learn domain-invariant and scale-invariant representations. Instead, \citet{JinX20StyleNormalizationRestitutionDGDA} consider  \textit{style normalization} and \textit{restitution} module to enhance the generalization capabilities, while preserving the discriminative power of the networks. The style normalization is performed by instance normalization to filter out the style variations and therefore foster generalization. To ensure high discriminative leverage, a restitution step adaptively distills task-relevant discriminative features from the residual (\ie the difference between original and style normalized features),
which are then exploited to learn the network.

\citet{LiuTMI20MSNetMultiSiteNetworkProstateSegmHeterogeneousMRIData} extend domain-specific BN layers proposed in~\citep{SeoECCV20LearningToOptimizeDomainSpecificNormalizationDG} for MRI image segmentation, where at inference time an ensemble of prediction is generated and their confidence-weighted average is considered as the final prediction.
\citet{ChoiCVPR21RobustNetImprovingDGUrbanSiSInstanceSelectiveWhitening} propose an instance selective whitening loss which disentangles domain-specific and domain-invariant properties from higher-order statistics of the feature representation, selectively suppressing the domain-specific ones.
\citet{LeeCVPR22WildNetLearningDomainGeneralizedSISFromTheWild}  learn
domain-generalized semantic features by leveraging a
variety of contents and styles from the wild, where they
diversify the styles of the source features with the help of wild styles. This is carried out by adding
several AdaIN~\citep{HuangICCV17ArbitraryStyleTransferRealTimeAdaptiveInstanceNormalization} layers to the feature extractor during the learning process and increasing the intra-class content variability with content extension to the wild in the latent embedding space.

Closely related with DGSiS, \citet{LengyelICCV21ZeroShotDayNightDAWithPhysicsPrior} propose \textit{zero-shot day-to-night domain adaptation} to improve performance on unseen illumination conditions without the need of accessing target samples. The proposed method relies on task agnostic physics-based illumination priors where a trainable Color Invariant Convolution layer is used to transform the input to a domain-invariant representation. It is shown that this layer allows reducing the day-night domain shift in the feature map activations throughout the network and, in turn, improves SiS on samples recorded at night.

\subsection{Semi-supervised domain adaptation}
\label{sec:SSDA}

Semi-supervised learning (SSL) methods exploit at training time accessibility to both a small amount of labeled data and a large amount of unlabeled data. After gaining traction for more standard classification tasks, recently several semi-supervised methods have emerged that address SiS problems (see  \secref{SSLearn}). 

The standard UDA setting shares with semi-supervised learning the availability at training time of labeled and unlabeled data; the core difference is that in the semi-supervised framework both sets are drawn from the same domain (\iid assumption), whereas in UDA they are drawn from different data distributions (source and target).
In Section~\ref{sec:dass_improvements} we have discussed how several strategies from the SSL literature such as pseudo-labeling, self-training, entropy minimization, self-ensembling, have been inherited by DASiS and tailored for cross-domain tasks.
Semi-supervised domain adaptation can be seen as a particular case of them, where on the one hand we can see part of pseudo-labels replaced by GT target labels, or on the other hand we can see the source labeled data extended with labeled target samples.

To address such scenario,~\citet{WangCVPRWS20AlleviatingSemLevelShiftSSDAForSemSegm} leverage a few labeled images from the target domain to supervise the segmentation task and the adversarial semantic-level feature adaptation.
They show that the proposed strategy improves also over a target domain's oracle.~\citet{ChenCVPR21SemiSupervisedDADualLevelDomainMixingSIS} tackle the semi-supervised DASiS problem with a method that relies on a variant of CutMix~\citep{YunICCV19CutMixRegStrategyToTrainStrongClassifWithLocFeats} and a \textit{student-teacher} approach based on self-training. Two kinds of data mixing methods are proposed: on the one hand, directly mixing labeled images from two domains from holistic view; on the other hand, region-level data mixing is achieved by applying two masks to labeled images from the two domains. The latter encourages the model to extract domain-invariant features about semantic structure from partial view. Then, a student model is trained by distilling knowledge from the two complementary \textit{domain-mixed teachers} -- one obtained by direct mixing and another obtained by region-level data mixing -- and which is refined in a self-training manner for another few rounds of teachers trained with pseudo-labels.

\cite{ZhuPAMI21ImprovingSemanticSegmentationEfficientSelfTraining} first train an ensemble of student models with various backbones and network architectures using both labeled source data and pseudo labeled target data where the labels are obtained with a teacher model trained on the labeled source. This model is further  finetuned for the target domain using a small set of labeled samples not only to further  adapt  to the new domain but also to address class-label mismatch across domains (see also~\secref{universal}).

\subsection{Active DASiS}
\label{sec:act-DASS}

Active DASiS~\citep{NingICCV21MultiAnchorActiveDASS,ShinICCV21LabORLabelingOnlyIfRequiredDASS} is related to semi-supervised DASiS. While for the latter we assume that a small set of target samples are already labeled, in the former, an algorithm selects itself the images or pixels to be annotated by human annotators,  and use them to update the segmentation model over iterations.
\citet{NingICCV21MultiAnchorActiveDASS} propose a multi-anchor based active learning strategy to identify the most complementary and representative samples for manual annotation by exploiting the feature distributions across the target and source domains. \citet{ShinICCV21LabORLabelingOnlyIfRequiredDASS} -- inspired by the maximum classifier discrepancy (MCD)~\citep{SaitoCVPR18MaximumClassifierDiscrepancyUDA} -- propose a method that selects the regions to be annotated  based on the mismatch in predictions across the two classifiers.

\begin{figure}[ttt]
\begin{center}
\includegraphics[width=0.9\textwidth]{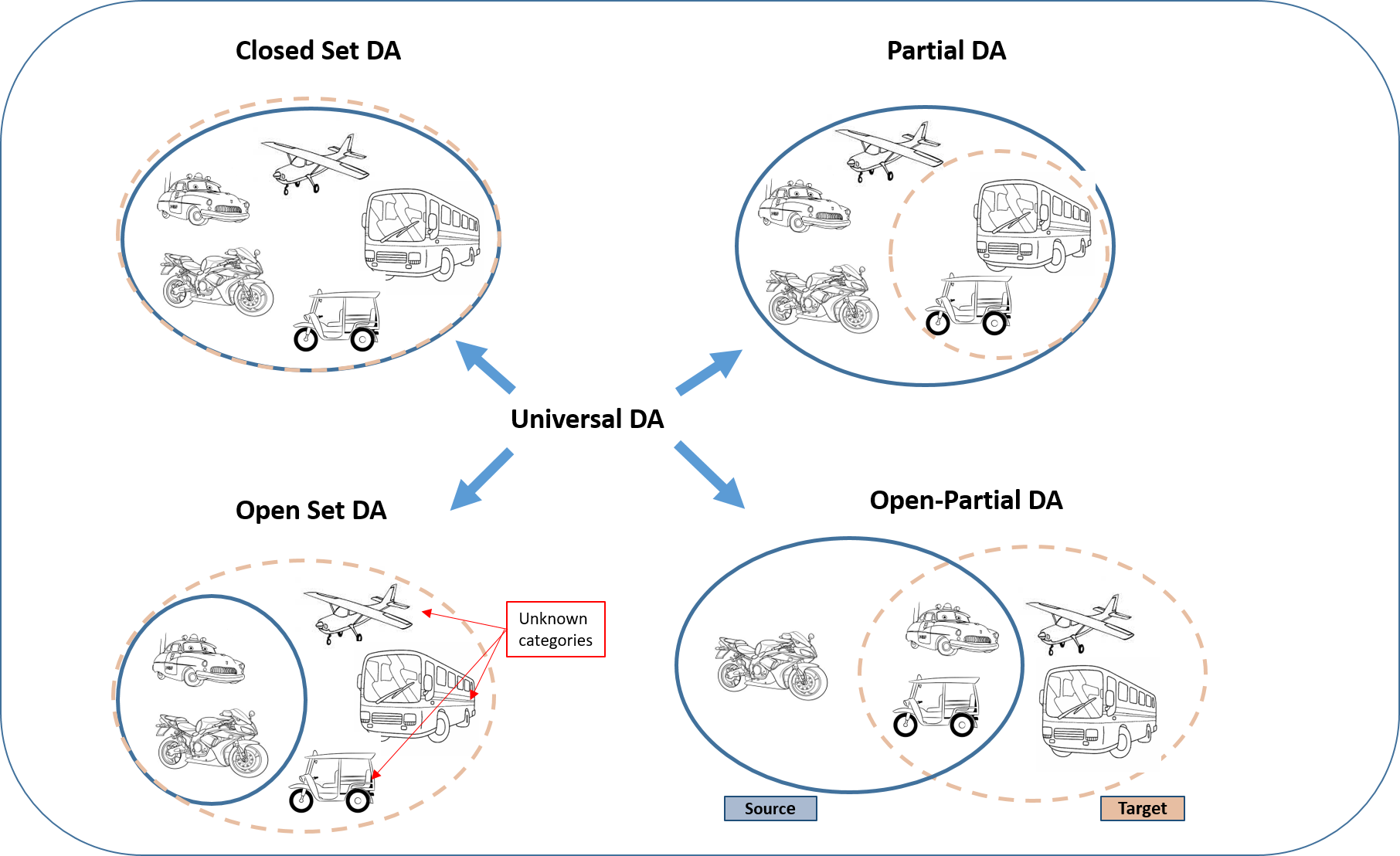}
\caption{A summary of the standard domain adaptation (also known as closed set), partial DA, open set DA and open-partial DA with respect to the overlap between the label sets of the source and the target domains. Universal DA tends to address all cases simultaneously.} 
\label{fig:OpenSetPartial}
\end{center}
\end{figure}

More recently, \citet{XieCVPR22TowardsFewerAnnotationsActiveLearningDASIS} propose a new region-based acquisition strategy for active DASiS, which relies on both region impurity and prediction uncertainty, in order to identify the image regions that are both diverse in spatial adjacency and uncertain in terms of output predictions.

\subsection{Class-label mismatch across domains}
\label{sec:universal}

Another way of sub-dividing domain adaptation approaches is by considering the mismatch between source and target class sets. Specifically, in \emph{partial domain adaptation}~\citep{ZhangCVPR18ImportanceWeightedAdversarialPartialDA,CaoECCV18PartialAdversarialDA,CaoCVPR19LearningtoTransferExamplesPartialDA} the class set of the source is a super-set of the target one, while \emph{open set domain adaptation} \citep{PanaredaBustoICCV17OpenSetDomainAdaptation,SaitoECCV18OpenSetDAbyBackpropagation,RakshitECCV20MultiSourceOpenSetDeepAdvDA,JingICCV21TowardsNovelTargetDiscoveryOpenSetDA} assumes that extra private classes exist in the the target domain. Finally, \emph{universal domain adaptation} \citep{FuECCV20LearningToDetectOpenClassesUniversalDA,LiCVPR21DomainConsensusClusteringUniversalDA,SaitoICCV21OVANetOneVsallNetworkUniversalDA,MaICCV21ActiveUniversalDA} integrates both open set and partial DA.

For what concerns segmentation,~\citet{GongICCV21mDALUMuSDALabelUnificationWithPartialDatasets} propose an MSDA strategy where the label space of the target domain is defined as the union of the label spaces of all the different source domains and the knowledge in different label spaces is transferred from different source domains to the target domain, where the missing labels are replaced by pseudo-labels.

\citet{LiuICCV21AdvUDAInferAlignIterate} 
propose an optimization scheme which alternates between 1) conditional distribution alignment with adversarial UDA relying on estimated \textit{class-wise balancing} in the target and 2) target \textit{label proportion estimates} with Mean Matching~\citep{GrettonBC09CovariateShiftKernelMeanMatching}, assuming conditional distributions alignment between the domains.

\chapter{Datasets and Benchmarks}
\label{c:benchmarks}

In this chapter, we discuss datasets and evaluation protocols commonly adopted in SiS (\secref{SISbench}), DASiS (\secref{DAbench}) and related problems -- such as class-incremental SiS (\secref{class-inc-protocol}) and online adaptation (\secref{oasis-protocol}). We further cover the main evaluation metrics used in SiS (\secref{evaluate}), also discussing more recent alternatives. Furthermore, we emphasize that segmentation performance 
in terms of accuracy or mIoU
is only one of the aspects one should consider when assessing the effectiveness of an SiS approach,  discussing the trade-off between accuracy and efficiency in \secref{efficiency} and the vulnerability of SiS models in \secref{vulnerability}.

\section{SiS Datasets and Benchmarks}
\label{sec:SISbench}

In SiS, we can mainly distinguish the following groups of datasets and benchmarks that we call \textit{object segmentation} (Obj) datasets, \textit{image parsing} (IP) datasets and \textit{scene understanding in autonomous driving} (AD) scenarios. Note that the separation between these datasets are not strict, for example the AD is a particular case of IP. There is also a large set of \textit{medical image} (Med) segmentation datasets and benchmarks (see a list in~\citet{LiuJS21AReviewDeepLearningMedicalImageSegm}) that we do not discuss here. 

PASCAL Visual Object Classes (VOC)~\citep{EveringhamIJCV10ThePascalVisualObjectClassesVOCChallenge} is one of the first and most popular object segmentation datasets. It contains 20 classes to be segmented plus the background. Several versions are available,
the most used ones being the Pascal-VOC 2007 (9,963 images) and Pascal-VOC 2012 (11,5K images). MS COCO \citep{LinECCV14MicrosoftCOCO} is another challenging object segmentation dataset  containing complex everyday scenes with  objects in their natural contexts. It contains 328K images with segmentations of 91 object class.

\begin{table*}
\begin{center}
\tiny
\begin{tabular}{@{}lcccccc@{}}
\toprule
\textbf{Dataset name (ref. paper)}
& \textbf{\# Classes}
& \begin{tabular}{@{}c@{}} \textbf{\# Annotated} \\ \textbf{samples} \end{tabular}
& \begin{tabular}{@{}c@{}} \textbf{Real or} \\ \textbf{sim.} \end{tabular}
& \begin{tabular}{@{}c@{}} \textbf{Video} \\ \textbf{seq.} \end{tabular}
& \begin{tabular}{@{}c@{}} \textbf{Environment/} \\ \textbf{geography} \end{tabular}
& \begin{tabular}{@{}c@{}} \textbf{Visual} \\ \textbf{conditions} \end{tabular} \\
\midrule
\midrule
Cityscapes~\citep{CordtsCVPR16CityscapesDataset} & $30$ & $5,000$* & Real & Yes & Germany; Zurich & $-$ \\
\midrule
BDD100K~\citep{YuCVPR20BDD100KDiverseDrivingDatasetHeterogeneousMultitaskLearning}  & $19$ & $10,000$ & Real & No & United States & $-$ \\
\midrule
KITTI~\citep{GeigerCVPR12AreWeReadyForADKITTI} & $28$ & $400$ & Real & Yes & Germany & $-$ \\
\midrule
CamVid~\citep{BrostowPRL09SemanticObjectClassesVideo}  & $32$ & $701$ & Real & Yes & Cambridge (UK) & $-$ \\
\midrule
Mapillary~\citep{NeuholdICCV2017MapillaryDataset} & $66$ & $25,000$ & Real & No & Worldwide & $-$ \\
\midrule
IDD~\citep{VarmaWACV19IDDDatasetExploringADUnconstrainedEnvironments} & $34$ & $10,004$ & Real & Yes & India & $-$ \\
\midrule
RainCityscapes~\citep{HuCVPR19DepthAttentionalFeaturesSingleImageRainRemoval} & $32$ & $10,620$ & Real & Yes & Germany & Artificial rain \\
\midrule
FoggyCityscapes~\citep{SakaridisECCV18ModelAdaptationSyntheticRealSemanticDenseFoggy} & $32$ & $15,000
$ & Real & Yes & Germany & Artificial fog \\
\midrule
ACDC~\citep{SakaridisICCV21ACDCAdverseConditionsDatasetSIS}  & 19 & $4,006$ & Real & Yes & Switzerland &  Daytime; Weather  \\
\midrule
FoggyZurich~\citep{SakaridisECCV18ModelAdaptationSyntheticRealSemanticDenseFoggy} &
$19$ & $40$** 
& Real & Yes & Zurich & Fog \\
\midrule
GTA-5~\citep{RichterECCV16PlayingForData} & $19$ & $24,966$ & Sim & Yes & $-$ & $-$ \\
\midrule
SYNTHIA~\citep{RosCVPR16SYNTHIADataset} & $13$ & $200,000$ & Sim &  Yes &  Highway; NYC; EU & Season; Daytime; Weather \\
\midrule
SYNTHIA-RAND~\citep{RosCVPR16SYNTHIADataset} & $11$ & $13,407$ & Sim & No & $-$ & $-$ \\
\midrule
KITTI-v2~\citep{CabonX20VirtualKITTI2} & $15$ & $21,260$ & Sim & Yes & Germany & Daytime; Weather \\
\midrule
Synscapes~\citep{WrenningeX18SynscapesPhotorealisticSyntheticDataset} & $19$ & $25,000$ & Sim & No & - & Daytime; Overcast; Scene param. \\
\bottomrule
\end{tabular}

\end{center}

\caption{Datasets for research on urban scene SiS. From leftmost to rightmost columns, we indicate the dataset name, the number of categories covered by annotations, whether the dataset contains real or simulated (rendered) images, whether samples are recorded as video sequences (\ie, mimicking an agent driving in an urban environment), the locations from which samples were recorded or the ones simulated by the engine, and whether different visual conditions can be set by the user.
*Cityscapes 
contains $20,000$ additional samples with coarse annotations and 
**FoggyZurich 
contains also $3,808$ unlabelled samples.}


\label{tab:semsegm_datasets}
\end{table*}



Image or scene parsing datasets contain both~\textit{things} (objects) and \textit{stuff} classes. One of the first such dataset is MSRC-21~\citep{ShottonECCV06TextonBoostJointAppearanceShapeContextSemSegm}, containing 21 categories and 591 images. The Pascal Context~\citep{MottaghiCVPR14TheRoleOfContext}, extends to IP the segmented images from Pascal-VOC 2010 by labeling the background. It has 10,1K images and 400 classes, however mainly a subset of 59 classes is used, ignoring the others as they have rather low frequency in the dataset. SiftFlow \citep{LiuCVPR09NonparametricSceneParsingLabelTransferViaDenseSceneAlignment} includes 2,688 images from the LabelMe database~\citep{RussellIJCV08LabelMeDatabaseToolAnnotation} annotated with 33 semantic classes. The Stanford background dataset~\citep{GouldICCV09DecomposingSceneGeometricSemanticallyConsistentRegions} contains 715 outdoor images from LabelMe, MSRC and Pascal-VOC where the aim is to separate the foreground (single class) from the background, identifying the seven following semantic~\textit{stuff} regions: ``sky'', ``tree'', ``road'', ``grass'', ``water'', ``mountain'' and ``buildings''. The most used IP dataset is ADE20K~\citep{ZhouIJCV19SemanticUnderstandingScenesAde20kDataset} which contains 20K images with 150 semantic categories.

There exists a large set of image parsing datasets proposed in the literature specifically built for urban scene understanding, targeting autonomous driving (AD) scenarios. One of the most popular datasets used to compare SiS methods  is Cityscapes~\citep{CordtsCVPR16CityscapesDataset}, but
with the increased interest for the AD scenarios,  recently a large set of labeled urban scene datasets have been proposed, both real and synthetically rendered with game-engines.  
In Table~\ref{tab:semsegm_datasets} we provide a summary of such AD oriented SiS datasets  with their most important characteristics: the number of classes, the number of annotated samples, whether images are real or rendered, whether the dataset contains video sequences (and not only temporally uncorrelated images), the geographical location (for what concerns simulated datasets, we report the simulated area indicated, if available), and whether the dataset allows setting arbitrary conditions (seasonal, weather, daylight, \etc). In addition, in Table~\ref{tab:semsegm_classes} we present a summary of the classes available in these different datasets, to ease the comprehension of the compatibility between different models. They are  also interesting in the light of incremental SiS (see \secref{class-inc}) and  new DASiS problems where the sets of semantic classes in the source and target sets do not  coincide (shortly discussed in \secref{universal}).


\begin{table*}
\begin{center}
{
\tiny
\begin{NiceTabular}{@{}|l|ccccccccccc|@{}}[code-before = 
\rowcolor{gray!10}{3}
\rowcolor{gray!10}{5}
\rowcolor{gray!10}{7}
\rowcolor{gray!10}{9}
\rowcolor{gray!10}{11}
\rowcolor{gray!10}{13}
\rowcolor{gray!10}{15}
\rowcolor{gray!10}{17}
\rowcolor{gray!10}{19}
\rowcolor{gray!10}{21}
\rowcolor{gray!10}{23}
\rowcolor{gray!10}{25}
\rowcolor{gray!10}{27}
]
\toprule


  \rotatebox{0}{ \textbf{Classes} }
& \rotatebox{90}{ Cityscapes }
& \rotatebox{90}{ BDD100K}
& \rotatebox{90}{ CamVid}
& \rotatebox{90}{ IDD}
& \rotatebox{90}{ ACDC}
& \rotatebox{90}{ GTA-5}
& \rotatebox{90}{ SYNTHIA-R}
& \rotatebox{90}{ SYNTHIA}
& \rotatebox{90}{ KITTI-v2}
& \rotatebox{90}{ FoggyZurich}
& \rotatebox{90}{ Synscapes}
\\
\midrule

%
Bicycle              & \checkmark & \checkmark & \checkmark & \checkmark & \checkmark & \checkmark & \checkmark & \checkmark &            & \checkmark & \checkmark \\
Bridge               & \checkmark &            & \checkmark & \checkmark &            &            &            &            &            &            &            \\
Building             & \checkmark & \checkmark & \checkmark & \checkmark & \checkmark & \checkmark & \checkmark & \checkmark & \checkmark & \checkmark & \checkmark \\
Bus                  & \checkmark & \checkmark & \checkmark & \checkmark & \checkmark & \checkmark & \checkmark &            &            & \checkmark & \checkmark \\
Car                  & \checkmark & \checkmark & \checkmark & \checkmark & \checkmark & \checkmark & \checkmark & \checkmark & \checkmark & \checkmark & \checkmark \\
Caravan              & \checkmark &            &            & \checkmark &            &            &            &            & \checkmark &            &            \\
%
Fence                & \checkmark & \checkmark & \checkmark & \checkmark & \checkmark & \checkmark & \checkmark & \checkmark &            & \checkmark & \checkmark \\
Guard rail           & \checkmark &            &            & \checkmark &            &            &            &            & \checkmark &            &            \\
Lane marking         &            &            & \checkmark &            &            &            & \checkmark & \checkmark &            &            &            \\
Motorcycle           & \checkmark & \checkmark & \checkmark & \checkmark & \checkmark & \checkmark & \checkmark &            &            & \checkmark & \checkmark \\
%
Parking              & \checkmark &            & \checkmark & \checkmark &            &            & \checkmark &            &            &            &            \\
Person               & \checkmark & \checkmark & \checkmark & \checkmark & \checkmark & \checkmark & \checkmark & \checkmark &            & \checkmark & \checkmark \\
Pole                 & \checkmark & \checkmark & \checkmark & \checkmark & \checkmark & \checkmark & \checkmark & \checkmark & \checkmark & \checkmark & \checkmark \\
Rail track           & \checkmark &            &            & \checkmark &            &            &            &            &            &            &            \\
Rider                & \checkmark & \checkmark &            & \checkmark & \checkmark & \checkmark & \checkmark &            &            & \checkmark & \checkmark \\
Road                 & \checkmark & \checkmark & \checkmark & \checkmark & \checkmark & \checkmark & \checkmark & \checkmark & \checkmark & \checkmark & \checkmark \\
%
Sky                  & \checkmark & \checkmark & \checkmark & \checkmark & \checkmark & \checkmark & \checkmark & \checkmark & \checkmark & \checkmark & \checkmark \\
Sidewalk             & \checkmark & \checkmark & \checkmark & \checkmark & \checkmark & \checkmark & \checkmark & \checkmark &            & \checkmark & \checkmark \\
Terrain              & \checkmark & \checkmark &            &            & \checkmark & \checkmark & \checkmark &            & \checkmark & \checkmark & \checkmark \\
%
Train                & \checkmark & \checkmark & \checkmark &            & \checkmark & \checkmark & \checkmark &            &            & \checkmark & \checkmark \\
Traffic light        & \checkmark & \checkmark & \checkmark & \checkmark & \checkmark & \checkmark & \checkmark &            & \checkmark & \checkmark & \checkmark \\
Traffic sign         & \checkmark & \checkmark & \checkmark & \checkmark & \checkmark & \checkmark & \checkmark & \checkmark & \checkmark & \checkmark & \checkmark \\
%
Truck                & \checkmark & \checkmark & \checkmark & \checkmark & \checkmark & \checkmark & \checkmark &            & \checkmark & \checkmark & \checkmark \\
Tunnel               & \checkmark &            & \checkmark & \checkmark &            &            &            &            &            &            &            \\
Vegetation           & \checkmark & \checkmark & \checkmark & \checkmark & \checkmark & \checkmark & \checkmark & \checkmark & \checkmark & \checkmark & \checkmark \\
Wall                 & \checkmark & \checkmark & \checkmark & \checkmark & \checkmark & \checkmark & \checkmark &            &            & \checkmark & \checkmark \\

\bottomrule
\end{NiceTabular}
} 
\end{center}
\caption{Categories of which annotation is provided in different SiS datasets. We report classes available in 
several (at least three) distinct datasets: some datasets, \eg  CamVid~\citep{BrostowPRL09SemanticObjectClassesVideo}, contain a variety
of other categories.}
\label{tab:semsegm_classes}
\end{table*}


\myparagraph{Annotating SiS datasets} 
Generally, software tools that allow to annotate images are based on an interface where the user can manipulate polygons that are shaped according to the image's instances; such polygons are further processed into segmentation maps. Some examples of popular, open-source annotation tools are LabelMe\footnote{\url{https://github.com/wkentaro/labelme}}, Label Studio\footnote{\url{https://github.com/heartexlabs/label-studio}} and VIA\footnote{\url{https://gitlab.com/vgg/via}}.

Initially taking an hour or more per image~\citep{CordtsCVPR16CityscapesDataset}, recent semi-automatic tools manage to reduce the annotation time for common urban classes (``people'', ``road'' ``surface'' or ``vehicles'') by relying, \eg, on pre-trained models for object detection\footnote{\url{https://github.com/virajmavani/semi-auto-image-annotation-tool}}, -- however they still require manual verification and validation. For an up-to-date collection of annotation tools, we refer to the dedicated page\footnote{\url{https://github.com/heartexlabs/awesome-data-labeling}.}

\subsection{Evaluating SiS performance}
\label{sec:evaluate}

To evaluate SiS, the \textit{overall pixel accuracy} and the \textit{per-class accuracy} have been proposed in \citep{ShottonECCV06TextonBoostJointAppearanceShapeContextSemSegm}. 
The former computes the proportion of correctly labeled pixels, while the latter calculates the proportion of correctly labeled pixels for each class and then averages over the classes.
The Jaccard Index (JI), more popularly known as \textit{intersection over the union} (IoU), takes into account both the false positives and the missed values for each class. It measures the intersection over the union of the labeled segments for each class and reports the average. This measure 
became the standard to evaluate SiS models, after having been introduced in the Pascal-VOC challenge \citep{EveringhamIJCV10ThePascalVisualObjectClassesVOCChallenge} in 2008. \citet{LongCVPR15FullyConvolutionalNetworksSegmentation} propose, in addition, a \textit{frequency weighted IoU} measure where the IoU for each class is weighted by the frequency of GT pixels corresponding to that class.

We schematize these main metrics below, following the notation used by \citet{LongCVPR15FullyConvolutionalNetworksSegmentation}. Let $n_{ij}$ be the number of pixels from the \ith class that are classified as belonging to the \jth class where  $i,j \in \{1,\cdots, C\}$, $C$ being the number of different semantic classes. Let $t_i = \sum_j n_{ij}$ be the total number of pixels of the \ith class. The metrics introduces above are defined as follows:

\begin{itemize}
  \item \textbf{Mean IoU}:
    $\frac{1}{C}\sum_i\frac{ n_{ii}}{(t_i + \sum_j n_{ji} - n_{ii})}$
  \item \textbf{Frequency weighted IoU}:
    $\frac{1}{\sum_k t_k}\sum_i\frac{t_i\cdot n_{ii}}{(t_i + \sum_j n_{ji} - n_{ii})}$
  \item \textbf{Pixel accuracy}:
    $\frac{\sum_i n_{ii}}{\sum_i t_i}$
  \item \textbf{Mean accuracy}:
    $\frac{1}{C}\sum_i\frac{ n_{ii}}{t_i}.$
\end{itemize}

The above measures are generally derived from the confusion matrix computed over the whole dataset having the main advantage that there is no need to handle the absent classes in each image. While these metrics are the most used to evaluate and compare SiS and DASiS models, we would like to mention below a few other metrics that have been introduced in the literature to evaluate SiS models, and  could also be interesting for evaluating DASiS.

Instead of relying on the confusion matrix computed over the whole dataset,
\citet{CsurkaBMVC13WhatGoodEvalMeasureSemSegm} propose to evaluate the pixel accuracy, the mean accuracy and the IoU for each image individually, where the IoU is computed by averaging only  over the classes present in the GT segmentation map of the image. The main rationale behind this is that the measures computed over the whole dataset do not enable to distinguish an algorithm that delivers a medium score on all images from an algorithm that performs very well on some images and very poorly on others (they could yield a very similar averages). To better assess such differences,~\citet{CsurkaBMVC13WhatGoodEvalMeasureSemSegm} propose to measure the percentage of images with a performance higher than a given threshold. Then,  given a pair of approaches, the percentage of images for which one of the method outperforms the other one is 
analyzed, \eg considering the statistical difference of two segmentation algorithms with t-test. Finally, it has also been noticed by~\citet{CsurkaBMVC13WhatGoodEvalMeasureSemSegm} that per-image scores reduce the bias \wrt large objects, as missing or incorrectly segmented small objects have  low impact on the global confusion matrix.

Another important aspect of semantic segmentation is the accurate semantic border detection. To evaluate the accuracy of boundary segmentation, \citet{KohliIJCV09RobustHigherOrderPotentialsEnforcingLabelConsistency} propose Trimap that defines a narrow band around each contour and computes pixel accuracies within the given band. Instead, to measure the quality of the segmentation boundary,   \citet{CsurkaBMVC13WhatGoodEvalMeasureSemSegm} extend the Berkeley contour matching (BCM) score~\citep{MartinPAMI04LearningToDetectNaturalImageBoundaries} -- proposed to evaluate similarity between unsupervised segmentation and human annotations -- to SiS, where a BCM score is computed between the GT and predicted contours corresponding to each semantic class (after binarizing first both segmentation maps). The scores are averaged over the classes present in the GT map.

\subsection{Trade-off between accuracy and efficiency}
\label{sec:efficiency}

The segmentation accuracy is not a unique metric when evaluating and comparing segmentation models. Indeed, SiS can be extremely demanding for high computational resources -- in particular due to the fact that it is a pixel-level task, as opposed to image-level tasks. In real applications where latency is crucial, one 
needs to trade-off accuracy for efficiency. Indeed, as previously discussed, being a key element of scene understanding for autonomous driving, robotic applications or augmented reality, semantic segmentation models should accommodate real-time settings.

Historical methods, in order to achieve reasonable performance, often required a costly post-processing. While deep neural network 
models have significantly boosted the segmentation performance, in most cases this improvement came with a significant cost increase both on model parameters and computation, both at train and inference time. 
 
Several solutions have been proposed to find a good trade-off between accuracy and efficiency. 
One possibility is to reduce the computational complexity by restricting the input size \citep{WuX17RealTimeSISViaSpatialSparsity,ZhaoECCV18ICNet4RealTimeSISonHighRes}; yet, this comes with the loss of fine-grain details and, hence, accuracy drops -- especially around the boundaries. An alternative solution is to boost the inference speed by pruning the channels of the network, especially in the early stages of the base model \citep{BadrinarayananPAMI17SegnetDeepConvEncoderDecoder,PaszkeX16ENetDeepNNForRealTimeSemSegm}. Due to the fact that such solutions weaken the spatial capacity,~\citet{PaszkeX16ENetDeepNNForRealTimeSemSegm} propose to abandon the downsampling operations in the last stage, at the cost of diminishing the receptive field of the model. 
To further overcome the loss of spatial details, these methods often use U-shape architectures to gradually increase the spatial resolution and to fill some missing details that however introduces additional computational cost. 

Instead,~\citet{YuECCV18BiSeNetBilateralSegmentationNetworkRealTimeSiS} propose the Bilateral Segmentation Network (BiSeNet) where two components -- the \textit{Spatial Path} and the \textit{Context Path} -- are devised to confront with the loss of spatial information and shrinkage of receptive field respectively.

The segmentation accuracy obtained with Deep Convolutional Networks has further been improved by Transformer-based SiS models (see some examples in~\secref{tranfSiS}). These networks rely on high-performing attention-based modules which have linear complexity with respect to the embedding dimension, 
but a quadratic complexity with respect to the number of tokens. In vision applications, the number of tokens is typically linearly correlated with the image resolution -- yielding a quadratic increase in complexity and memory usage in models strictly using self-attention, such as ViT~\citep{DosovitskiyICLR21AnImageIsWorth16x16WordsTransformersAtScale}. To alleviate this increase, local attention modules were proposed such as Swin~\citep{LiuICCV21SwinTransformerHierarchicalViTShiftedWindows}. Furthermore, \citet{VaswaniCVPR21ScalingLocalSelfAttention4ParamEfficientVisualBackbones} found that a combination of local attention blocks and convolutions result in the best trade-off between memory requirements and translational equivariance. Instead~\citet{HassaniX22NeighborhoodAttentionTransformer} propose the~\textit{Neighborhood Attention Transformer}, which limits each query token’s receptive field to a fixed-size neighborhood around its corresponding tokens in the \textit{key-value} pair, controlling the receptive fields in order to balance between translational invariance and equivariance. \citet{ZhangCVPR22TopFormerTokenPyramidTransformer4MobileSemSegm} propose a mobile-friendly architecture named Token Pyramid Vision Transformer (TopFormer) which takes tokens from various scales as input to
produce scale-aware semantic features with very light computation cost.

Finally, the recent ConvNeXt architecture proposed by~\citet{LiuCVPR22AConvNet4The2020s} competes favorably with Transformers in terms of accuracy, scalability and robustness across 
several tasks including SiS, while maintaining the efficiency of standard ConvNets. 

\subsection{Vulnerability of SiS models}
\label{sec:vulnerability}

While very effective when handling samples from the training distribution, it is well known that deep learning-based models can suffer when facing~\textit{corrupted} samples~\citep{HendrycksICLR19BenchmarkingNNRobustnessCommonCorruptionsPerturbations}. Crucially, these models suffer from perturbations that are imperceptible to the human eye, but causing severe prediction errors~\citep{SzegedyICLR14IntriguingPropertiesNeuralNetworks}. Modern SiS models are also vulnerable in this sense, therefore increasing their robustness against natural or adversarial perturbations is an active research area. Finally, models with a finite set of classes, including SiS models, can suffer when instances of previously unseen categories appear in a scene.

\myparagraph{Adversarial perturbations}
\cite{XieICCV17AdversarialExamplesforSemSegmandObjectDetection} and
\cite{MetzenICCV17UniversalAdversarialPerturbationsSiS} concurrently show for the first time that semantic segmentation 
models can also be fooled by perturbations that are imperceptible to the human eye. 
\cite{MetzenICCV17UniversalAdversarialPerturbationsSiS} show that it is possible to craft \textit{universal} perturbations~\citep{MoosaviDezfooliCVPR17UniversalAdversarialPerturbations}, namely perturbations that are sample-agnostic, that can make the network consistently miss-classify a given input. In particular, they show how to craft perturbations to 1) make the SiS model provide always the same output and 2) make the model avoid predicting ``cars'' or ``pedestrians''. \cite{XieICCV17AdversarialExamplesforSemSegmandObjectDetection} instead focus on sample-specific adversarial perturbations, proposing the ``Dense Adversary Generation'' algorithm. Both works raise security issues on the reliability of SiS models, and therefore the overall systems they are embedded into.

\myparagraph{Corruptions}
\cite{HendrycksICLR19BenchmarkingNNRobustnessCommonCorruptionsPerturbations} showed that deep neural network models for image classification are extremely brittle against simple input miss-specification, such as Gaussian and salt-and-pepper noises, but also to artificial corruptions and contrast or brightness modifications such as simulated fog and snow.~\cite{KamannCVPR20BenchmarkingRobustnessSemSegmModels} extend this analysis to SiS models and show that the same conclusions hold: the models are very vulnerable against simple corruptions, which -- even though perceptible -- would not cause particular difficulties to a human eye.

\myparagraph{Unseen classes}
The out-of-distribution (OOD) detection~\citep{HendrycksICLR17BaselineforDetectingMisclassidfiedandOODExamplesinNN} literature is a very active topic in computer vision: given that the number of classes a model can predict is finite, it is important to be able to handle images with unknown instances. In the case of SiS models, this results in being able to determining when \textit{some pixels} in an image are related to a class the model had never been trained on. 

\cite{BlumICCVWS19FishyscapesBenchmark4SafeSiSAD} and \cite{ChanNeurIPS21SegmentMeIfYouCanDataset} propose the ``Fishyscapes'' and the ``SegmentMeIfYouCan'' benchmarks, that allow to evaluate and compare SiS models on the task of determining which pixels are related to unknown classes. The latter further introduces a new problem where the task is to determine pixels associated with road obstacles (from known and unknown classes). For what concerns methods for the task of determining pixels from unknown classes, most of them are derived from the OOD literature~\citep{HendrycksICLR17BaselineforDetectingMisclassidfiedandOODExamplesinNN,LiangICLR18EnhancingTheReliabilityofOODImageDetectionInNN} and the uncertainty literature~\citep{KendallNIPS17WhatUncertaintiesDoWeNeedInBayesDLForCV}. While methods in both fields are typically designed for classification tasks, they can be extended to SiS by applying them at pixel level instead of image level.

\subsection{Class-incremental SiS protocols}
\label{sec:class-inc-protocol}

In Section~\ref{sec:class-inc} we formulated the problem of class-incremental learning -- in the context of SiS. In the following lines, we review the main protocols used to evaluate such class-incremental SiS algorithms. For reference, the first protocols for this tasks have been
proposed in~\citep{CermelliCVPR20ModelingBackgroundIncrementalLearningSemSegm,MichieliCVIU21KnowledgeDistillationIncLearnSemSegm}. 

The learning procedure, as typical in continual learning, is divided in a sequence of different tasks. In the context of class-incremental SiS, solving a task means learning to segment novel classes, given images where the classes of interest are annotated with GT,  and the others are considered as ``background''. The first task is defined as a learning procedure over a multitude of different classes (as generally happens during model's pre-training). In the following tasks, one or more classes are learned, but generally in inferior numbers with respect to number of categories learned during the first task. 

Formally, given a dataset $D$ with $N$ classes, we will indicate the benchmark as $M-K$, which means that the model is  first trained  on $M$ classes, then it learns $K$ new classes at the time (resulting in $1+(N-M)/K$ consecutive learning steps). Current class-incremental SiS approaches  were evaluated mainly on Pascal
VOC'12~\citep{EveringhamIJCV10ThePascalVisualObjectClassesVOCChallenge} (20 classes) and ADE20K~\citep{ZhouIJCV19SemanticUnderstandingScenesAde20kDataset} (150 classes) datasets. Following the notations above, the following benchmarks have been considered by the community: for Pascal-VOC 2012,
$19-1$ (2 tasks), $15-5$ (5 tasks) and  $15-1$ (2 tasks) and  for ADE20K, $150-50$ (2 tasks), $150-50$ (2 tasks) and $50-50$ (3 tasks).

Furthermore, two different setups are considered in~\citep{CermelliCVPR20ModelingBackgroundIncrementalLearningSemSegm}: the \textit{Disjoint} one, where each task is defined by images that are unique for that task only -- which cannot contain classes associated with classes that will be seen in the future; and the \textit{Overlapped} one, where future classes may be present, and images can be replicated across different tasks.

\section{DASiS Benchmarks}
\label{sec:DAbench}

Understanding traffic scene images taken from vehicle mounted cameras is important for such advanced tasks as autonomous driving and driver assistance. It is a challenging problem due to large variations under different weather or illumination conditions~\citep{DiTITS18CrossDomainTrafficSceneUnderstandingTL} or when a model needs to cope with different environments such as city, countryside and highway. 

\begin{figure*}[ttt]
\begin{center}
\includegraphics[width=0.9\textwidth]{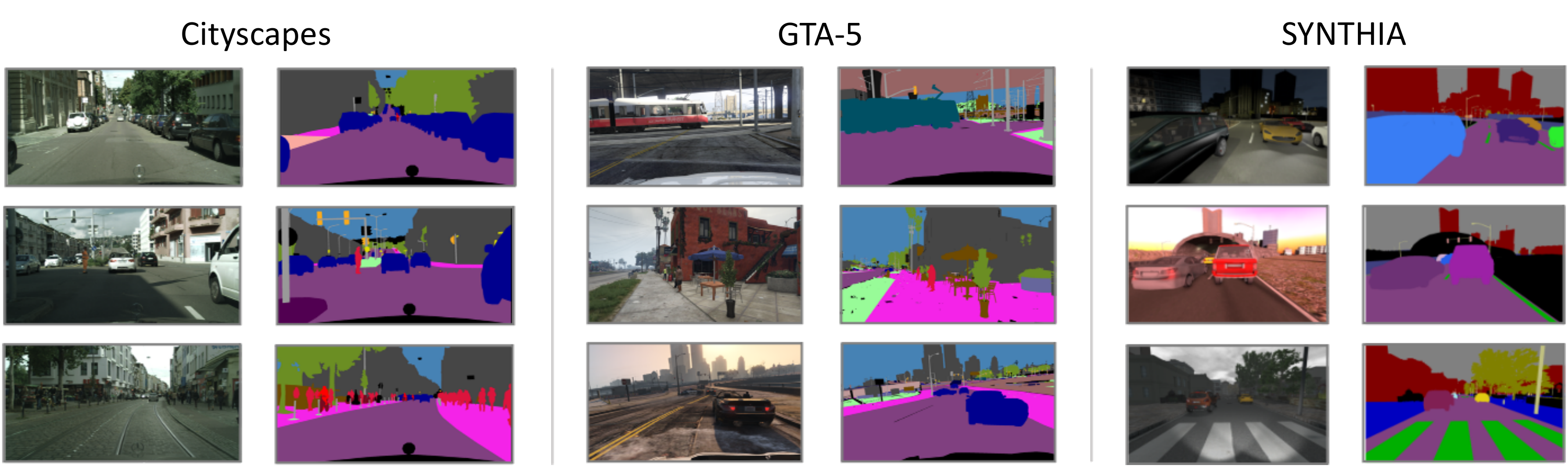}
\caption{
{\bf Left:} Samples from Cityscapes~\citep{CordtsCVPR16CityscapesDataset} recorded in the real world. They allow to evaluate the model performance on images that resemble the ones an agent will cope with at deployment; the difficulty of collecting real, large-scale datasets is the huge cost required to obtain fine annotations.
{\bf Middle:} Synthetic data from GTA-5~\citep{RichterECCV16PlayingForData},
obtained with high quality game engines, which makes easy the  pixel-wise annotation for SiS
and scene understanding. However, if the domain shift between real and synthetic data is not addressed, models trained on GTA-5 perform poorly on Cityscapes.
{\bf Right:} An autonomous car must cope with large variations, such as day vs. night, weather condition changes, or structural differences, which might affect the image appearance even when the image is taken from the same viewpoint. Simulation engines allow generating large number of samples from urban environments in different conditions, as for example in the SYNTHIA~\citep{RosCVPR16SYNTHIADataset} dataset.}
\label{fig:segmDAex}
\end{center}
\end{figure*}

Even though relying on real samples (such as the datasets listed in Table~\ref{tab:semsegm_datasets})
allows assessing model performance in conditions that are more similar to deployment ones, manually annotating an image at pixel level for SiS is a very tedious and costly operation. Recent progresses in computer graphics and modern graphics platforms such as game engines raise the prospect of easily obtaining labeled, synthetic datasets. Some examples in this direction are SYNTHIA~\citep{RosCVPR16SYNTHIADataset} and GTA-5~\citep{RichterECCV16PlayingForData} 
(see examples in~\fig{segmDAex}, the middle and right sides).

However, models learned on such datasets might not be optimal due to the domain shift between synthetic and real data. To tackle this problem, a large set of DASiS methods have been proposed, most of which we surveyed in Chapter~\ref{c:DAsemsegm}. These methods start with a model pre-trained on the  
simulated source data (typically GTA-5 or SYNTHIA) which is adapted to real target data, for which it is  assumed no access to ground-truth annotations.  Typically, the Cityscapes~\citep{CordtsCVPR16CityscapesDataset} dataset is considered in most papers (see examples in \fig{segmDAex} (left)), however more recent methods started to provide results on newer  dataset (listed in Table~\ref{tab:semsegm_datasets}).  This scenario mimics the realistic conditions such that a large database of simulated, labeled samples is available for training, and the model needs to be adapted to  real world conditions without having access to ground-truth annotations.

We summarize the most common settings used in the DASiS research  in Table~\ref{tab:semsegm_benchmarks}. They have been introduced in the pioneering DASiS study by \citet{HoffmanX16FCNsInTheWildPixelLevelAdversarialDA}. As the first row in the table indicates, the most widely used benchmark is GTA-5~\citep{RichterECCV16PlayingForData} $\rightarrow$ Cityscapes~\citep{CordtsCVPR16CityscapesDataset} task. It represents a sim-to-real adaptation problem, since GTA-5 
was conceived to be consistent with Cityscapes
annotations. Following the notation from \secref{UDA}, the source dataset $\calD_\calS$ is defined by GTA-5~\citep{RichterECCV16PlayingForData} annotated samples, and the target dataset $\calD_\calT$ is defined by Cityscapes~\citep{CordtsCVPR16CityscapesDataset} (non-annotated) samples.

 \begin{table}[ttt]
\begin{center}
{
\footnotesize 
\begin{tabular}{@{}lcc@{}}
\multicolumn{3}{c}{\textbf{Main benchmarks for DASiS}} \\
\toprule

  \begin{tabular}{@{}c@{}} \textbf{Source} \\ \textbf{domain} \end{tabular}
& \begin{tabular}{@{}c@{}} \textbf{Target} \\ \textbf{domain} \end{tabular} 
& \begin{tabular}{@{}c@{}} \textbf{Adaptation} \\ \textbf{type} \end{tabular}  \\

\midrule
\midrule

  GTA-5
& Cityscapes
& Sim-to-real \\
\midrule

  SYNTHIA-RAND
& Cityscapes
& Sim-to-real \\
\midrule

  Cityscapes (Train)
& Cityscapes (Val)
& Cross-city (real) \\
\midrule

  SYNTHIA (Fall)
& SYNTHIA (Winter)
& Cross-weather (sim) \\


\bottomrule

\end{tabular}
} 
\end{center}
\caption{The most widely used benchmarks within the DASiS community. The first column 
indicates the source dataset (labeled images available); the second column 
indicates the target dataset (unlabeled images available); the third column indicates the
type of adaptation problem.} 
\label{tab:semsegm_benchmarks}
\end{table}

\begin{figure*}[ttt]
\begin{center}
\includegraphics[width=0.9\textwidth]{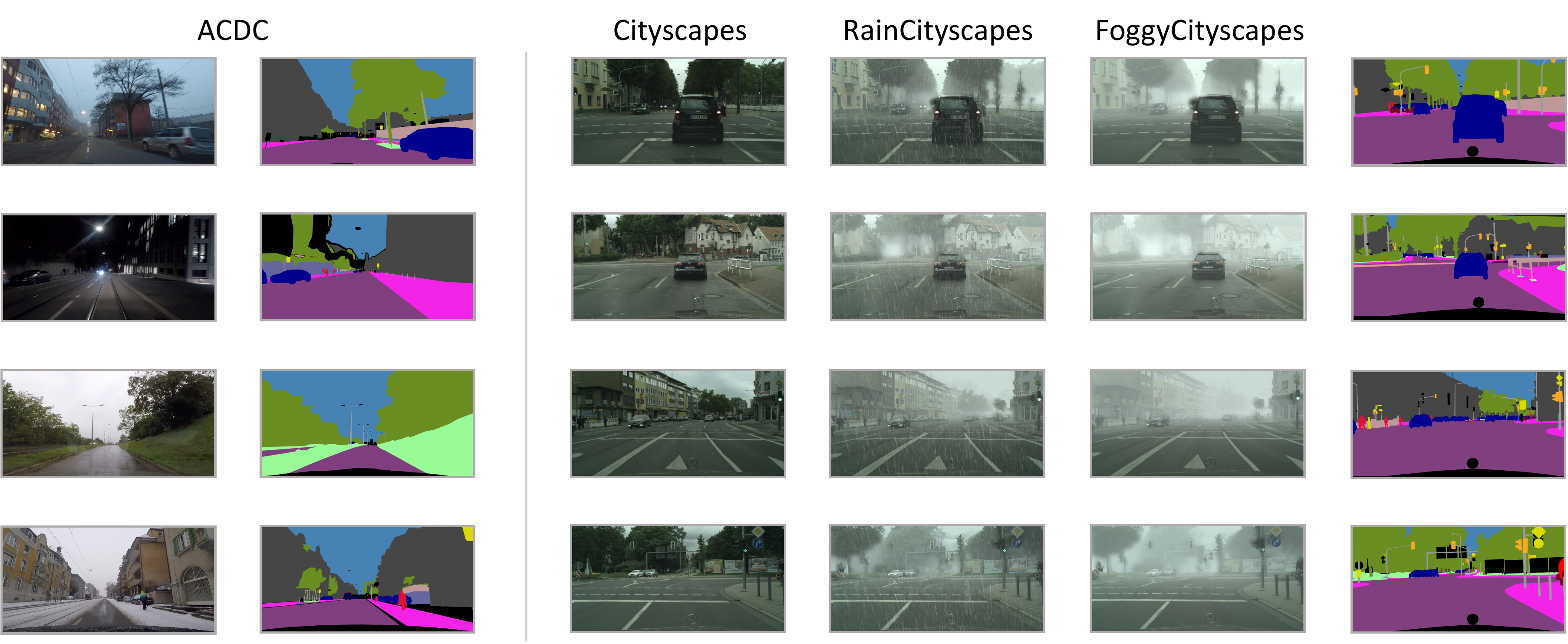}
\caption{
{\bf Left:} Example images from ACDC dataset~\citep{SakaridisICCV21ACDCAdverseConditionsDatasetSIS} which permits to assess the model performance on real-world weather condition changes (\textit{fog}, \textit{night}, \textit{snow}, \textit{rain}).
{\bf Right:} Example images from RainCityscape~\citep{HuCVPR19DepthAttentionalFeaturesSingleImageRainRemoval} and
FoggyCityscape~\citep{SakaridisECCV18ModelAdaptationSyntheticRealSemanticDenseFoggy}, which provide Cityscapes~\citep{CordtsCVPR16CityscapesDataset} images with simulated \textit{rain} and \textit{fog}, respectively.}
\label{fig:ACDC}
\end{center}
\end{figure*}

Naturally, datasets generated with the help of simulation engines are significantly larger, as they are able to generate synthetic data under a broad set of conditions (the only exception is GTA-5~\citep{RichterECCV16PlayingForData}, that is considerably large but does not allow the user to set different visual conditions). Still, in order to evaluate how the models will perform in the real environment on various real conditions, these synthetic  datasets  might be not sufficient. Therefore, an important contribution to the semantic segmentation landscape is the real-image ACDC dataset~\citep{SakaridisICCV21ACDCAdverseConditionsDatasetSIS}, that is both reasonably large (slightly smaller than Cityscapes~\citep{CordtsCVPR16CityscapesDataset}) and flexible in terms of visual conditions: researchers can indeed choose between \textit{foggy}, \textit{dark}, \textit{rainy} and \textit{snowy} scenarios.  More importantly, samples are recorded from the same streets in such different conditions, allowing to properly assess the impact of adverse weather/daylight on the models (see examples in \fig{ACDC} (left)).  RainCityscape~\citep{HuCVPR19DepthAttentionalFeaturesSingleImageRainRemoval} and FoggyCityscape~\citep{SakaridisECCV18ModelAdaptationSyntheticRealSemanticDenseFoggy} (see examples in \fig{ACDC} (right)) are also extremely valuable in this direction, but in this case the weather conditions are simulated (on top of the real Cityscapes images). We think that these datasets are better suited than the currently used Cityscapes dataset and we expect that in the future DASiS methods will be also evaluated on these or similar datasets.

\subsection{DA and DASiS evaluation protocols}
\label{sec:evlauation}

There exist two main evaluation protocols in DA, namely, \textit{transductive} and \textit{inductive}. Transductive DA aims to learn prediction models that directly assign labels to the target instances available during training. In other words, the model aims to perform well on the sample set $\calD_\calT$ used to learn the model. Instead, the inductive UDA measures the performance of the learned models on held-out target instances that are sampled from the same target distribution, $\widehat{\calD_\calT} \sim D_\calT$. While in classical DA most often the transductive protocol is considered, in the case of DASiS, the \textit{inductive} setting is the preferred one.

Selecting the best models, hyper-parameter settings is rather challenging in practice. As described in \citep{SaitoICCV21TuneItTheRightWayUnsupervisedValidationDASoftNeighborhoodDensity}, many methods do hyper-parameter optimization using the risk computed on target domain's annotated samples, which contradicts the core assumption of UDA -- \ie not using any labels from the target set. Furthermore, in many papers, a clear description about how the final model has been selected for evaluation is often missing, making the comparisons between different methods rather questionable. Even if in the inductive evaluation protocol a different set is used to select the model, an obvious question arises: \textit{If the model has access to target labels for evaluation, why not using those labeled target samples to improve the model in a semi-supervised DA fashion?}

Fairer strategies such as transfer cross-validation \citep{ZhongECML10CrossValidationTL}, reverse cross-validation \citep{GaninJMLR16DomainAdversarialNN}, importance-weighted cross-validation~\citep{LongNIPS18ConditionalAdversarialDomainAdaptation} and deep embedded validation~\citep{YouICML19TowardsAccurateModelSelectionDA} rely on source labels, evaluating the risk in the source domain and/or exploiting the data distributions. However, these strategies remain sub-optimal due to the fact that they still rely on the source risk which is not necessarily a good estimator of the target risk in the presence of a large domain gap~\citep{SaitoICCV21TuneItTheRightWayUnsupervisedValidationDASoftNeighborhoodDensity}.

Instead, \citet{SaitoICCV21TuneItTheRightWayUnsupervisedValidationDASoftNeighborhoodDensity}  revisit the unsupervised validation criterion based on the classifier entropy and show that when the classification model produces confident and low-entropy outputs on target samples the target features are discriminative and the predictions likely reliable. However, they claim that such criterion is unable to detect when a DA method falsely align target samples with the source and incorrectly changes the neighborhood structure. To overcome this limitation, they propose a model selection method based on  soft neighborhood density measure to evaluate the discriminability of target features.

\begin{figure}[ttt]
\begin{center}
\includegraphics[width=\textwidth]{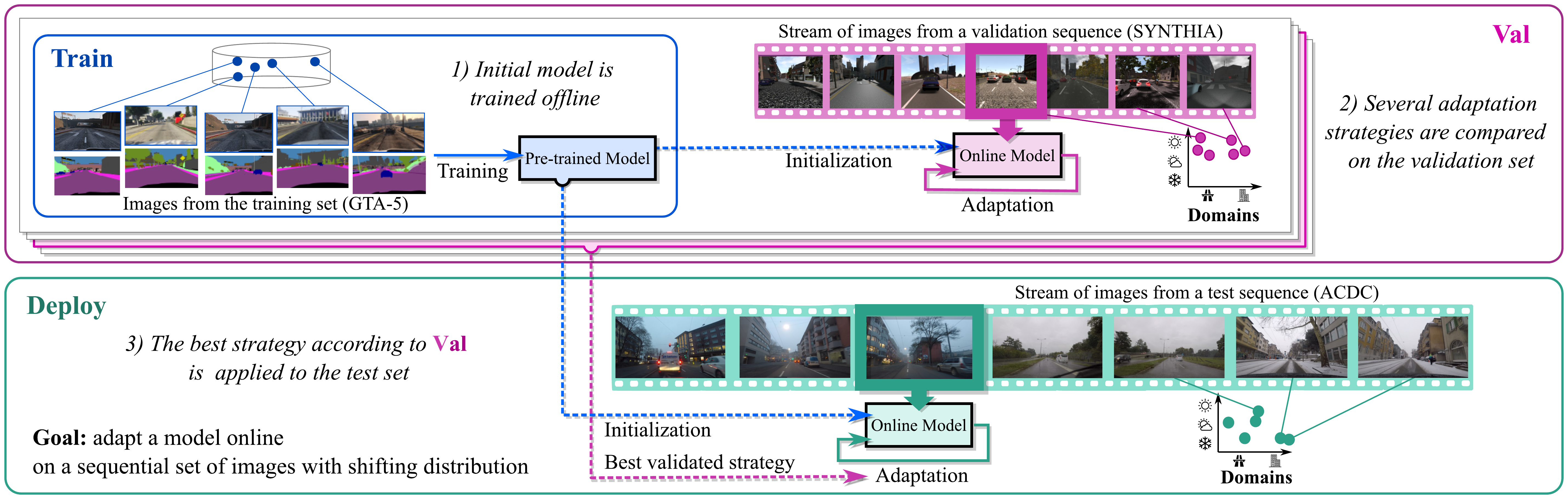}
\caption{The OASIS benchmark  addressing  evaluation of the  online, unsupervised adaptation 
    of semantic segmentation three steps. The  model is \textbf{trained} offline on simulated data (top-left), 
     several adaptation strategies can be  \textbf{validated} on simulated data organized in sequentially shifting domains
    (\eg, \textit{sunny-to-rainy}, \textit{highway-to-city}), to mimic deploy  (top-right), and \textbf{tested} on real data  (bottom).}
\label{fig:oasis}
\end{center}
\end{figure}

\subsection{Online adaptation for SiS protocols}
\label{sec:oasis-protocol}

In Section~\ref{sec:oasis} we had introduced the problem of online adaptation for SiS for which
\citet{VolpiCVPR22OnRoadOnlineAdaptSiS} propose a three-stage benchmark to train, validate and test corresponding 
algorithms 
(the OASIS benchmark). In general, the three steps are 1) pre-train a model on simulated data; 2) validate the adaptation algorithm on simulated \textit{sequences} of \textit{temporally correlated} samples; 3) test the validated model/method  on real sequences (see illustration in \fig{oasis}). In practice, they propose to use  GTA-5 dataset~\citep{RichterECCV16PlayingForData}  in 1), the SYNTHIA dataset~\citep{RosCVPR16SYNTHIADataset} in 2), and Cityscapes~\citep{CordtsCVPR16CityscapesDataset} (original and with artificial weather conditions) and ACDC~\citep{SakaridisICCV21ACDCAdverseConditionsDatasetSIS} datasets  for final testing in 3). The proposed pipeline allows evaluating the algorithm performance on environments that are unseen, both at training and validation,  mimicking real-world deployment in unfamiliar environments.

\chapter{Related Segmentation Tasks}
\label{c:relatedtasks}

In this chapter we discuss briefly some tasks that are closely related to SiS such as instance segmentation (Section~\ref{sec:instanceSegm}), panoptic segmentation (Section~\ref{sec:panopticSegm}) and medical image segmentation (Section~\ref{sec:MedSegm}).

\section{Instance Segmentation (InstS)}
\label{sec:instanceSegm}

\begin{figure}[ttt]
\begin{center}
\includegraphics[width=0.9\textwidth]{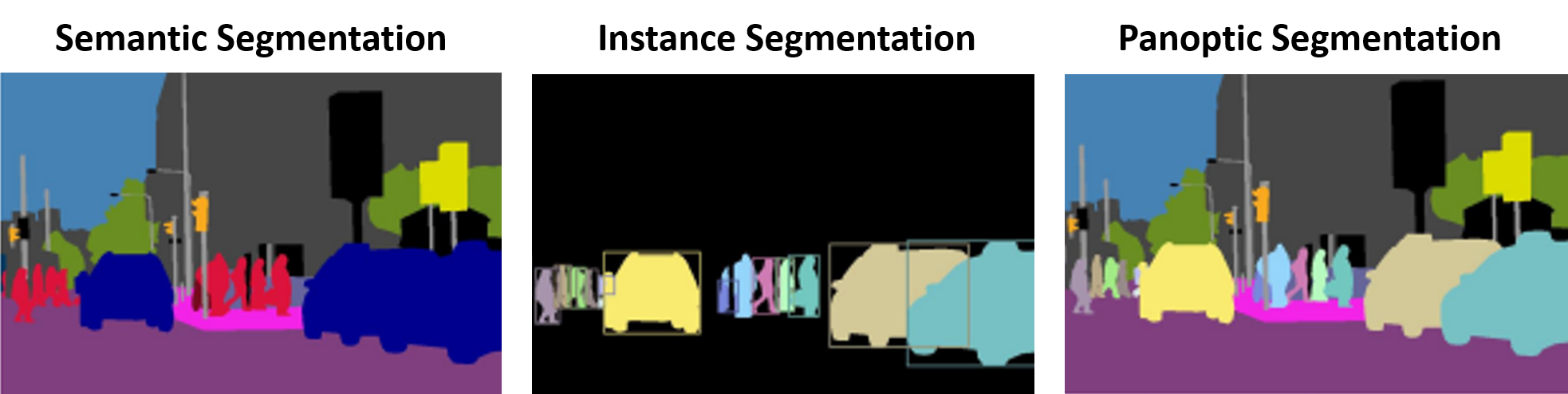}
\caption{Semantic image segmentation is related to Instance  Segmentation~\citep{YangPAMI12LayeredObjectModels4ImgSegm} and to Panoptic Segmentation~\citep{KirillovCVPR19PanopticSegmentation}. Instance Segmentation can be seen as a combination of object detection and semantic  segmentation  where the aim is to detect and segment all instances of a category in an image and such that each instance is uniquely identified. Panoptic Segmentation mixes semantic and instance segmentation, where  for some \textit{things} classes -- countable objects such as ``cars'', ``pedestrians'', \etc -- each instance is segmented individually, while for other classes especially those belonging to \textit{stuff} -- ``road'', ``sky'', ``vegetation'', ``buildings'' -- all classes are labeled with a single class label.}
\label{fig:InstSegm}
\end{center}
\end{figure}

SiS is strongly related to Instance Segmentation \citep{YangPAMI12LayeredObjectModels4ImgSegm}, which can be seen as a combination of object detection and semantic segmentation. The goal in InstS is indeed to detect and segment all instances of a category in a given image, 
while also ensuring
that each instance is uniquely identified (see illustration in~\fig{InstSegm} (middle)).

Early instance segmentation methods are based on complex graphical models \citep{SilbermanECCV12IndoorSegmentation,ZhangCVPR16InstanceLevelSegmentation4ADDeepDenselyConnectedMRFs,ArnabCVPR17PixelwiseInstSegmDynamicallyInstantiatedNetwork}, 
post-processing object detection~\citep{YangPAMI12LayeredObjectModels4ImgSegm,TigheCVPR14SceneParsingObjectInstancesOcclusionOrdering,ChenCVPR15MultiInstanceObjSegmentationWithOcclusionHandling}, or models built on top of segment region proposals~\citep{HariharanECCV14SimultaneousDetectionAndSegmentation,PinheiroNIPS15LearningToSegmentObjectCandidates}. 

Amongst more recent deep methods relying on object detectors, Mask R-CNN~\citep{HeICCV17MaskRCNN} is one of the most successful ones. It employs an R-CNN object detector~\citep{GirshickCVPR14RichFeatureHierarchies} and region of interest (RoI) operations -- typically RoIPool or RoIAlign -- to crop the instance from the feature maps.~\citet{LiuCVPR18PathAggregationNetwork4InstSegm} propose to further improve Mask R-CNN by 1) bottom-up path augmentation, which shortens the information path between lower layers and top most features, 2) by adaptive feature pooling and 3) by including a complementary branch that captures different views for each proposal.

\citet{NovotnyECCV18SemiConvolutionalOperators4InstSegm} extend Mask R-CNN with semi-convolutional operators,  which mix information extracted from the convolutional network with information about the global pixel location. YOLACT~\citep{BolyaICCV19YOLACTRealTimeInstanceSegmentation} and BlendMask \citep{ChenCVPR20BlendMaskTopDownMeetsBottomUpInstSegm} can be seen as a reformulation of Mask R-CNN, which decouple RoI detection and feature maps used for mask prediction. MaskLab~\citep{ChenCVPR18MaskLabInstanceSegmByRefiningObjDet} builds on top of Faster-RCNN~\citep{RenNIPS15FasterRCNNObjDetRegProposal} and for each RoI perform foreground/background segmentation by exploiting semantic segmentation and direction logits.

In contrast to the above \textit{detect-then-segment} strategies, many recent methods build on the top of deep segmentation models reviewed in \secref{deepSS}. FCN models, discussed in \secref{FCNSiS}, are amongst the most popular ones. One of such models, InstanceCut~\citep{KirillovCVPR17InstanceCutFromEdgesToInstances}, combines the output of two pipelines -- a FCN based SiS model and an instance-aware edge detection, processed independently, -- with an image partitionning block that merges the super-pixels into connected components with a class label assigned to each component. InstanceFCN~\citep{DaiECCV16InstanceSensitiveFCN}, instead of generating one score map per semantic class, computes  3$\times$3 position-sensitive score maps where each pixel corresponds to a classifier prediction concerning its relative positions to an object instance. \citet{LiCVPR17FullyConvolutionalInstanceAwareSiS} propose a fully convolutional Instance-aware SiS model where position-sensitive inside/outside score maps are used to perform object segmentation and detection jointly and simultaneously. The SOLO models~\citep{WangECCV20SOLOSegmentingObjectsByLocations,WangNIPS20Solov2DynamicFastInstSegm} assign categories to each pixel within an instance according to the instance’s location and size,  converting instance segmentation into a single-shot classification-solvable problem using FCNs to output dense predictions.

Dilated Convolutional Models (discussed in~\secref{DilatedSiS}), and in particular DeepLab-CRF-LargeFOV~\citep{ChenICLR15SemanticImgSegmFullyConnectedCRFs}, are fine-tuned and refined for InstS by~\citet{LiangPAMI18ProposalFreeNetwork4InstanceLevelObjSegm} and by~\citet{ZhangCVPR16InstanceLevelSegmentation4ADDeepDenselyConnectedMRFs}. The latter combines it with Densely Connected MRFs to improve instance boundaries (similarly to methods in \secref{GraphModSiS}).

\citet{RenCVPR17End2EndInstSegmRecurrentAttention} propose an end-to-end  RNN architecture with an attention mechanism
(see also
\secref{AttBasedSiS}). This model combines a box proposal network responsible for localizing objects of interest with a DeconvNet~\citep{NohICCV15LearningDeconvolutionNNSegmentation} to segment image pixels within the box.~\citet{ArunECCV20WeaklySupInstSegmLearningAnnotationConsistentInstances} modify a U-Net architecture where they explicitly model the uncertainty in the pseudo label generation process using a conditional distribution.

A transformer-based model (see \secref{tranfSiS}) is applied
by~\citet{XuICCV21CoScaleConvAttentionalImageTransformers} who propose a co-scale mechanism to image transformers, where encoder branches are maintained at separate scales while engaging attention across scales. They also design a Conv-attention module which performs relative position embeddings with convolutions in the factorized attention module. 

Finally, as for SiS (\secref{WSSiS}), a large set of weakly supervised methods that rely on bounding box supervision have been proposed also for instance segmentation.
For example,~\citet{TianNIPS19WeaklySupervisedInstSegmBBoxTightnessPrior} train jointly a Mask R-CNN detection and segmentation branches, estimate the object instance map inside each detected bounding box and then generate the positive and negative bags using the bounding box annotations. 
\citet{TianCVPR21BoxInstHighPerfInstSegmBoxAnnotations} extends this architecture with 
CondInst~\citep{TianECCV20ConditionalConvolutionsInstanceSegmentation} employing dynamic instance-aware networks, conditioned on instances which eliminates the need for RoI operations.~\citet{LanICCV21DiscoBoxWeakSupInstSegmSemCorrFromBoxSupervision} propose a self-ensembling framework where instance segmentation and semantic correspondences are jointly learned by a structured teacher and bounding box supervision. The teacher is a structured energy model incorporating a pairwise potential and a cross-image potential to model the pairwise pixel relationships both within and across the boxes.

\section{Panoptic Segmentation (PanS)}
\label{sec:panopticSegm}
 
Panoptic Segmentation \citep{KirillovCVPR19PanopticSegmentation}
unifies semantic and instance segmentation, where for several \textit{things} classes -- countable objects such as ``cars'', ``pedestrians'', \etc -- each instance is segmented individually, while for classes belonging to \textit{stuff} -- ``road'', ``sky'', ``vegetation'', ``buildings'' -- all pixels are labeled with a single class label (see illustration in~\fig{InstSegm}(right)).

\citet{KirillovCVPR19PanopticSegmentation}, emphasizing the importance to tackle semantic and instance segmentation jointly, they introduce the panoptic quality metric in order to evaluate jointly semantic and instance segmentation,  and thus open the path to a new set of methods called Panoptic Segmentation. The key idea is that for the \textit{things} classes the model has to predict both the belongings to the given \textit{things} class as well as distinguish the instances within the class, while for \textit{stuff} only the semantic class label needs to be assigned to the relevant pixels. 

To solve PanS, \citet{KirillovCVPR19PanopticSegmentation} propose to combine PSPNet~\citep{ZhaoCVPR17PyramidSceneParsingNetwork} with Mask R-CNN~\citep{HeICCV17MaskRCNN},  where the models process the inputs independently and then their outputs are combined in a post-processing step. \citet{deGeusX18PanopticSegmJointSiSAndInstSNetwork} propose to jointly train two branches with a shared backbone, one being a Mask R-CNN for the InstS and a second one relying on an Augmented Pyramid Pooling module for SiS. The Attention Guided Unified Network~\citep{LiCVPR19AttentionGuidedUnifiedNetworkPanopticSegm} combines a proposal attention module that selects regions potentially containing~\textit{things} with a mask attention module to refine the boundary between~\textit{things} and \textit{stuff}.

\citet{LiuCVPR19AnEnd2EndNetworkPanSegm} propose an end-to-end occlusion aware pipeline, where 1) the instance segmentation and stuff segmentation branches -- sharing the backbone features -- are optimized by the accumulated losses and 2) the head branches are fine-tuned on the specific tasks. A spatial ranking module addresses the ambiguities of the overlapping relationship. Instead,~\citet{XiongCVPR19UPSNetAUnifiedPanopticSegmentationNetwork} design a deformable convolution based SiS head and a Mask R-CNN based InstS head, and solve the two subtasks simultaneously. \citet{SofiiukICCV19AdaptISAdaptiveInstanceSelectionNetwork} propose a fully differentiable end-to-end network for class-agnostic instance segmentation which, jointly trained with an SiS Branch, can perform panoptic segmentation. 

\citet{LiECCV18WeaklySemiSupPanopticSegm}, building on top of the Dynamically Instantiated Network~\citep{ArnabCVPR17PixelwiseInstSegmDynamicallyInstantiatedNetwork}, propose a weakly supervised model for PanS where~\textit{things} classes are weakly supervised by bounding boxes, and \textit{stuff} classes with image-level tags.

Several PanS methods have been proposed on the top of DeepLab~\citep{ChenICLR15SemanticImgSegmFullyConnectedCRFs}. For instance,~\citet{PorziCVPR19SeamlessSceneSegmentation} propose an architecture which seamlessly integrates multi-scale features generated by an FPN~\citep{LinCVPR17FeaturePyramidNetworksObjDet} with contextual information conveyed by a lightweight DeepLab-like module.~\citet{YangX19DeeperLabSingleShotImageParser} adopt the encoder-decoder paradigm where SiS and InstS predictions are generated from the shared decoder output and then fused to produce the final image parsing result. This model has been extended by \citet{ChengCVPR20PanopticDeepLabBottomUpPanopticSegm}, by adding a dual-ASPP and a dual-decoder structure for each sub-task branch, and by \citet{WangECCV20AxialDeepLabStandAloneAxialAttPanopticSegm} where axial-attention blocks are used instead of ASPP. 

\citet{GaoFuICCV19SSAPSingleShotInstSegmAffinityPyramid} propose to jointly train semantic class labeling with a pixel-pair affinity pyramid that computes -- in a hierarchical manner -- the probability that two pixels belong to the same instance. Furthermore, they incorporate, with the learned affinity pyramid, a novel cascaded graph partition module to sequentially generate instances from coarse to fine. 
\citet{YuanECCV20ObjectContextualReprSemSegm} proposed  the Object-Contextual Representations (OCR) for SiS and generalized it to Panoptic Segmentation  where  the Panoptic-FPN~\citep{KirillovCVPR19PanopticFeaturePyramidNetworks}  head computes  soft object regions and then the OCR head predicts a refined semantic segmentation map.

The Efficient Panoptic Segmentation architecture~\citep{MohanIJCV21EfficientPSEfficientPanopticSegm} combines a semantic head that aggregates fine and contextual features coherently with a Mask R-CNN-like instance head. The final panoptic segmentation output is obtained by the panoptic fusion module that congruously integrates the output logits from both heads.

Amongst recent transformer-based solutions we can mention the Masked-attention Mask Transformer (Mask2Former)~\citep{ChengCVPR22MaskedAttentionMaskTransformer4UniversalSiS}  which extracts localized features by constraining cross-attention within predicted mask regions.

\section{Medical Image Segmentation}
\label{sec:MedSegm}
 
Medical image segmentation has an important role in sustainable medical care. With the proliferation of the medical imaging equipment,~\ie {\it computed tomography} (CT), {\it magnetic resonance imaging} (MRI), {\it positron-emission tomography}, {\it X-ray} and {\it ultrasound imaging} (UI), microscopy and fundus retinal images are widely used  in clinics, and medical images segmentation can effectively help doctors in their diagnoses
\citep{GreenspanTMI16GuestEditorialDeepLearningInMedicalImaging,AhujaPEERJ19TheImpactAIInMedicineFutureRolePhysician,KingJrJACR18ArtificialIntelligenceRadiologyFutureHold,EggerB21ComputerAidedOralMaxillofacialSurgery}. 
 
Here we only briefly mention a few works  on medical image segmentation  that heavily rely on architectures discussed in Chapters~\ref{c:semsegm} and~\ref{c:DAsemsegm}; for a detailed survey on medical image segmentation we refer the interested reader to~\citep{LiuJS21AReviewDeepLearningMedicalImageSegm}.
 
FCN and 3D-FCN based methods have been applied
for segmenting brain tumors~\citep{MyronenkoMICCAWSI173DMRIBrainTumorSegmAutoencoder,NieTC193DFCN4MultimodalInfantBrainImgSeg} or pathological lung tissues in MRI~\citep{NovikovTMI18FullyConvolutionalArchitecturesMultiClassSegmentationChestRadiographs,AnthimopoulosTC193DFCN4SemanticSegmPathologicalLungTissueDilatedFCN}, eye vessels in fundoscopy images ~\citep{EdupugantiICIP17AutomaticOpticDiskCupSegmFundusImages}, or skin lesions in dermatology images~\citep{MirikharajiMICCAI18StarShapePriorFCNSkinLesionSegm}. 

3D-Unet has been used by~\citet{BorneMIDL19Combining3DUNetCorticalSulciRecognition} to segment brain in MRI, by~\citet{YeTCSVT19MultiDepthFusionNetworkWholeHeartCTImgSegm} to segment heart in CT, by~\citet{ZhangMICCAI18DeepSupervisionAdditionalLabelsRetinalVesselSegm} to segment eye vessel in fundoscopy images.~\citet{OktayX18AttentionUNetLearningWhereToLook4Pancreas} propose Attention U-Net to segment pancreas in CT. 

A SegNet based network has been applied to segment musculoskeletal MRI images~\citep{LiuCGF18DeepCNN3DDeformableApproachTissueSegmMusculoskeletalMRI} and cells on microscopic images~\citep{TranICCCECE18BloodCellImagesSegmentation}. 
Different works rely on GAN-based models, in order to predict segmentation maps that are similar to humans' annotations. Such models have been used for MRI image segmentation~\citep{RezaeiMICCAWSI17ConditionalAdvNetworkBrainTumorSegmentation,MoeskopsMICCAWSI17AdversarialTrainingDilatedConvBrainMRISegmentation,HanMIM18SpineGANSemanticSegmentationMultipleSpinalStructures} and in histopathology~\citep{WangICMLC17AdversarialNN4BasalMembraneSegmHistopathology}.
 
DASiS solutions have been designed for MRI segmentation of liver and kidney~\citep{ValindriaMIDL18DomainAdaptationMRIOrganSegmentationReverseClassificationAccuracy}, neuroanatomy~\citep{NovosadX19UDAAutomatedSegmNeuroanatomyMRI},
retinal vessel~\citep{HuangMICCAIWS20DAPRNetDARetinalVesselSegm},
white matter hyper-intensities~\citep{OrbesArteagaMICCAIWS19MultiDomainAdaptationBrainMRIAdvLearning}, and multiple sclerosis lesions~\citep{AckaouyFCN20UDAOTMultipleSclerosisLesionsMRI}.
Furthermore,~\citet{BermudezChaconISBI18ADomainAdaptiveUNetElectronMicroscopy}
apply DASiS to microscopic image segmentation; ~\citet{DouIJCAI18UnsupervisedCrossModalityDABiomedicalImSegm} and ~\citet{JiangMICCAI18TumorAwareAdvDAFromCTtoMRILungCancerSegmentation} perform adaptation between CT and MRI images for cardiac structure segmentation and for lung cancer segmentation, respectively. \citet{VenkataramaniISBI19TowardsContinuousDA4MedicalImaging} propose a continuous DA framework for X-ray lung segmentation. Cross-center adaptation results of multiple sclerosis lesions and brain tumor segmentation have been considered by \citet{LiX20DAMedicaImSegmAdvLearningDiseaseSpecificSpatialPatterns}
and adaptation between gray matter segmentations \cite{PeroneNI19UDAMedicalImSegmSelfEnsembling}. 

\citet{LiMICCAI21FewShotDAWithPolymorphicTransformers} insert a polymorphic transformer (polyformer) into a U-Net model which relying on prototype embeddings, dynamically transforms the target-domain features making them semantically compatible with the source domain. They showcase their model on optic disc/cup segmentation in fundus images and polyp segmentation in colonoscopy images.

\chapter[Summary and Perspectives]{Conclusive Remarks}
\label{c:conclusions}

\section{Book Summary} 

In this book, we provide a comprehensive and up-to-date review of both  semantic image segmentation (SiS) in general as well as the domain adaptation of semantic image segmentation (DASiS) literature. We describe in both cases the main trends and organize  methods according to their most important characteristics. 

We extend the discussions on the two topics with scenarios that depart from the classical setting. In the case of SiS, we overview methods  exploiting unlabeled or weakly labeled data, curriculum or self-supervised strategies or methods learning the semantic classes incrementally. Concerning DASiS, we go beyond the typical single labeled source single unlabeled target and survey proposed methods for tasks such as multi-source or multi-target DA, domain incremental learning, source-free adaptation and domain generalization. We also discuss semi-supervised, active and online domain adaptation. 

We complement the discussion around SiS and DASiS topics with an extensive list of the existing datasets, evaluation metrics and protocols -- designed to compare different approaches. Finally, we conclude the book with a brief overview of three strongly related task, instance segmentation, panoptic segmentation and medical image segmentation. 

As the survey shows, both SiS and DASiS are very active research fields, with an increasing number of approaches being developed by the community and actively integrated in advanced industrial applications and solutions for autonomous driving, robot navigation, medical imaging, remote sensing, \etc
Therefore, we believe that the community can benefit from our survey -- in particular, PhD students and researchers who are just beginning their work in these fields, but also developers from the industry, willing to integrate SiS or DASiS in their systems, can find answers to their numerous questions.

\section{SiS with Additional Modalities} 

This book mainly focuses on SiS and DASiS, where raw images represent the only information available for scene understanding. However, both SiS and DASiS can benefit from additional visual information such as depth, 3D maps, text or other -- when available. There exists already a large amount of work in this direction and we expect that this line of research will grow further. Though out of the book scope, for the sake of completeness we highlight here some of the key directions. The interested reader can find more details in~\citep{ZhangIVC21DeepMultimodalFusion4SiSSurvey,FengITSC21DeepMultiModalObjDetSiSADDatasetsMethodsChallenges,ZhouIVC19AReviewDeepLearningMedicalImageSegmentationMultiModalityFusion}.

Additional visual modalities include near-infrared images \citep{SalamatiX14IncorporatingNearInfraredInformationIntoSemanticImageSegmentation,LiangCVPR22MultimodalMaterialSegmentation}, thermal images~\citep{HaIROS17MFNetTowardsRealTimeSiSADMultiSpectralScenes,SunRAL19RTFNetRGBThermalFusionNetworkSiSUrban}, 
depth~\citep{WangCVPR15TowardsUnifiedDepthAndSemanticPredictionSingleImage,QiICCV173DGraphNeuralNets4RGBDSiS,SchneiderSCIA17MultimodalNeuralNetworksRGBDSiSObjDetection}, 
surface-normals~\citep{EigenCVPR15PredictingDepthSurfaceNormalsSISMultiScaleCNN}, 3D LiDAR point clouds~\citep{KimBC18SeasonInvariantSiSDeepMultimodalNetwork,Jaritz3DV18SparseAndDenseDataCNNsDepthCompletionSiS,CaltagironeRAS19LIDARCameraFusionRoadDetectionFCN},~\etc. Any of these modalities brings additional information about a scene and can be used to learn a better segmentation model. 

One solution to address semantic segmentation with extra modalities is to deploy multi-modal fusion networks~\citep{HazirbasACCV16FuseNetIncorporatingDepthintoSemSegmViaFusionBasedCNN,LiECCV16LSTMCFUnifyingContextModelingFusionWithLSTMsRGBDSceneLabeling,ValadaICRA17AdapNetAdaptiveSemSegmAdverseEnvironmentalConditions,SchneiderSCIA17MultimodalNeuralNetworksRGBDSiSObjDetection,CaltagironeRAS19LIDARCameraFusionRoadDetectionFCN,SunRAL19RTFNetRGBThermalFusionNetworkSiSUrban} where multiple modalities are given as input to the system -- both at training and inference time -- and the model outputs pixel-level semantic labeling. To enhance the fusion between RGB and depth, \citet{HuICIP19ACNETAttentionBasedNetwork2ExploitComplementaryFeaturesRGBDSiS} propose to add Attention Complementary Modules between the single modality branches and the Fusion branch allowing the model to   selectively gather features from the RGB and depth branches. 

Alternatively, the second \textit{modality} is considered as privileged information given at training time but not at test time. 
Most works on this direction focused on joint monocular depth estimation and semantic segmentation showing that joint training allows improving the performance of both tasks~\citep{WangCVPR15TowardsUnifiedDepthAndSemanticPredictionSingleImage,Mousavian3DV16JointSemSegmDepthDCN,ZhangECCV18JointTaskRecursiveLearningSiSDepth,KendallCVPR18MultiTaskLearningUncertaintyWeighSceneGeometrySemantics,ChenICML18GradNormGradientNormalization4AdaptiveLossBalancingDeepMultitaskNetworks,HeNC21SOSDNetJointSemanticObjectSegmentationDepth}. A 
multi-task guided Prediction-and-Distillation Network was designed by \citet{XuCVPR18PADNetMultiTasksGuidedPredictionDistillationNetworkDepthSceneParsing}, where the model first predicts a set of intermediate auxiliary tasks ranging from low to high level, and then such predictions are used as multi-modal input to a multi-modal distillation module, opted at learning the final tasks.
\citet{JiaoECCV18LookDeeperIntoDepthMonocularDepthEstSiSAttentionDrivenLoss} rely on a synergy network to automatically learn information propagation between the two tasks. \citet{GaoAPPI22CINetContextInfoJointSiSDepth} use a shared attention block for the two tasks with contextual supervision and rely on a feature sharing module to fuse the task-specific features.

Similarly, extra modality was used as privileged information to improve the segmentation accuracy of DASIS methods, in particular using depth information available for the source data by~\citet{LeeICLR19SPIGANPrivilegedAdversarialLearningFromSimulation,VuICCV19DADADepthAwareDASemSegm,ChenCVPR19LearningSemSegmSyntheticGeomGuidedDA} and \citet{MordanHEART20BilinearMultimodalDiscriminatorAdvDAPrivilegedInformation}. Instead of using depth information as explicit supervision, \cite{GuiziliniICCV21GeometricDASS} infer and leverage depth in the target domain through self-supervision from geometric video-level cues, and use it as the primary source of domain adaptation.


\section{Perspectives in SIS}  

Concerning the perspectives, the most important one comes from the introduction of {\it foundation models}~\citep{YuanX21FlorenceNewFoundationModelCV} aimed at gaining and applying knowledge with good transferability. They consider the lifecycle of multiple deep learning applications as divided into two stages: pre-training and fine-tuning. 
In the first stage, the deep model is pre-trained on an upstream task with large-scale data (labeled or unlabeled) for gaining transferable knowledge. In the second stage, the pre-trained model is adapted to a downstream task in the target domain with labeled data. 

If the downstream task only has unlabeled data, then additional labeled data from another source domain of identical learning task but different data distribution 
can be used to improve the performance.
Compared with supervised pre-training,
self-supervised pre-training leads to competitive or sometimes even better performance on downstream tasks such as object detection and semantic segmentation~\citep{YuanX21FlorenceNewFoundationModelCV}. 

We believe that while these models provide good initialization for the methods discussed in this book, yet without undermining their value, we can foresee that future solutions will exploit and combine the strengths of both world. 

As an example we can mention \textit{Language driven Semantic Segmentation}~\citep{LiICLR22LanguageDrivenSemanticSegmentation} and 
\textit{Referring Image Segmentation}~\citep{HuECCV16SegmentationFromNaturalLanguageExpressions} which are emerging and challenging segmentation problems. Their  aim  is to segment a target semantic region in an image by understanding a given natural linguistic expression
In early solutions the models were trained on specific 
referring image segmentation datasets and where visual and linguistic features are simply concatenated \citep{LiuICCV17RecurrentMultimodalInteraction4ReferringImageSegmentation,LiCVPR18ReferringImageSegmentationViaRecurrentRefinementNetworks} or combined with Cross-Modal Self-Attention \cite{YeCVPR19CrossModalSelfAttentionNetworkReferringImageSegmentation}, 
using linguistic features to choose amongst visual target 
regions (proposed by \eg  Mask R-CNN) \citep{YuCVPR18MAttNetModularAttentionNetworkReferringExpressionComprehension},  or in a multi-task setting, by optimizing expression comprehension and segmentation simultaneously \citep{LuoCVPR20MultiTaskCollaborativeNetworkJointReferringExpressionComprehensionSegmentation}. More recent solutions are vision-language transformer based architectures~\citep{DingCVPR21VisionLanguageTransformerQueryGenerationReferringSegmentation}, which build upon and exploit the inherited knowledge of transformer-based joint language and vision models pretrained in a self-supervised manner on very large datasets. It is worth mentioning the successful Contrastive Language-Image Pre-training (CLIP)  model~\citep{RadfordICML21LearningTransferableVisualModelsFromNaturalLanguageSupervision}, used by \cite{WangCVPR22CRISCLIPDrivenReferringImageSegmentation} for referring image segmentation.

\backmatter  



\end{document}